\documentclass[lettersize,journal]{IEEEtran}
\usepackage{amsmath,amsfonts}
\usepackage[ruled,vlined]{algorithm2e}
\SetKwInOut{Input}{Input}
\SetKwInOut{Output}{Output}
\usepackage{array}
\usepackage{textcomp}
\usepackage{stfloats}
\usepackage{url}
\usepackage{verbatim}
\usepackage{graphicx}
\usepackage{cite}
\usepackage{float}
\usepackage{gensymb}

\begin{document}

\bstctlcite{IEEEexample:BSTcontrol}

\title{Active Cross-Modal Visuo-Tactile Perception\\ of Deformable Linear Objects}

\author{Raffaele Mazza, Ciro Natale, and Pietro Falco
\thanks{Raffaele Mazza and Ciro Natale are with Dipartimento di Ingegneria,
        Università degli Studi della Campania Luigi Vanvitelli, 81031 Aversa (CE), Italy, Pietro Falco is with Dipartimento di Ingegneria dell'Informazione, Università degli Studi di Padova, 35122 Padova, Italy.\\ (Corr. author: R. Mazza {\tt\small raffaele.mazza@unicampania.it})}}%


\markboth{IEEE/ASME Transactions on Mechatronics}%
{Shell \MakeLowercase{\textit{et al.}}: A Sample Article Using IEEEtran.cls for IEEE Journals}


\maketitle

\begin{abstract}
This paper presents a novel cross-modal visuo–tactile perception framework for the 3D shape reconstruction of deformable linear objects (DLOs), with a specific focus on cables subject to severe visual occlusions. Unlike existing methods relying predominantly on vision, whose performance degrades under varying illumination, background clutter, or partial visibility, the proposed approach integrates foundation-model-based visual perception with adaptive tactile exploration. The visual pipeline exploits SAM for instance segmentation and Florence for semantic refinement, followed by skeletonization, endpoint detection, and point-cloud extraction. Occluded cable segments are autonomously identified and explored with a tactile sensor, which provides local point clouds that are merged with the visual data through Euclidean clustering and topology-preserving fusion. A B-spline interpolation driven by endpoint-guided point sorting yields a smooth and complete reconstruction of the cable shape. Experimental validation using a robotic manipulator equipped with an RGB-D camera and a tactile pad demonstrates that the proposed framework accurately reconstructs both simple and highly curved single or multiple cable configurations, even when large portions are occluded. These results highlight the potential of foundation-model-enhanced cross-modal perception for advancing robotic manipulation of deformable objects.
\end{abstract}


\section{Introduction}
Robotic manipulation of deformable linear objects (DLOs), such as cables, hoses, and wires, is a fundamental capability for automation in manufacturing, logistics, and domestic environments. Tasks including wiring harness assembly, cable routing, and switchgear cabling require accurate knowledge of the object geometry during interaction. However, the elastic and highly deformable nature of DLOs makes their shape difficult to estimate reliably, particularly under realistic sensing conditions. Most existing approaches rely predominantly on visual perception, using RGB or RGB-D sensing to segment and reconstruct the object shape. While vision-based pipelines can achieve high accuracy in controlled settings, their performance degrades in the presence of occlusions, background clutter, illumination changes, or partial visibility—conditions that frequently occur in real industrial scenarios. As a result, visual perception alone is often insufficient to provide a complete and reliable representation of DLO geometry.
\begin{figure}[t]
    \centering 
\includegraphics[width=\columnwidth]{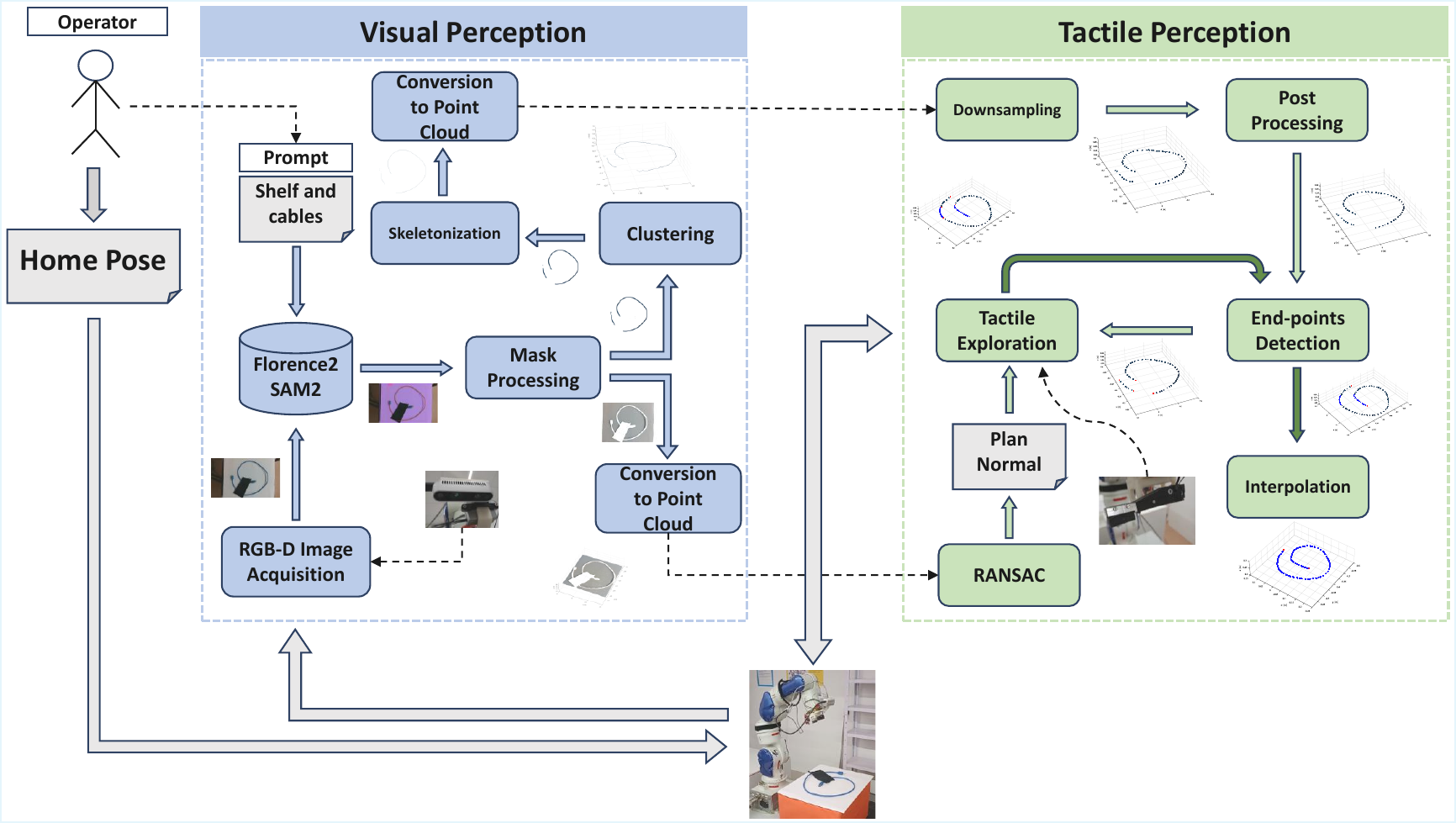} %
    \caption{Block diagram of the proposed cross-modal visuo-tactile perception framework: visual perception is triggered by a natural language prompt for a state-of-the-art segmentation network; a clustering algorithm isolates different cables based on their colour; the segmented image is converted into point clouds and post-processed triggering the tactile exploration; finally a B-spline interpolation provides a model of the DLO.}
    \label{fig:pipeline}
\end{figure}
Tactile sensing offers a complementary modality capable of capturing local geometric information even when visual data are unavailable. Prior work has shown that tactile feedback can significantly improve robustness during manipulation. However, tactile sensing is typically exploited only locally, after grasping, and is rarely integrated into a global shape reconstruction framework.
Inspired by human perception, which adaptively combines vision and touch based on local reliability, this paper proposes a cross-modal visuo--tactile perception framework for global DLO shape reconstruction under partial observability. The basic idea was preliminary shown in the extended abstract~\cite{Mazza2024}. The approach integrates foundation-model-based visual perception with autonomous tactile exploration to compensate for missing or unreliable visual information. Visual segmentation leverages state-of-the-art foundation models to extract a topology-aware representation of cable segments, while visually occluded regions are actively explored using a tactile sensor. The resulting visual and tactile point clouds are fused into a unified geometric model, yielding a smooth and continuous reconstruction of the DLO shape.

The proposed method is validated on a real robotic platform equipped with an RGB-D camera and a tactile sensor. Experimental results demonstrate reliable reconstruction of single and multiple cables under severe occlusions, inclined surfaces, and self-intersections, highlighting the effectiveness of foundation-model-enhanced cross-modal perception for deformable object manipulation.

 \section{Related Work}
\label{sec:related}
In this section, we first review general surveys on deformable object manipulation, and then focus on deformable linear objects (DLOs) such as cables and wires. We finally position our contribution with respect to the existing works.

\subsubsection{Surveys on Deformable Object Manipulation}
An early comprehensive survey by S{\'a}nchez et al.\ review robotic manipulation and sensing of deformable objects across domestic and industrial applications, emphasizing the variety of sensing modalities and control strategies required compared to rigid objects \cite{Sanchez2018Survey}. Yin et al.\ provide a structured overview of modeling, learning, control, and planning methods for deformable object manipulation, highlighting the need to combine analytical models with data-driven approaches \cite{Yin2021Modeling}. 
Zhu et al.\ outline key challenges and outlooks in deformable object manipulation (DOM), stressing the need for integrated perception–planning–control pipelines and standardized benchmarks \cite{Zhu2022Challenges}. These works underline that, despite rapid progress, perception of deformable objects under occlusions and in cluttered industrial scenes remains a central bottleneck.

\subsubsection{State Estimation of Deformable Linear Objects}
Within DOM, DLOs such as cables and wires are especially relevant for industrial tasks such as wiring harness assembly. A first family of works focuses on visual segmentation and geometric reconstruction. Caporali et al.\ proposed FASTDLO, a fast instance-segmentation pipeline tailored to DLOs that combines synthetic training data and skeleton extraction for robust segmentation in RGB images \cite{Caporali2022FASTDLO}. Sun et al.\ extend this idea to 3D, presenting a robust pipeline from RGB-D segmentation to 3D reconstruction of DLOs that explicitly addresses sparse depth, self-occlusions, and background clutter \cite{Sun2023}. Keipour et al.\ introduce a deformable one-dimensional object detector for routing and manipulation, focusing on robust detection of cables and hoses in cluttered scenes \cite{Keipour2022D1}.
Beyond pure vision, several works utilise multi-view or temporal information for DLO state estimation. Xiang et al.\ introduce TrackDLO, which tracks DLOs under occlusion by enforcing motion coherence over a sequence of RGB-D frames, achieving real-time performance without requiring markers or simulation \cite{Xiang2023TrackDLO}. 
Huang et al.\ address the perception of multiple DLOs of unknown number in complex backgrounds and propose a deep-learning approach for untangling and unknotting from visual input \cite{Huang2024UntanglingMultiDLO}. 
Learning-based state estimation bridges the gap between perception and physics. Yan et al.\ propose a self-supervised pipeline that learns to estimate the state of linear deformable objects from images, combined with a differentiable dynamics model for model-predictive control \cite{Yan2020SelfSupervised}.  Zhang et al.\ address robust vision-based control of DLOs in the presence of occlusions, coupling vision-based state estimation with feedback control \cite{Zhang2022RobustVB}. These works provide important building blocks, but typically assume static camera viewpoints and do not explicitly optimize sensor poses to reduce occlusions.
\subsubsection{Active Perception}
The notion of active perception, formalized by Bajcsy as the intentional control of sensing to improve perception \cite{Bajcsy1988Active}, is particularly relevant for occlusion-heavy DLO scenarios. 
Breyer et al.\ presented a closed-loop NBV strategy for target-driven grasping, where future camera poses are selected based on grasp metrics and uncertainty \cite{Breyer2022NBV}. 
In the context of deformable objects, Weng et al.\ introduce Dynamic Active Vision Space (DAVS) for interactive perception of deformable objects, coupling camera motion with object manipulation in a sequential decision-making framework \cite{Weng2024DAVS}. 
Recent work explores occlusion-aware active perception of tangled DLOs using Gaussian Splatting, where a volumetric representation is used to reason about occluded cable parts and plan informative viewpoints \cite{Chen2024GaussianDLO}. 
These contributions show that NBV and active vision can drastically improve perception under occlusions. However, they typically either address rigid objects, or treat DLOs at a purely geometric level, without exploiting the specific topological structure of cables or integrating tactile/force cues.

\subsubsection{Interactive Perception and Visuo–Tactile Fusion}
Interactive perception uses purposeful manipulation to probe the environment and improve state estimation. For DLOs, Caporali et al.\ proposed a method for manipulating DLOs while estimating model parameters online, thereby coupling interaction and parameter identification in the loop \cite{Caporali2024OnlineParams}. S{\"u}berkr{\"u}b et al.\ present an interactive segmentation strategy based on motion correlation: by moving the cable, the robot segments it from the background via correlated motion in the image \cite{Suberkrub2025MovingCable}. McConachie and Berenson study learning-based interactive perception for deformable objects without demonstrations, using reinforcement learning to decide exploratory actions \cite{McConachie2021Interactive}. 
From a mechatronic perspective, visuo–tactile fusion is particularly attractive for the scientific community. 
Pecyna et al.\ integrate visual and tactile sensing in an RL framework for following deformable linear objects, demonstrating that tactile information significantly improves robustness \cite{Pecyna2022VisuoTactile}. 
Palli et al.\ report a case study on tactile-based control for switchgear cabling, showing that tactile feedback can be used to stabilize contact and reduce damage in industrial wiring tasks \cite{Palli2021Wires}. Datta et al.\ propose 3D-ViTac, a visuo–tactile framework for fine-grained manipulation that combines 3D vision and tactile features in a unified representation \cite{Datta2024ViTac}. These works highlight that multi-modal sensing is key for robust DLO manipulation, but active viewpoint planning is limited or not explicitly optimized.


\subsubsection{Emerging Generative Approaches}
Large-scale learning and foundation models are beginning to influence DLO perception and manipulation. Caporali et al.\ introduced DLO Perceiver, which grounds a large language model to perform text-guided segmentation and perception of DLOs from images \cite{Caporali2024DLOPerceiver}. Emerging works explore multi-robot assembly of DLOs using multi-modal perception \cite{Chen2025MultiRobotDLO} or generative 3D state estimation from partial views \cite{tang2025generative}. These methods suggest that combining rich generative models, multi-modal sensing, and active perception can improve performance in complex manipulation tasks of DLOs.

\subsubsection{Our Contribution}
To the best of our knowledge, no prior work provides a 
\textbf{fully autonomous foundation-model-driven} and \textbf{cross-modal} 
reconstruction pipeline capable of estimating the \emph{global} shape of a DLO 
when large portions are visually occluded. Existing cross-modal approaches 
focus mainly on object recognition \cite{murali2022, falco2019}, but do not tackle the reconstruction 
of continuous deformable structures such as cables.
Within this context, the present work advances the state of the art by introducing a novel visuo–tactile perception framework that unifies foundation models, autonomous tactile exploration, and spline-based geometric reconstruction. The main contributions are:
\begin{itemize}
    \item \textbf{Topology-aware visual representation through skeletonization and endpoint detection.}
    Unlike classical vision pipelines that treat DLOs as unstructured point 
    clusters, our method extracts the underlying curve structure of the cable.
    This yields a representation that is robust to noise, downsampling, and 
    partial occlusions, and facilitates principled integration with tactile data.

    \item \textbf{Autonomous tactile exploration of visually occluded regions.}
    When the visual reconstruction is incomplete, an occlusion detector triggers 
    an active scanning procedure with a tactile sensor. The resulting tactile point 
    clouds are segmented and merged with visual data to restore missing geometry. 
    Such an active compensation mechanism is absent from existing DLO reconstruction 
    methods.

    \item \textbf{Unified point-cloud fusion and B-spline reconstruction.}
    We propose an endpoint-guided point sorting strategy followed by B-spline 
    interpolation to obtain a smooth and globally consistent cable model. This 
    reconstruction remains effective even in challenging cases involving multiple 
    disconnected occluded segments.
\end{itemize}

\section{Methodology}
\label{sec:methodology}
This work proposes a deterministic cross-modal visuo-tactile methodology for reconstructing the shape of DLOs, specifically cables, under partial observability due to visual occlusions or imperfect processing of visual data. The approach integrates visual perception and active tactile exploration within a unified geometric reconstruction framework depicted in Fig. \ref{fig:pipeline} whose modules are discussed in detail hereafter, while a formal description is provided in Algorithm~\ref{alg:crossmodal_overview}. The active tactile exploration strategy employed to recover occluded cable segments is formalized in Algorithm~\ref{alg:tactile_exploration}.

\subsection{Visual Perception}
An RGB-D camera mounted on the robot end effector acquires an image $\mathcal{I}$ of the scene. The pre-trained network Florence2 \cite{ravi2025sam} performs a semantic segmentation by using the prompt \textit{``shelf and cables"}. The results of Florence2 are the bounding boxes of the objects in the scene, then processed by SAM2 \cite{Xiao_2024_CVPR} to create their segmentation masks $\mathcal{M}$. 

\subsection{Image Processing}
The corresponding pixels of the segmentation masks are processed in a different way. The pixels of the shelf represented by $\mathcal{I}_B$ are converted into a point cloud $\mathcal{P}_B$ to apply the RANSAC algorithm, available in the Point Cloud Library (PCL), which creates a new point cloud matching the model of a plane. The plane coefficients allow obtaining the vector $\hat{z}$, normal to the plane, used to set the orientation of the end effector during the tactile exploration. The pixels of the detected cables $\mathcal{I}_{DLO}$ are processed with two standard operations available in the OpenCV library: image blurring and contour removal. These preliminary steps are necessary to facilitate the unsupervised clustering algorithm HDBSCAN \cite{HDBSCAN}, which returns the set of clusters $\mathcal{C}$. Subsequently, a skeletonization algorithm is applied to reduce the number of pixels of interest without losing information about the topology of the DLO. Finally, the skeletonized cluster $\mathcal{C}_i$ is converted into a point cloud $\mathcal{P}_{skeleton}$.

\subsection{Point Cloud Processing}

To reduce the number of points, a downsampling method is applied: all points in a 3D cube are approximated to their centroids. The downsampled point cloud $\mathcal{P}_{down}$ may have some points that are too close to each other and, in turn, can generate a non-smooth curve, making the sorting algorithm used to search for the DLO endpoints hard to converge because the algorithm performs well only when the point cloud of the cables is smooth with respect to the variation in direction between two points of the point cloud. So, when the Euclidean distance between two points of the point cloud is below a threshold $t_P$, they are replaced by the midpoint. Finally, the plane normal $\hat{z}$ is used to project all the points of $\mathcal{P}_{down}$ onto the plane and obtain the point cloud $\mathcal{P}_{proj}$.

\subsection{Endpoints Search}
The active tactile perception requires the endpoints of each cluster; therefore, endpoint detection is performed using a sorting algorithm based on the idea of following the cable direction on the plane. The details of the algorithm can be found in \cite{Mazza2024}.
Finally, the first point and the last point of the sorted point cloud $\mathcal{P}_{sorted}$ represent the endpoints of the cables. However, the detected points may not be the true endpoints due to either the presence of occluded parts of the DLO or imperfect segmentation, which can be emphasized through the removal of some pixels or points during the previous steps of the pipeline. In conclusion, a single cluster that represents a cable could be split into a certain number of segments with their own endpoints after the sorting. The set of endpoints $\mathcal{E}$ is exploited during the tactile exploration to reconstruct the missing parts of the DLO.

\subsection{Active Tactile Exploration}
\begin{figure}[b]
    \centering 
\includegraphics[width=0.5\columnwidth]{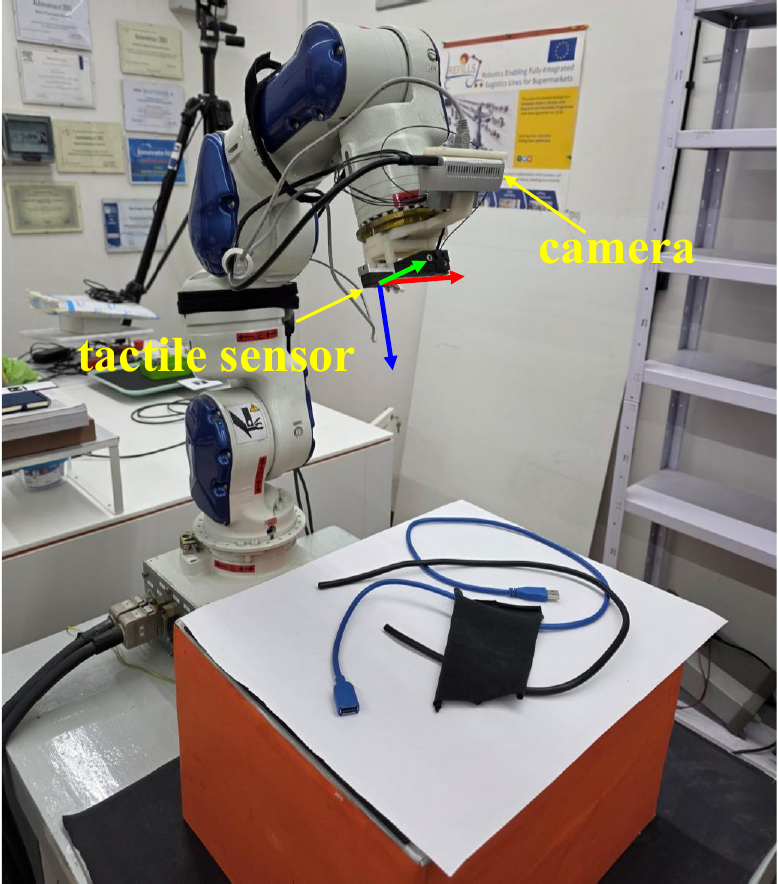} %
    \caption{Experimental setup: robot arm with camera and tactile sensor.}
    \label{fig:setup_picture}
\end{figure}

The robot relies on both the knowledge acquired through visual perception and endpoint detection to perform the gradient-based tactile exploration illustrated in Algorithm \ref{alg:tactile_exploration}. For each endpoint $e$, the initial end-effector orientation $R_d$ is computed by using the plane normal $\hat{z}$ and the sorted point cloud $\mathcal{P}_{sorted}$. The sorted point cloud is useful to select the previous point with respect to the endpoint $e$ and to compute the desired direction along the $y$ axis of the end effector frame, while the direction along the $z$ axis is defined by the estimated normal $\hat{z}$. Note how the $y$ direction has been chosen as the direction along the cable because it is orthogonal to the longest side of the tactile pad (see Fig. \ref{fig:setup_picture}), hence more taxels will likely be activated at the contact. Since the $y$ axis represents the direction along the cable, the new exploration point is defined as the endpoint with an additional term $\Delta y$ along the direction of the cable. Once the robot has been brought to the new exploration point $t^{b}_{e}$, there is a descending phase along the direction $\hat{z}$ by the offset $\Delta z$ until the tactile sensor detects a touch. The tactile map $\mathcal{T}$ is acquired at each contact. Let $H_{ij}$ be the $ij$-th element of a $6 \times 2$ matrix $H$ containing the norms of the Hessian matrices $\nabla^2 \mathcal{T}_{ij}$ computed in each point of the tactile map $\mathcal{T}$. The \textit{indicator} computed as $\bigl\lVert H \bigr\rVert$ is used as a curvature-based metric to discriminate between contacts with the flat support surface and contacts with the cable. If the metric exceeds a predefined threshold $t_H$, the centroid of the tactile map is projected onto the plane and added to the visual point cloud $\mathcal{P}_{tactile}$. The last added point (the endpoint $e$ at the first iteration) and the actual sampled point $p_{new}$ are used to compute the new end effector orientation, while the target point is defined by following the same logic explained above. The algorithm proceeds iteratively and terminates after all the endpoints are processed. The exploration along the cable is stopped when the Euclidean distance between $p_{new}$ and at least one endpoint is below the threshold $d_{min}$. After reaching an endpoint of the cable, it is marked as a non-endpoint and will not be evaluated as a starting point for another iteration of the algorithm. If the metric does not exceed the threshold $t_H$, the desired orientation $R_d$ is modified by applying a rotation about $\hat{z}$ with a rotation matrix $R_z(\theta)$ in order to explore along other directions before concluding that there are no cable parts. While the thresholds are empirically chosen, they have been demonstrated to be stable across all tested scenarios, suggesting robustness to variations in contact conditions. All parameters of the algorithms are summarized in Tab. \ref{t:parameters}.

\subsection{Interpolation}

The point cloud $\mathcal{P}_{merged}$ is obtained through the merging operation between $\mathcal{P}_{sorted}$ and $\mathcal{P}_{tactile}$, and it needs to be sorted again to apply the B-spline and obtain the model of the cable. In conclusion, the result is the point cloud $\mathcal{P}_{interpolated}$ where the first point and the last point of the cloud represent the true endpoints of the DLO.

\begin{algorithm}
\scriptsize
\caption{Cross-Modal Visuo-Tactile DLO Reconstruction}
\label{alg:crossmodal_overview}
\Input{Visual point cloud $\mathcal{P}_v$; parameters $d_m, t_P, t_H$}
\Output{Reconstructed B-spline curve $\mathcal{S}$}

$\mathcal{I} \gets$ \texttt{AcquireImage}$()$ \\
$\mathcal{M} \gets$ \texttt{SemanticSegmentation}$(\mathcal{I})$ \\
$\mathcal{I}_B,$ $\mathcal{I}_{DLO}$ $\gets$ \texttt{MaskProcessing}$(\mathcal{M})$ \\
$\mathcal{P}_B \gets$ \texttt{ImageToPointCloud}$(\mathcal{I_{B}})$ \\
$\hat{z} \gets$ \texttt{RANSAC}$(\mathcal{P}_B)$ \\
$\mathcal{C} \gets$ \texttt{HDBSCAN}$(\mathcal{I}_{DLO})$ \\

\ForEach{$\mathcal{C}_i \in \mathcal{C}$}
{
    $\mathcal{C}_S \gets$ \texttt{Skeletonization}$(\mathcal{C}_i)$ \\
    $\mathcal{P}_{skeleton} \gets$ \texttt{ImageToPointCloud}$(\mathcal{C}_S)$ \\
    $\mathcal{P}_{down} \gets$ \texttt{DownsampleAndFilter}$(\mathcal{P}_{skeleton}, \ d_m, \ t_P)$ \\
    $\mathcal{P}_{proj} \gets$ \texttt{PlaneProjection}$(\mathcal{P}_{down}, \ \hat{z})$ \\
    $\mathcal{P}_{sorted}, \ \mathcal{E} \gets$ \texttt{EndpointsDetection}$(\mathcal{P}_{proj})$ \\
    $\mathcal{P}_{tactile} \gets$ \texttt{TactileExploration}$(\mathcal{P}_{sorted}, \ \mathcal{E}, \ \hat{z}, \ t_H)$ \\
    $\mathcal{P}_{merged} \gets$ $\mathcal{P}_{sorted} \ \cup \ \mathcal{P}_{tactile}$ \\
    $\mathcal{P}_{sorted}, \ \mathcal{E} \gets$ \texttt{EndpointsDetection}$(\mathcal{P}_{merged})$ \\
    $\mathcal{P}_{interpolated} \gets$ \texttt{BSplineInterpolation}$(\mathcal{P}_{sorted})$ \\
}

\end{algorithm}

\begin{algorithm}
\scriptsize
\caption{Gradient-based Tactile Exploration}
\label{alg:tactile_exploration}
\Input{Sorted Visual Point Cloud  $\mathcal{P}_{sorted}$, Endpoints Set $\mathcal{E}$, Estimated Plan Normal $\hat{z}$, parameters $t_H, \ \Delta y, \ \Delta z, \ d_{min}, \ \theta$}
\Output{Tactile Point Cloud $\mathcal{P}_{tactile}$}
    \ForEach{$e \in  \mathcal{E}$}
    {
        $R_d \gets $ \texttt{ComputeEEOrientation($e, \ \hat{z} , \ \mathcal{P}_{sorted}$)} \\
        \While{$\neg$ stop}
        {
            $t^b_{e} \gets $ \texttt{ComputeExplorationPoint($\Delta y$)} \\
            \texttt{Robot.goTo($R_d$,$t^b_e$)} \\
            \While{$\neg$ touched}
            {
                $t^b_{e} \gets $ \texttt{ComputeExplorationPoint($\Delta z$)} \\
                \texttt{Robot.goTo($R_d$,$t^b_e$)} \\
                \If{touched}
                {
                    $\mathcal{T} \gets $ \texttt{AcquireTactileMap()} \\
                    \texttt{indicator} $\gets$ \texttt{compute\_indicator($\mathcal{T}$)} \\
                    \If{indicator $> t_H$}
                    {
                        $p_{new} \gets $ \texttt{ComputeCentroid($\mathcal{T}$)} \\
                        $\mathcal{P}_{tactile}$.\texttt{add($p_{new}$)} \\
                        \ForEach{$e_p \in \mathcal{E}$}
                        {
                            \If{EuclideanDistance($p_{new}, e_p$) $< d_{min}$ }
                            {
                                \texttt{stop} $\gets$ \texttt{true} \\
                            }
                        }
                        $R_d$ $\gets$ \texttt{ComputeEEOrientation($p_{new}, \ \hat{z}$, \ $\mathcal{P}_{sorted}$)}
                    }
                    \Else
                    {
                            $R_d \gets R_dR_z(\theta)$ \\
                    }
                }
            }
        }
    }
    
\end{algorithm}
\begin{table}[tph]
\caption{Parameters of the cable reconstruction algorithm}
\begin{center}
\begin{tabular}{|c|c|c|c|c|c|c|}
\hline
$d_{min}\,$[m] & $d_m\,$[m] & $t_P\,$[m] & $t_H\,$[m] & $\Delta y\,$[m] & $\Delta z\,$[m] & $\theta\,$[\degree] \\
\hline
0.0150 & 0.0200 & 0.0080 & 0.0011 & 0.0100 & 0.0015 & 15 \\
\hline
\end{tabular}
\label{t:parameters}
\end{center}
\end{table}

\section{Experimental Results}
The experimental setup (see Fig. \ref{fig:setup_picture}) includes a 7-axis Yaskawa Motoman SIA5F robot equipped with an eye-in-hand Intel Realsense D435i RGB-D camera and a SUNTouch tactile\cite{costanzo2019suntouch} sensor. The camera is used with an RGB resolution of $1280\times 720$ and a depth resolution of $848\times 480$ both at $30\,$fps, while the tactile sensor has a taxel matrix with dimensions $6 \times 2$ and provides tactile readings at $1000\,$Hz.

The experiments are carried out in two case studies: a single cable on an inclined plane (CS1) and two cables on a horizontal plane (CS2). For each case study, we show two situations: (i) cables are not occluded, (ii) some cable parts are occluded by another object.

\subsection{Single cable on an inclined plane}
\subsubsection*{Cable with no occlusions}
\begin{figure}[t]
    \centering
    \includegraphics[width=0.4\columnwidth]{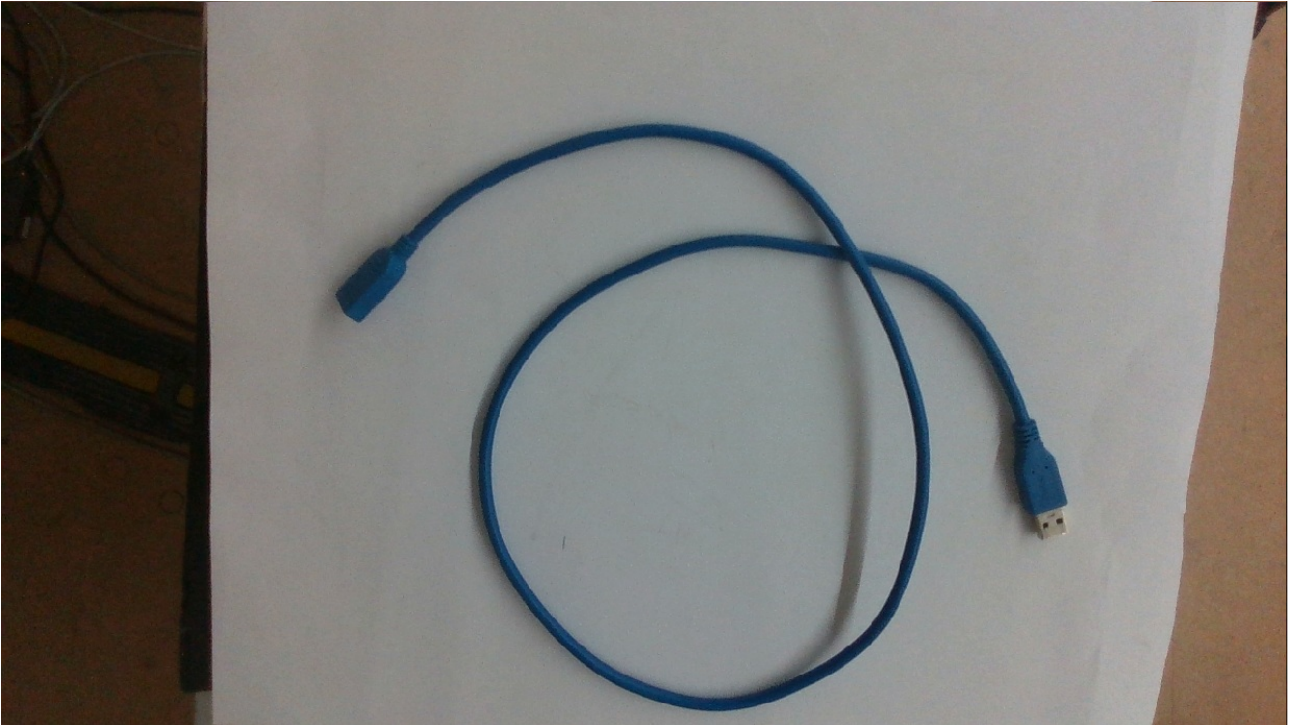}
    \includegraphics[width=0.4\columnwidth]{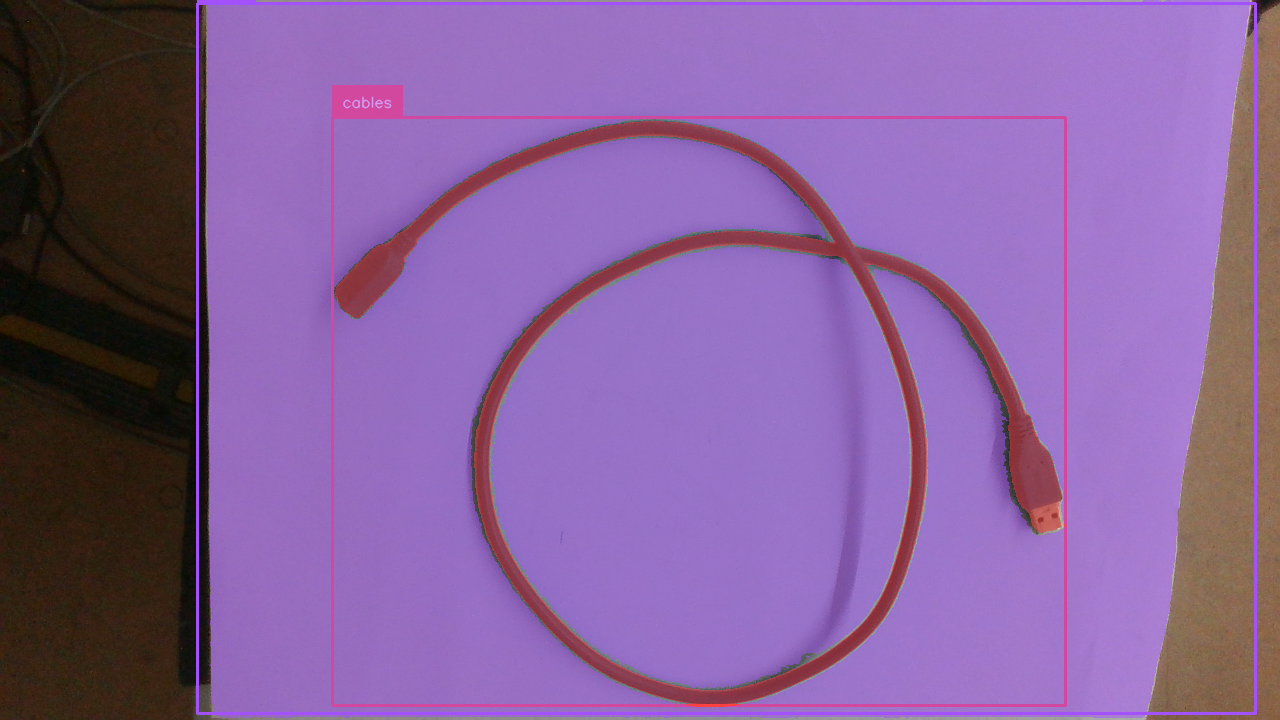}\\[0.5cm]
    \includegraphics[width=0.4\columnwidth]{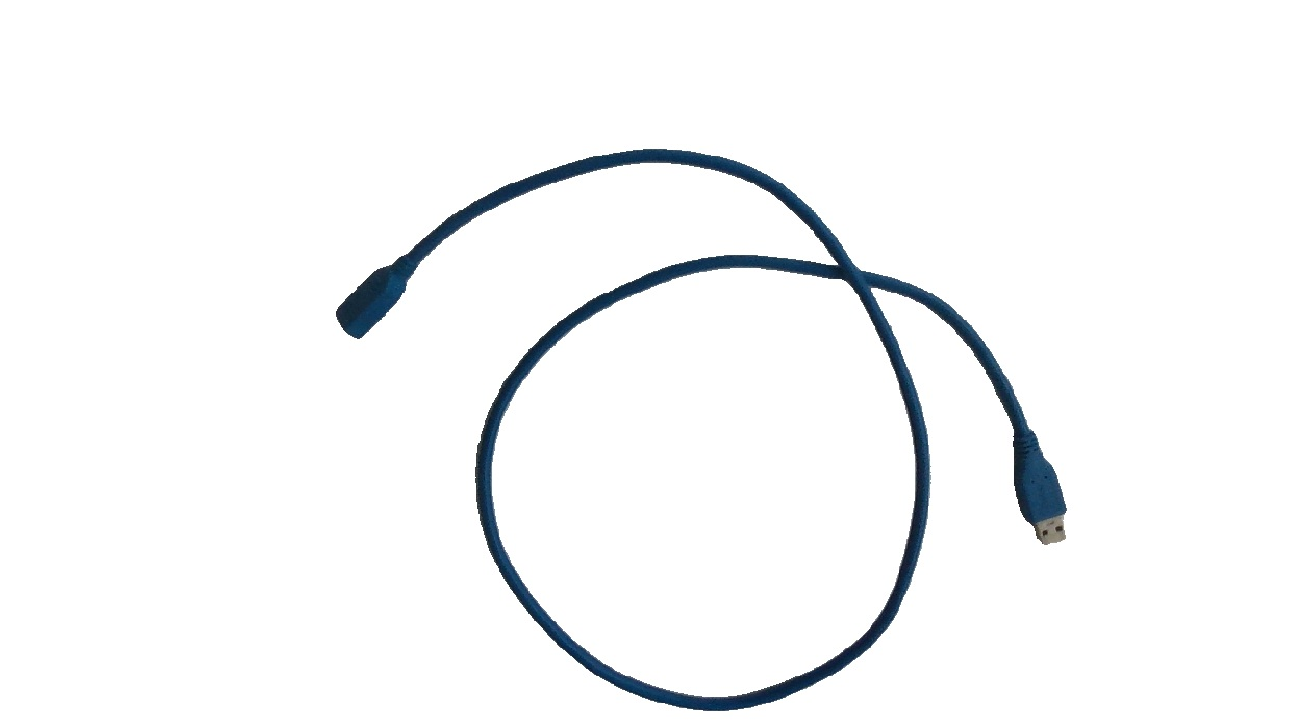}
    \includegraphics[width=0.4\columnwidth]{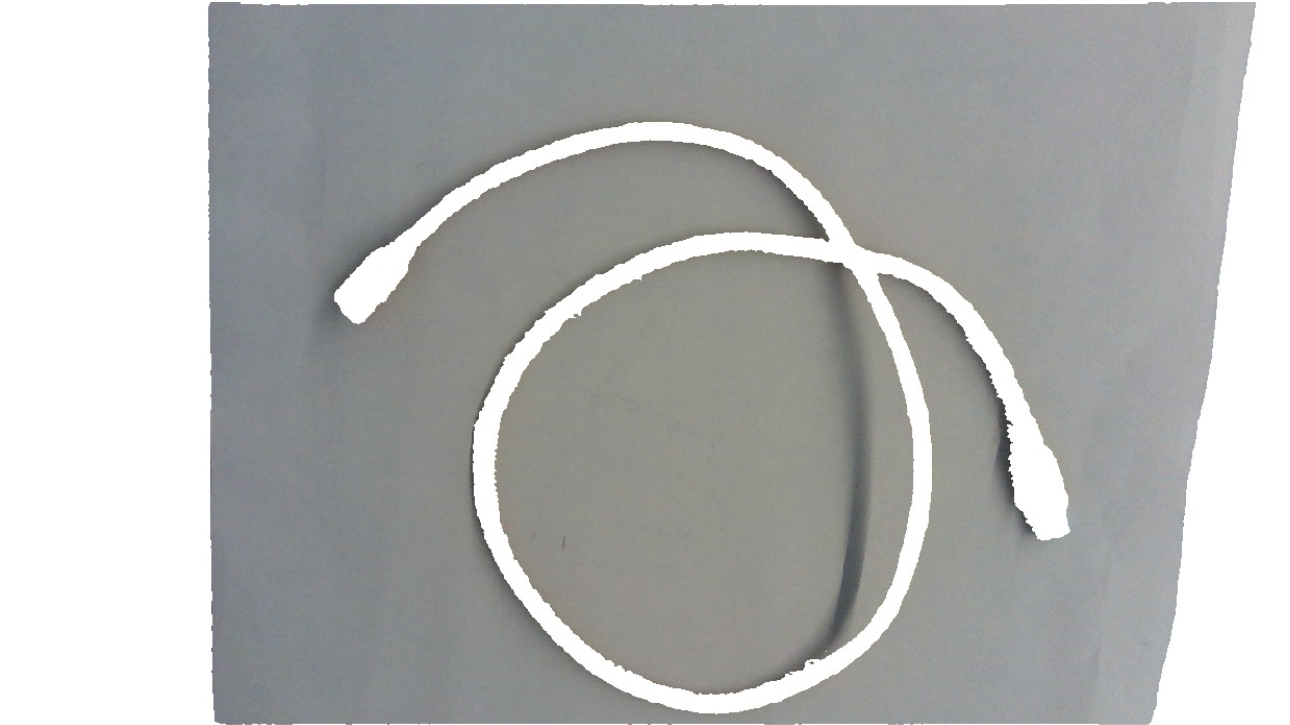}
\caption{CS1 (i): camera RGB image (top-left), result of the semantic segmentation by Florence2/SAM2 (top-right), segmented cable (bottom-left) and support surface (bottom-right).}
\label{fig:1cable_Inc_noOcc_image+sam2}
\end{figure}

\begin{figure}[t]
    \centering
    \includegraphics[width=0.4\columnwidth]{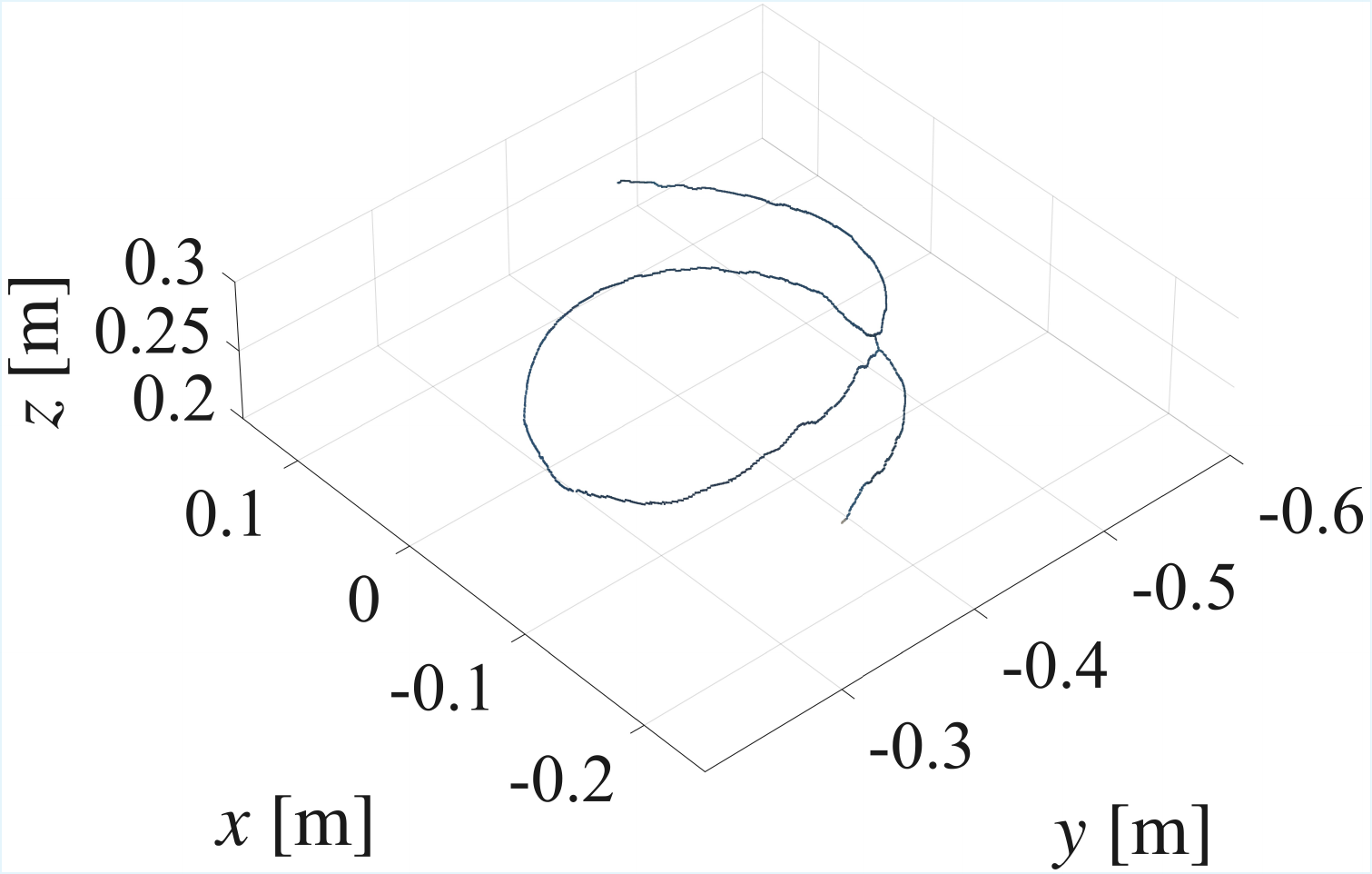}
    \includegraphics[width=0.4\columnwidth]{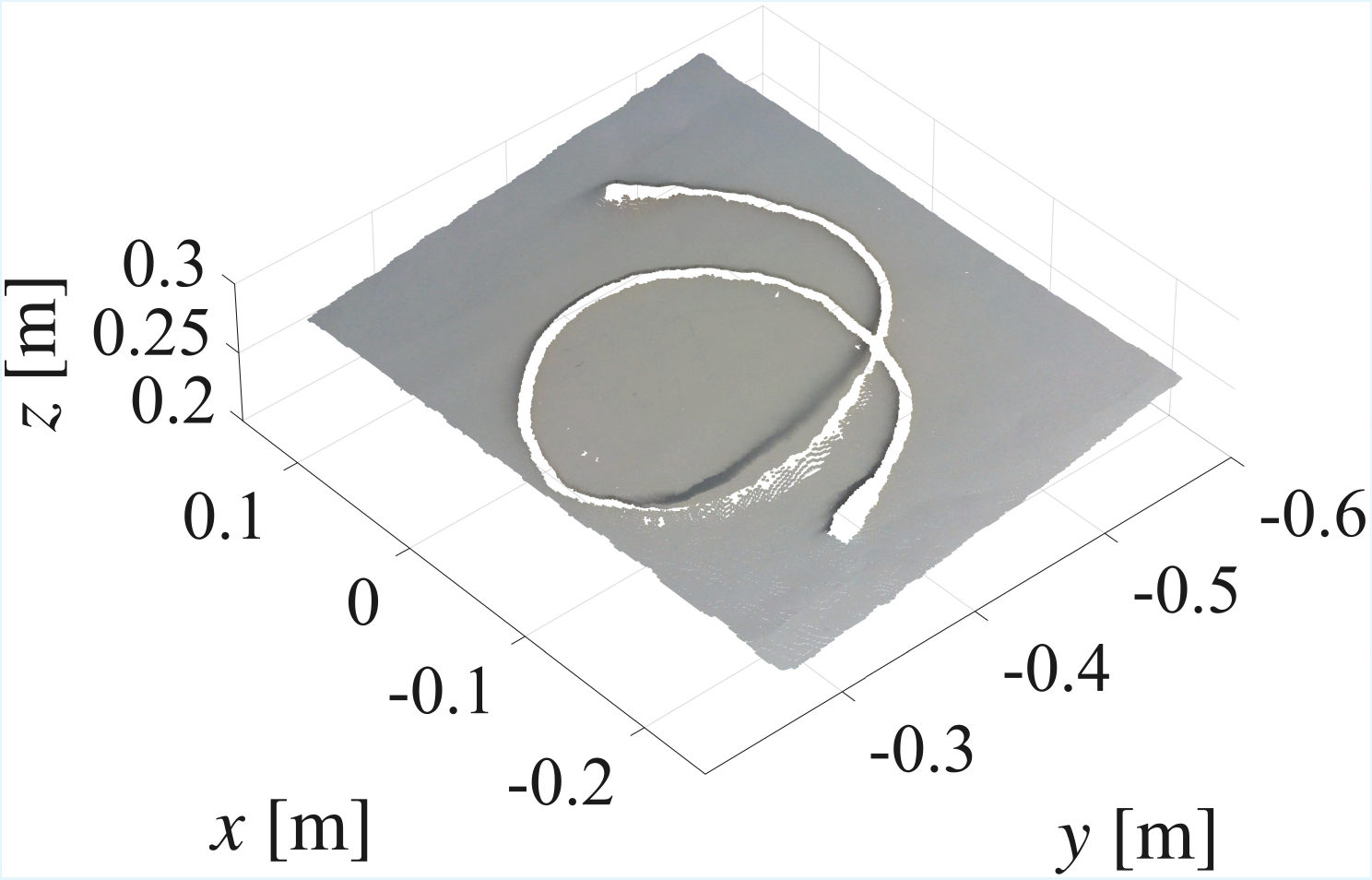}\\[0.5cm]
    \includegraphics[width=0.4\columnwidth]{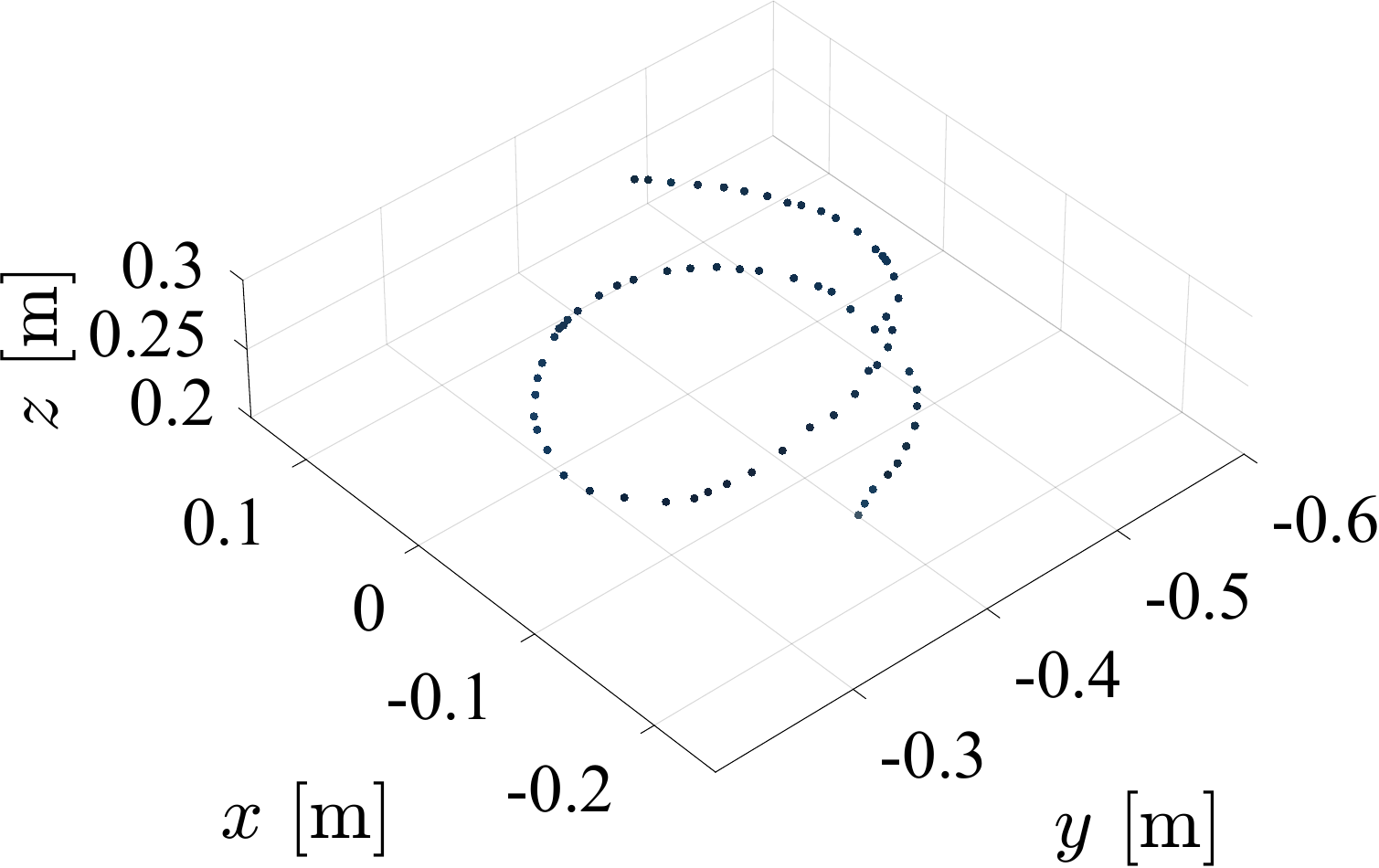}
    \includegraphics[width=0.4\columnwidth]{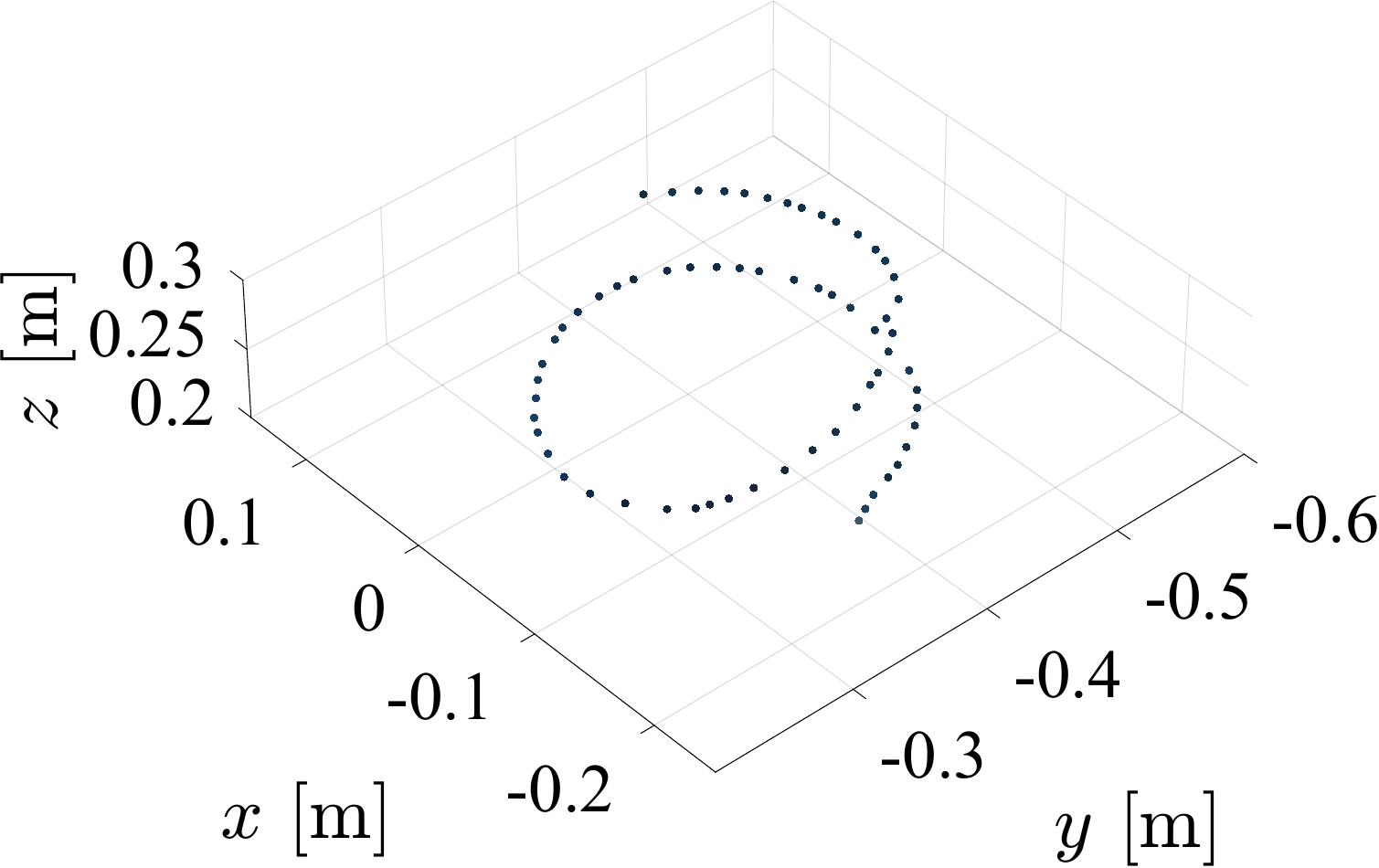}
\caption{CS1 (i): point clouds of the skeletonised cable (top-left) and of the inclined support surface (top-right). The skeleton is downsampled (bottom-left) and subsequently processed to replace the closer points with the midpoint, and then all points are projected onto the plane identified by RANSAC (bottom-right).}
\label{fig:1cable_Inc_noOcc_skeletonPC+backPC}
\end{figure}

The robot is brought into the home position to acquire the image of the scene, subsequently, the image is processed through the semantic segmentation with the prompt \textit{shelf and cables}. By exploiting the segmentation masks, it is possible to isolate the pixels of the cable and of the support surface (see Fig. \ref{fig:1cable_Inc_noOcc_image+sam2}).
The experiments in this section treat the case of a cable intersecting itself on an inclined plane (as illustrated in the video attached as supplementary material). The estimation of the plane is necessary to correctly define the end-effector orientation during tactile exploration. Fig. \ref{fig:1cable_Inc_noOcc_image+sam2} shows the results of the image processing.
Since there is a single cable, the result of the clustering algorithm is the same image obtained after the segmentation and reported at the bottom-left in Fig. \ref{fig:1cable_Inc_noOcc_image+sam2}. Both the cable image and the segmented surface are then converted into point clouds to apply the processing steps reported in Fig. \ref{fig:1cable_Inc_noOcc_skeletonPC+backPC}.

\begin{figure}[th]
    \centering
    \includegraphics[width=0.4\columnwidth]{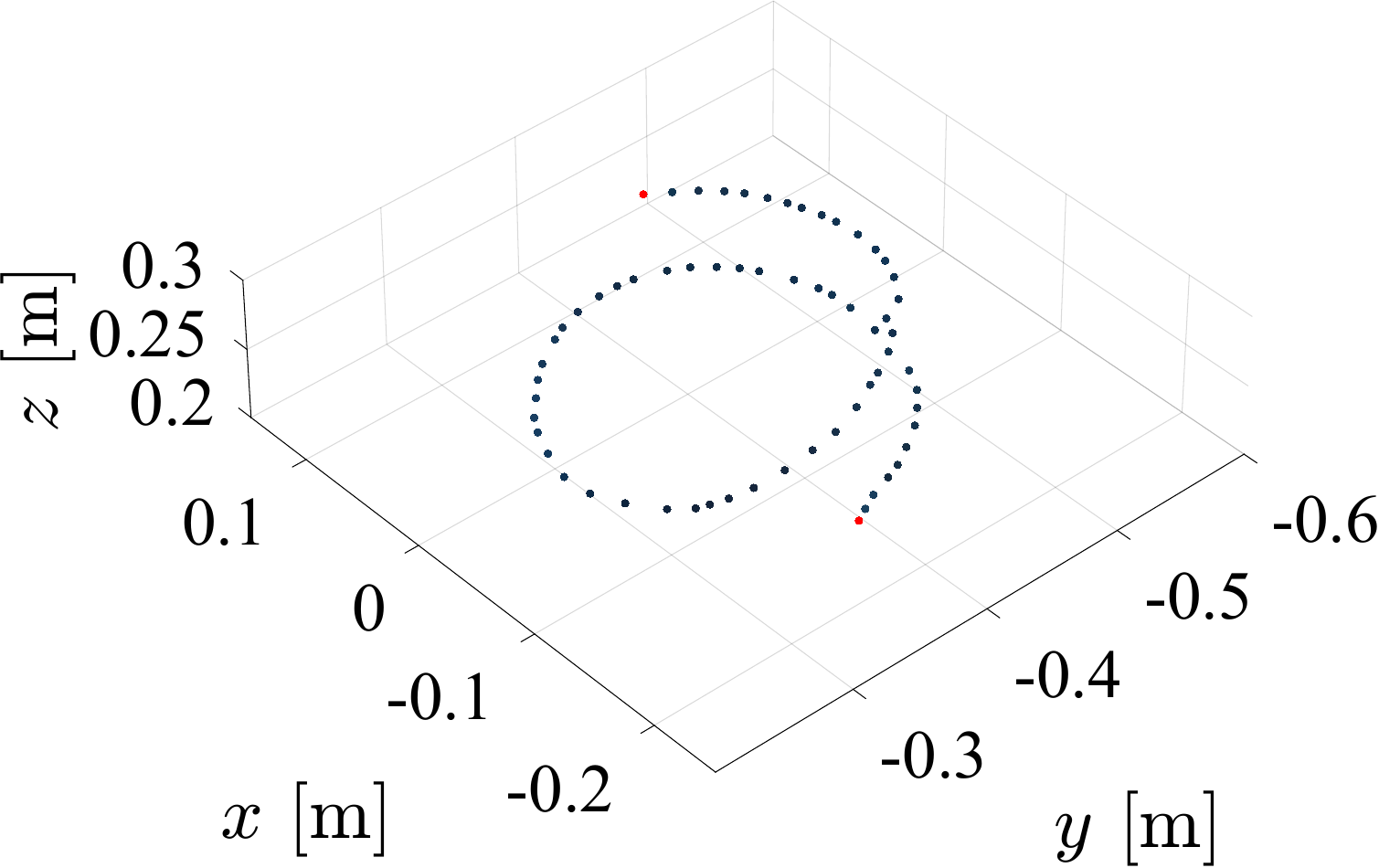}
    \includegraphics[width=0.4\columnwidth]{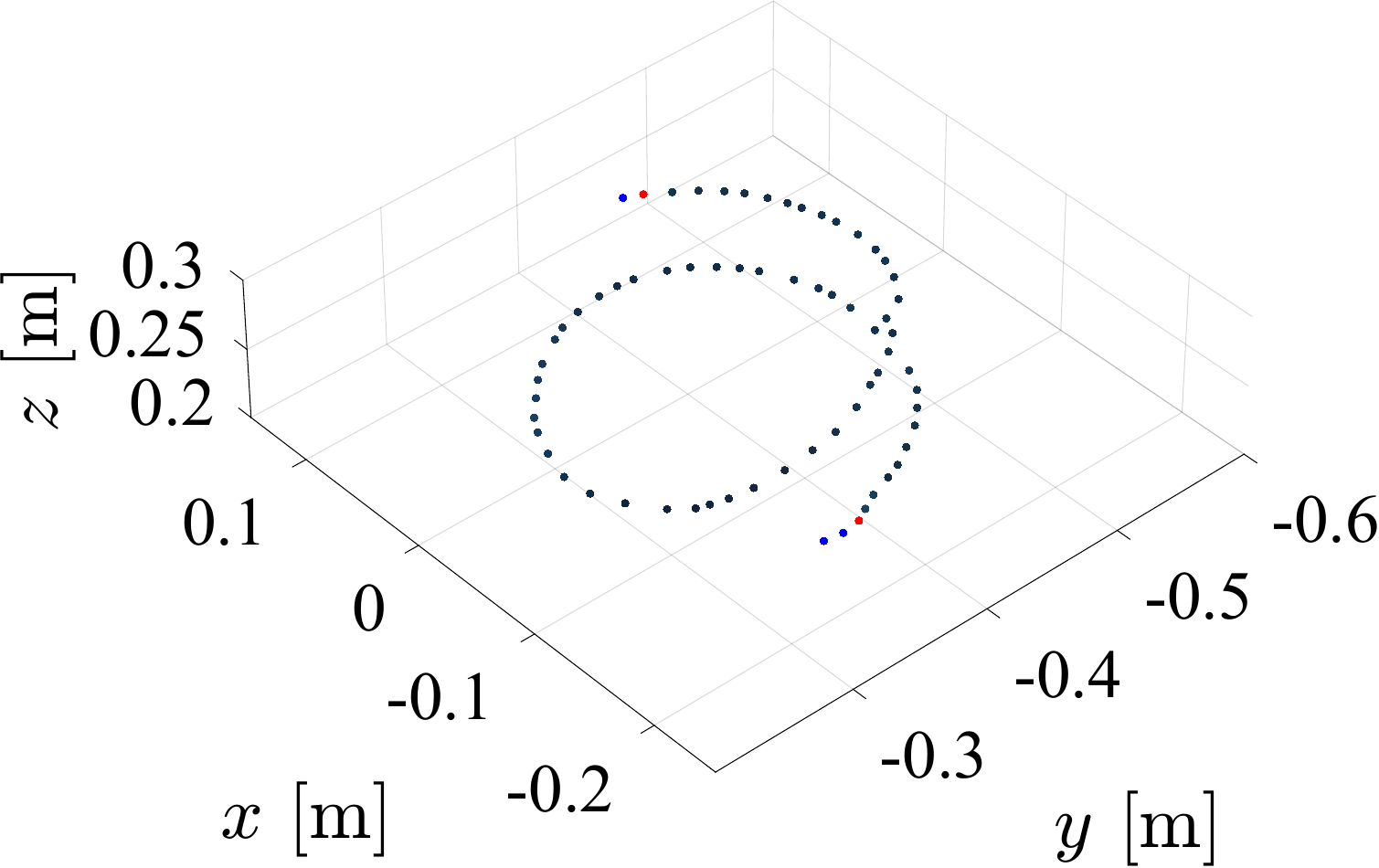}\\[0.5cm]
    \includegraphics[width=0.4\columnwidth]{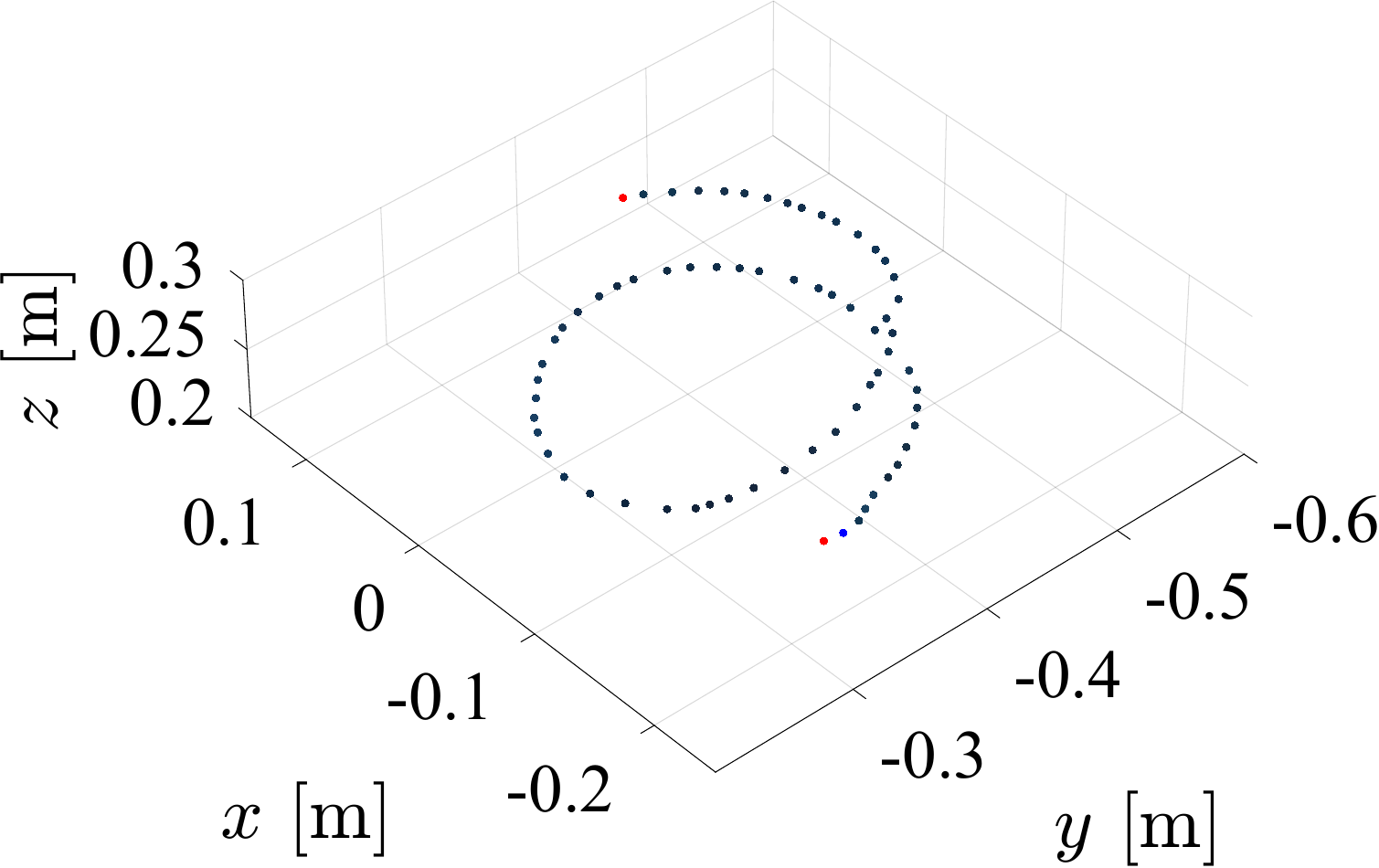}
    \includegraphics[width=0.4\columnwidth]{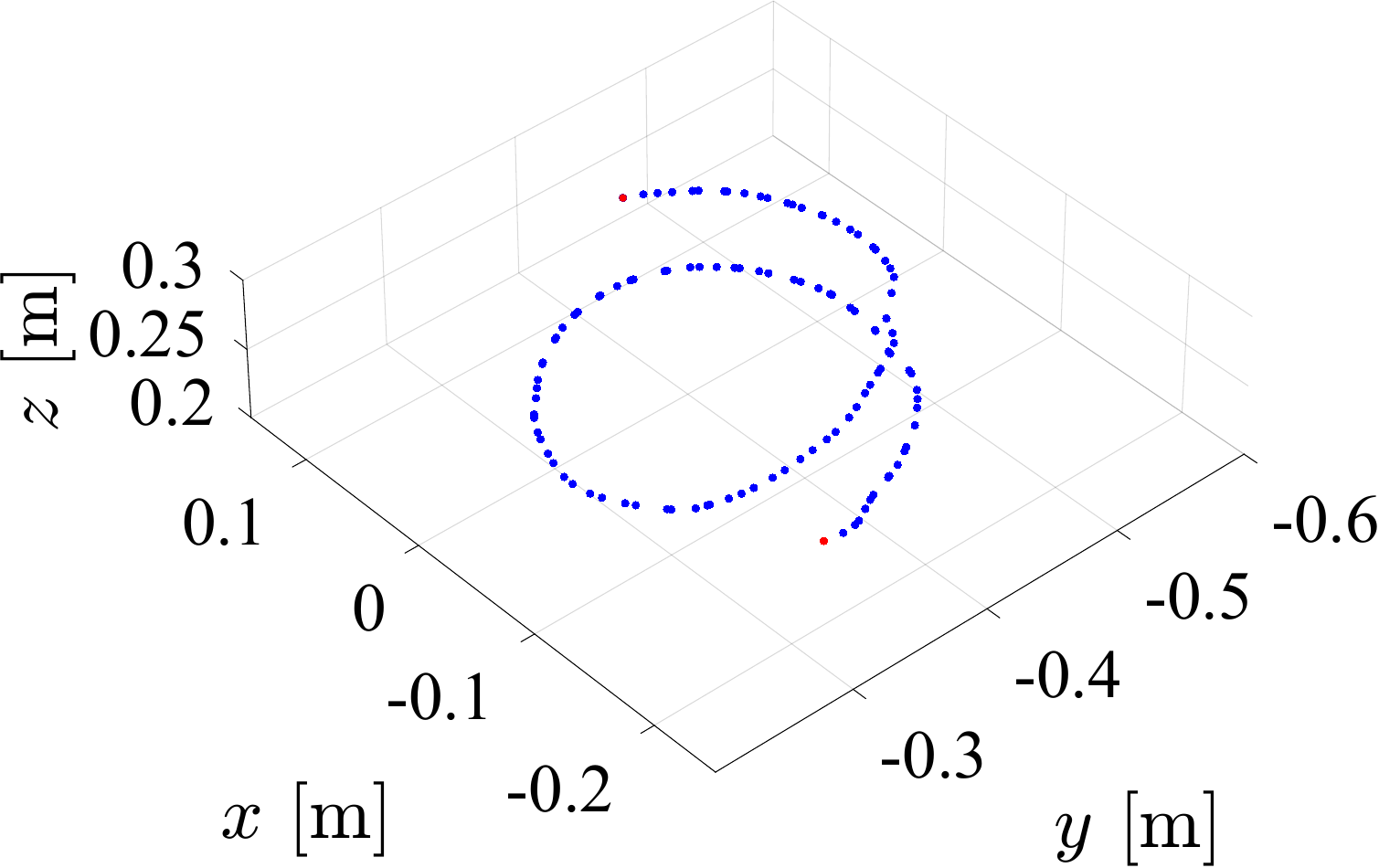}
\caption{CS1 (i): first sorted point cloud with endpoints in red (top-left), point cloud obtained by merging both visual and tactile point clouds (top-right), resulting point cloud of the new sorting step (bottom-left), interpolated point cloud (bottom-right).}
\label{fig:1cable_Inc_noOcc_endpointsPC+reconstruction}
\end{figure}

The tactile exploration phase introduces additional points belonging to the connectors of the cable, and the detection of the true endpoints, now completed successfully, is followed by the last interpolation step that produces the B-spline modeling the cable (see Fig. \ref{fig:1cable_Inc_noOcc_endpointsPC+reconstruction}).

\subsubsection*{Cable with occlusions}
\begin{figure}[th]
    \centering
    \includegraphics[width=0.4\columnwidth]{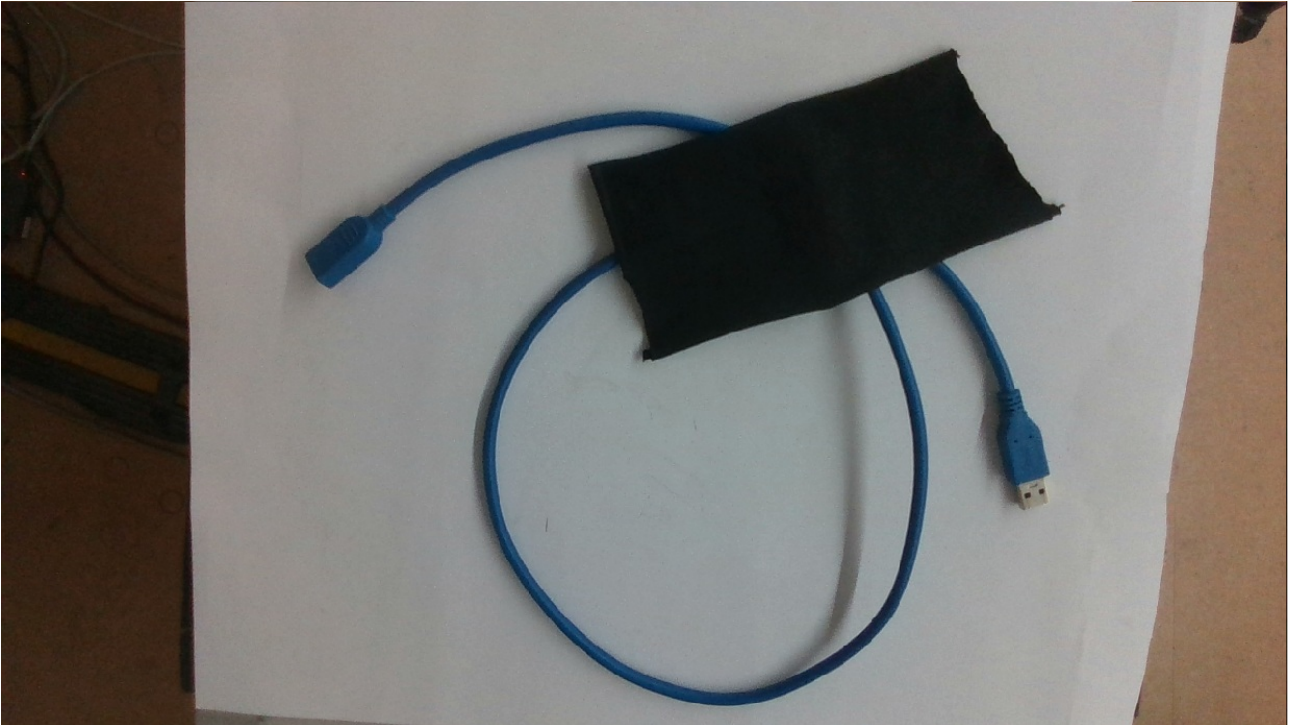}
    \includegraphics[width=0.4\columnwidth]{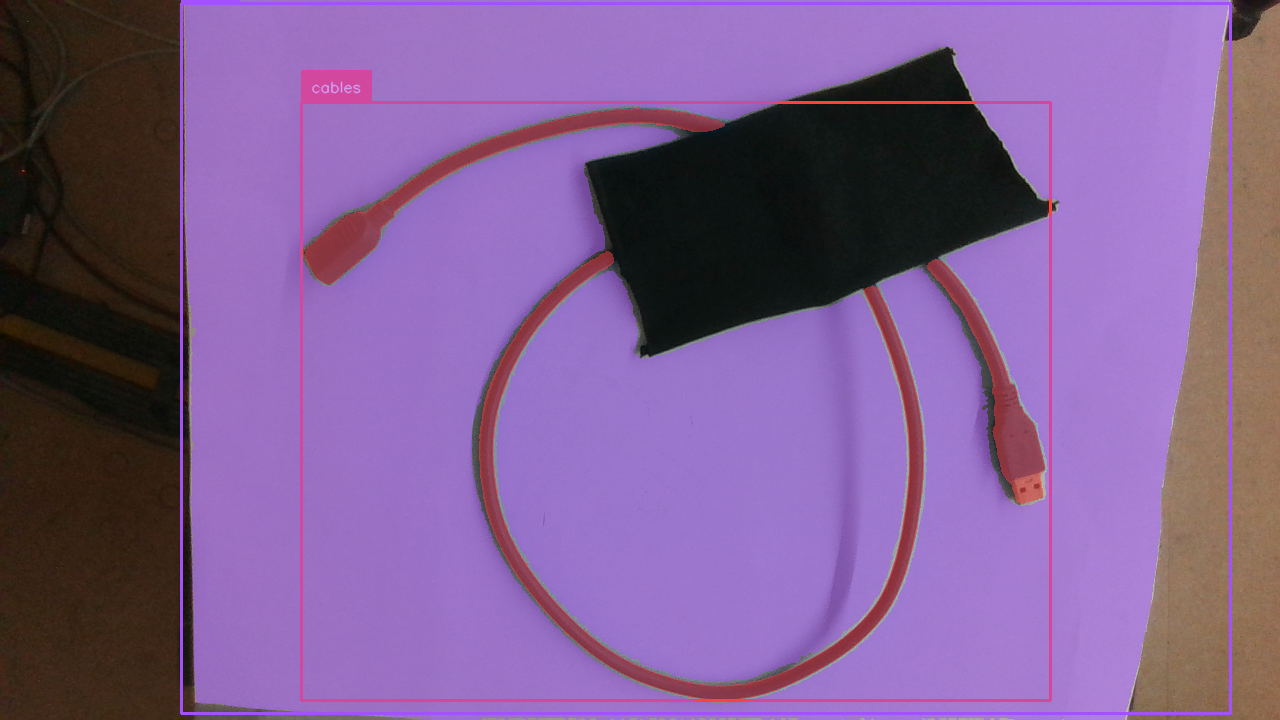}\\[0.5cm]
    \includegraphics[width=0.4\columnwidth]{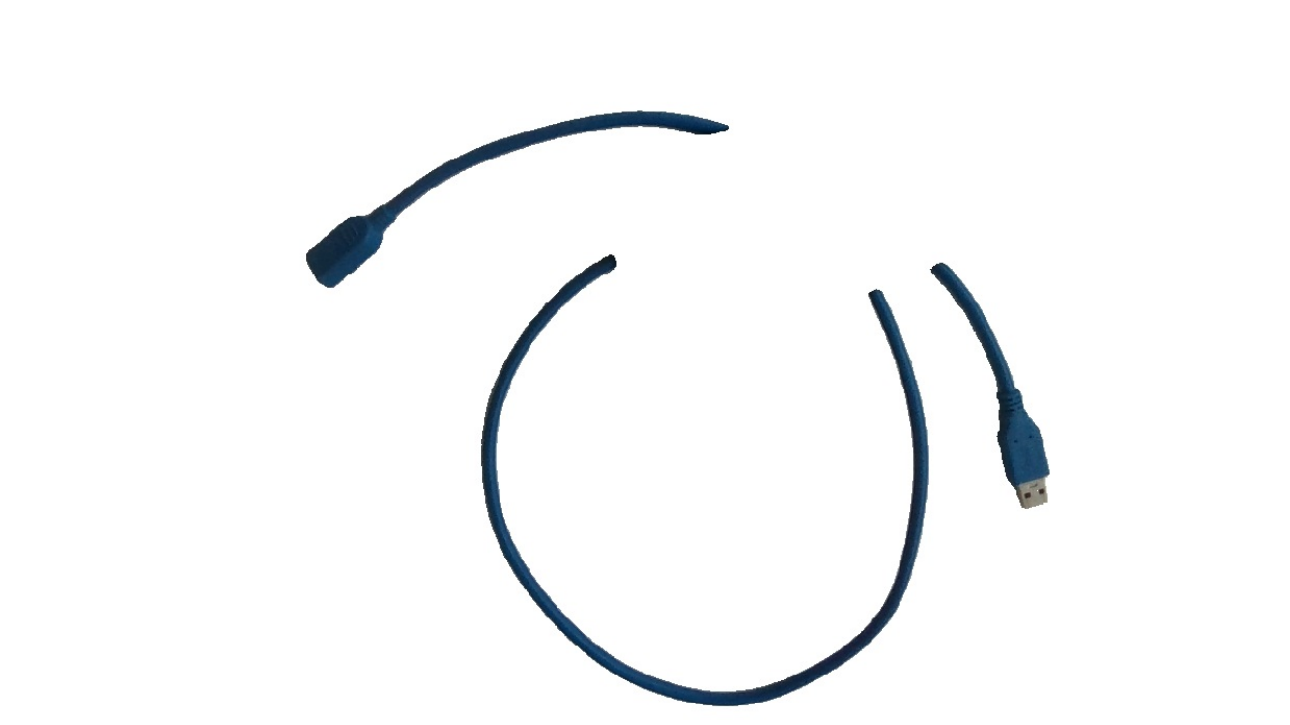}
    \includegraphics[width=0.4\columnwidth]{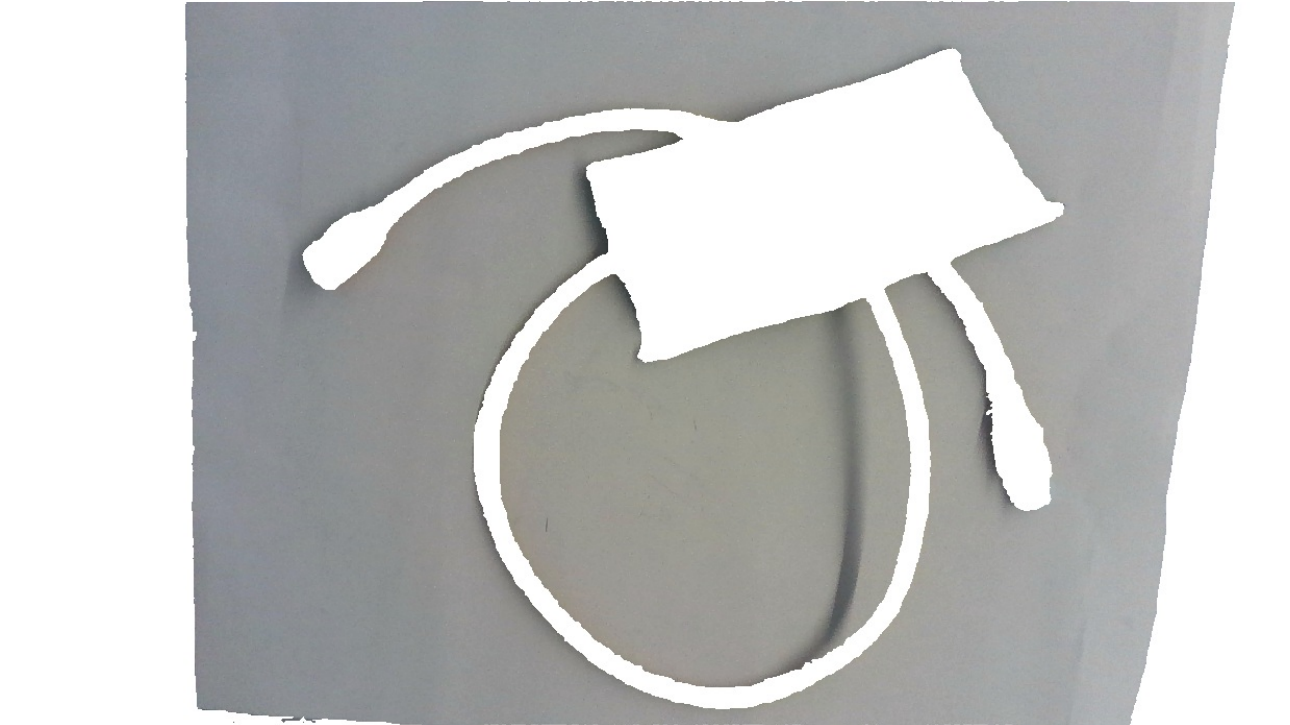}
\caption{CS1 (ii): camera RGB image (top-left), result of the semantic segmentation by Florence2/SAM2 (top-right), segmented cable (bottom-left) and support surface (bottom-right).}
\label{fig:1cable_Inc_Occ_image+sam2}
\end{figure}

\begin{figure}[th]
    \centering
    \includegraphics[width=0.4\columnwidth]{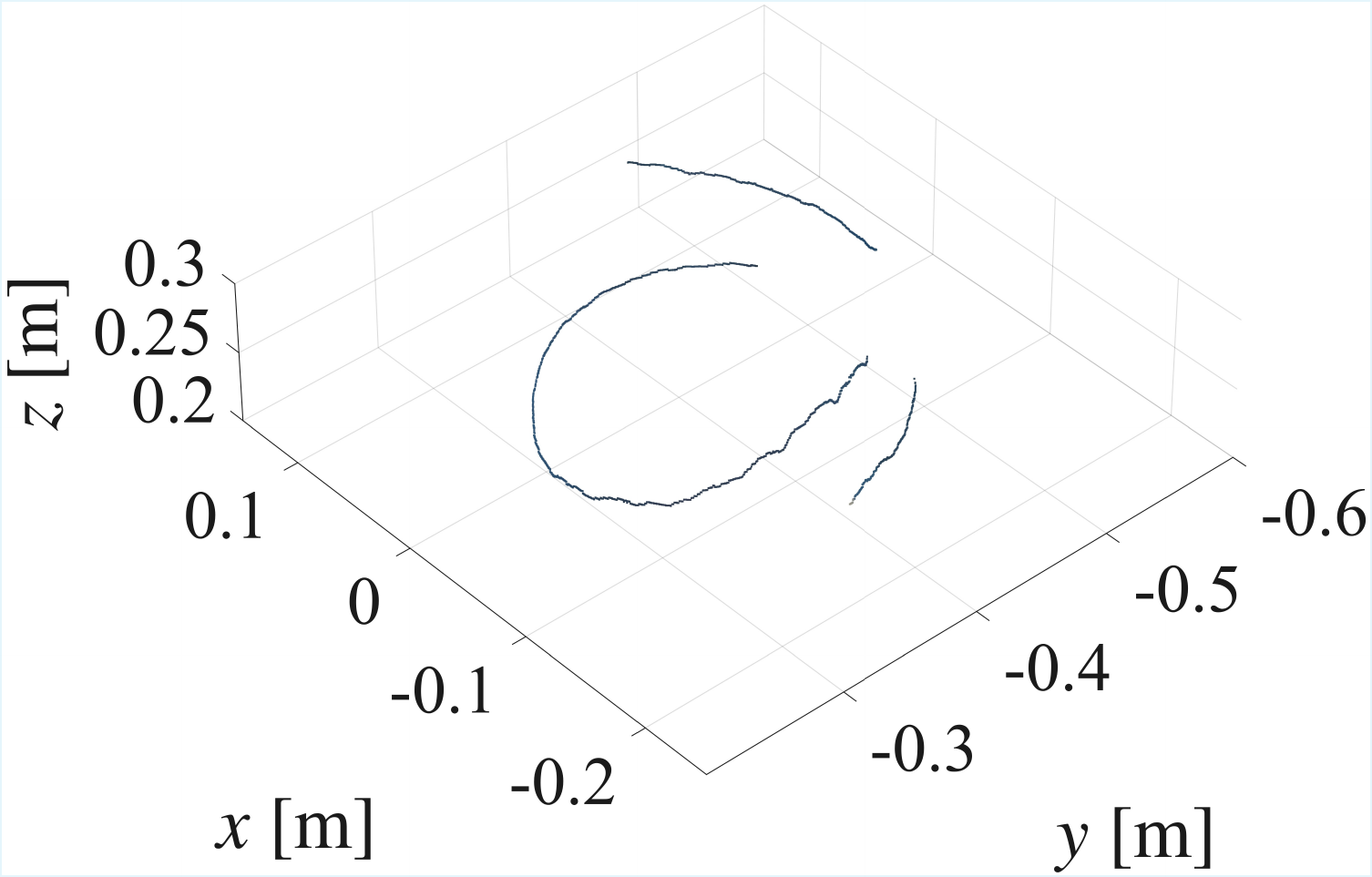}
    \includegraphics[width=0.4\columnwidth]{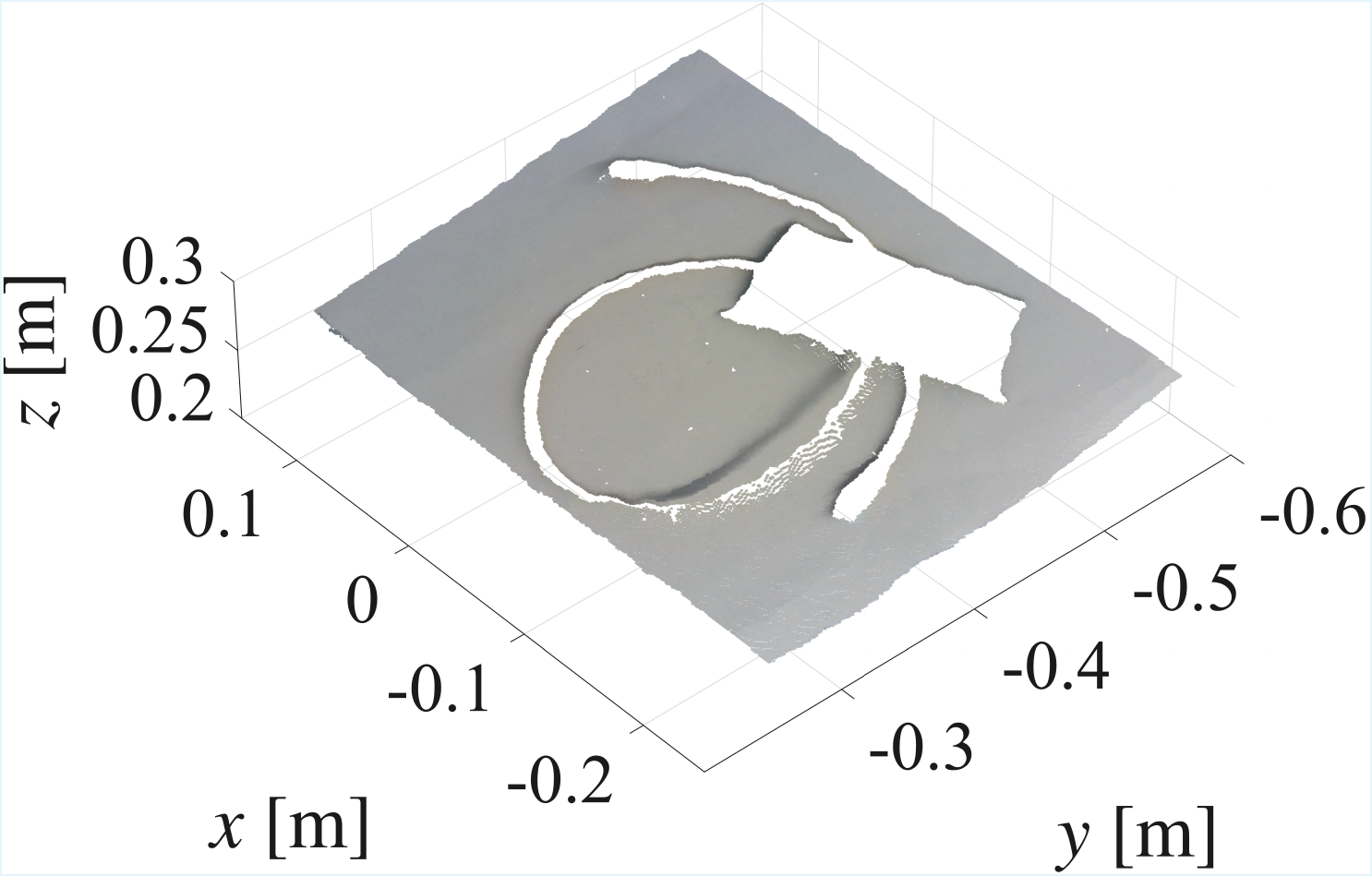}\\[0.5cm]
    \includegraphics[width=0.4\columnwidth]{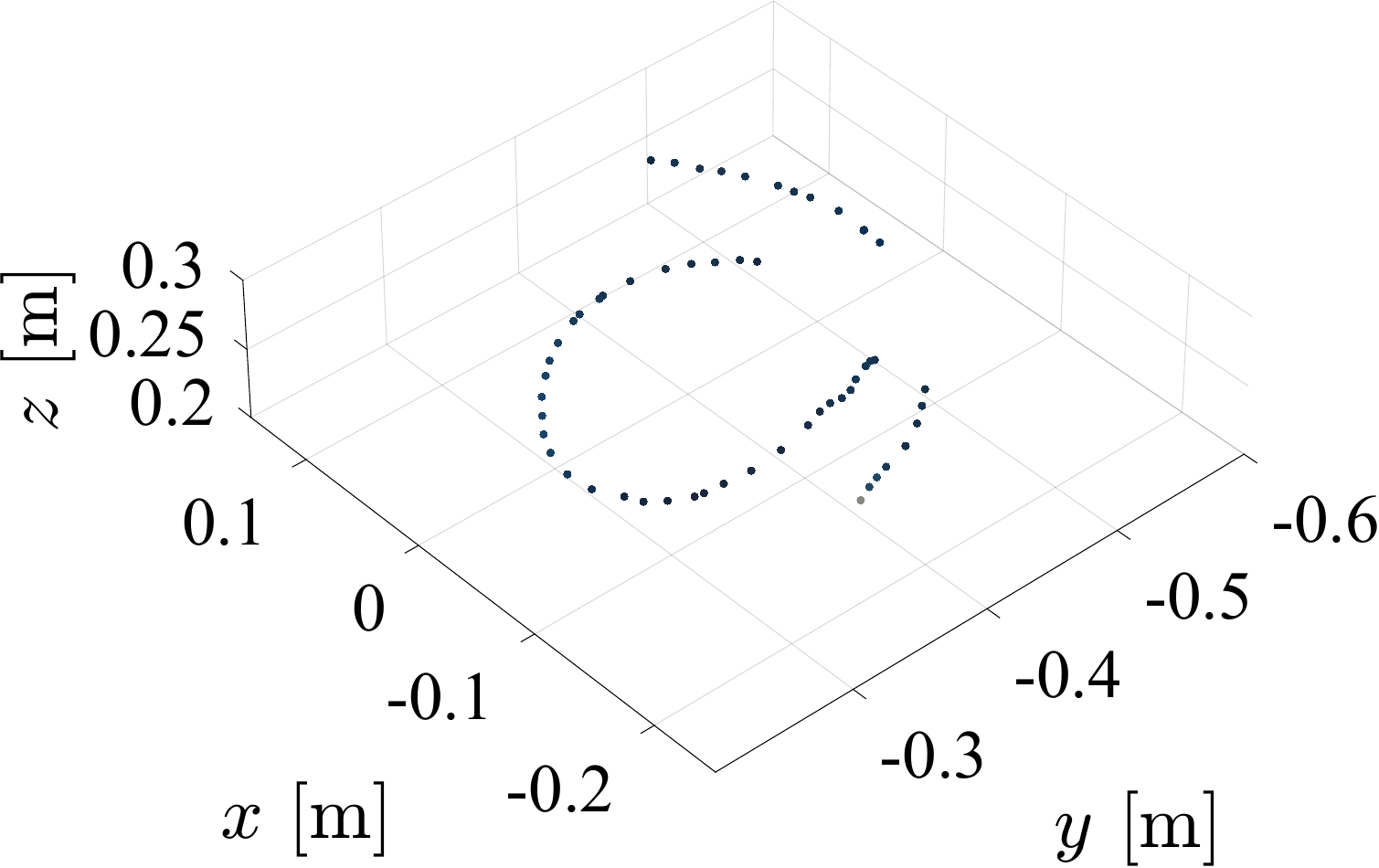}
    \includegraphics[width=0.4\columnwidth]{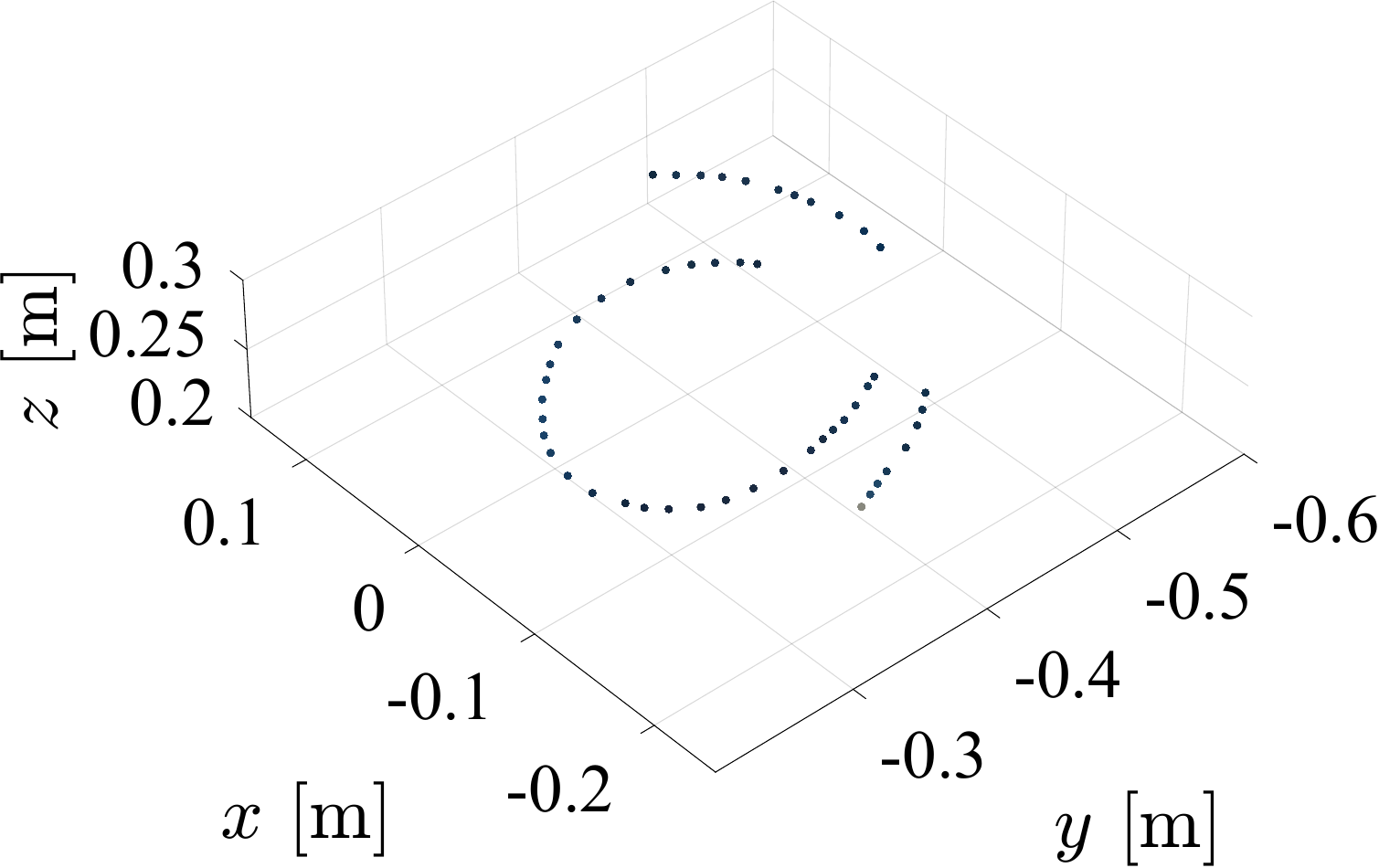}
\caption{CS1 (ii): point clouds of the skeletonised cable (top-left) and of the inclined support surface (top-right). The skeleton is downsampled (bottom-left) and subsequently processed to replace the closer points with the midpoint, and then all points are projected onto the plane identified by RANSAC (bottom-right).}
\label{fig:1cable_Inc_Occ_skeletonPC+backPC}
\end{figure}

\begin{figure}[th]
    \centering
    \includegraphics[width=0.4\columnwidth]{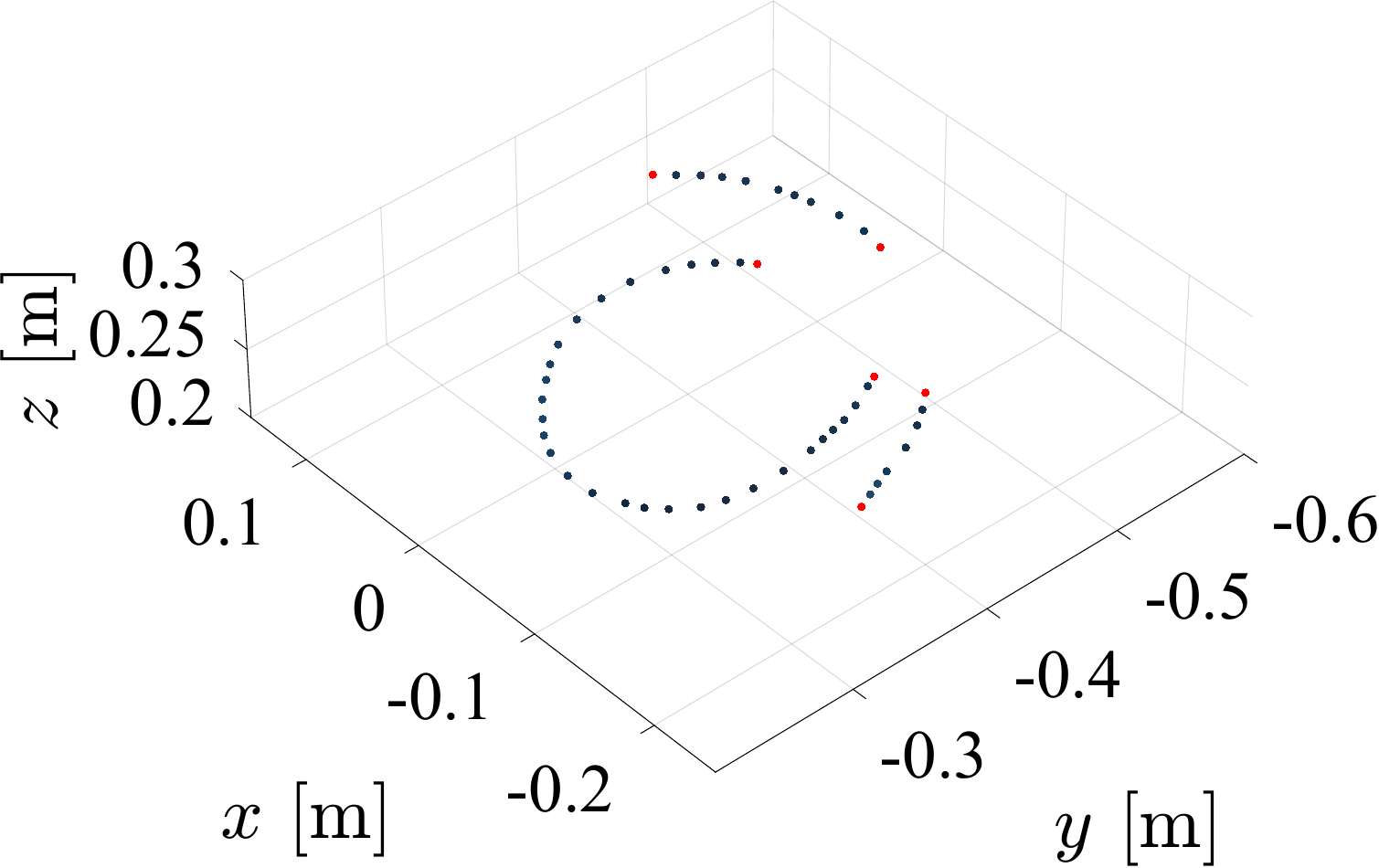}
    \includegraphics[width=0.4\columnwidth]{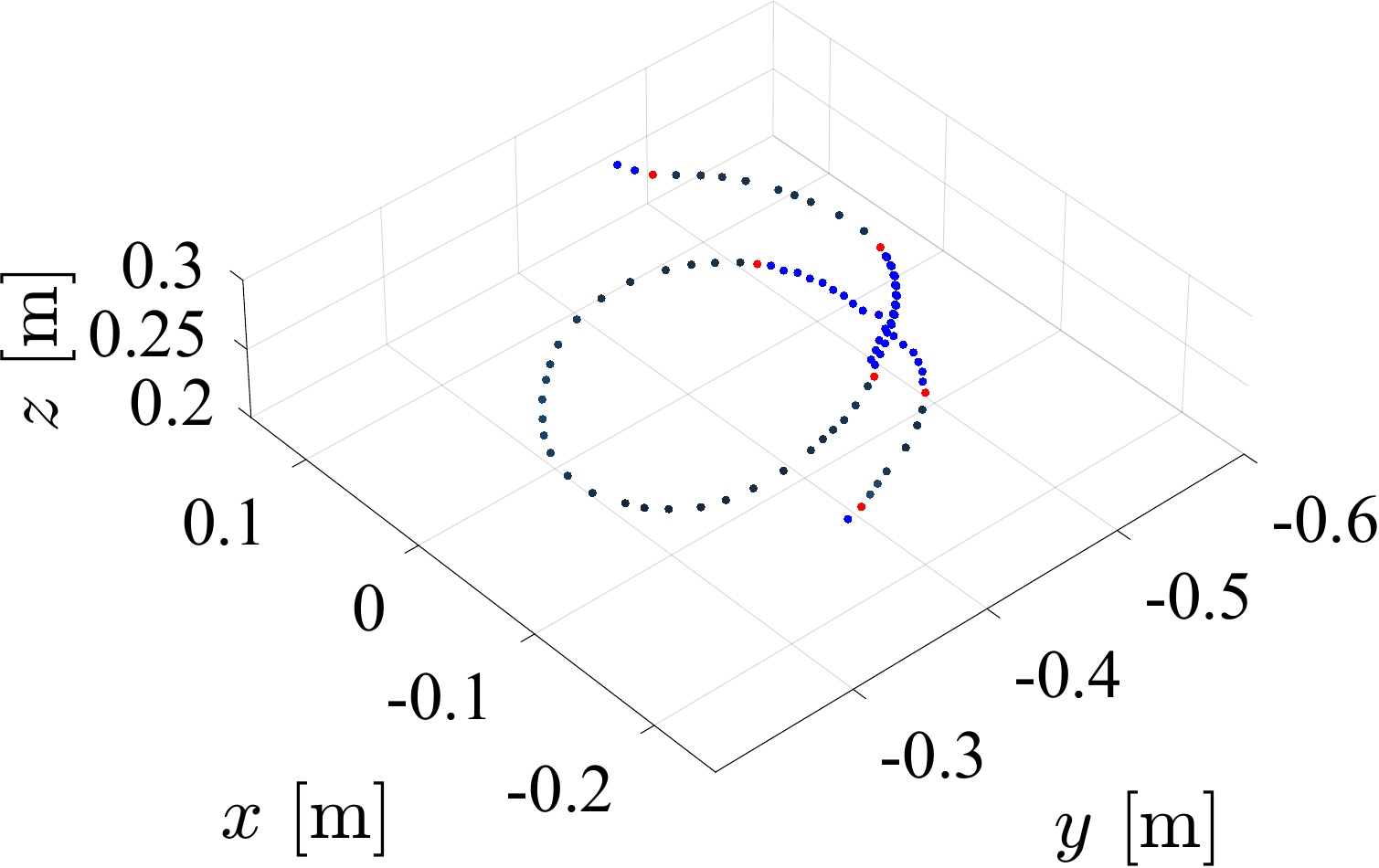}
\caption{CS1 (ii): first sorted point cloud with endpoints in red (left), point cloud obtained by merging both visual and tactile point clouds (right).}
\label{fig:1cable_Inc_Occ_critic_tactile_exploration}
\end{figure}

In this experiment, as shown in Fig. \ref{fig:1cable_Inc_Occ_image+sam2}, the cable is in the same configuration as the previous one, but a cable part is now occluded by an object. This scenario emphasizes a critical aspect of the proposed approach. In particular, the occluding object is placed over a cable intersection, and this could introduce some problems during tactile exploration as discussed below. Fig. \ref{fig:1cable_Inc_Occ_image+sam2} reports the results of the image processing.
Fig. \ref{fig:1cable_Inc_Occ_skeletonPC+backPC} illustrates the related point clouds, where empty parts correspond to cable segments under the occluding object.
The endpoint detection on the visual point cloud is correctly performed as shown in Fig. \ref{fig:1cable_Inc_Occ_critic_tactile_exploration} (left). The result of the tactile exploration (right) is characterised by additional points along the cable portions that were already explored. This behaviour is clearly shown in the attached video, and it is due to the cable intersection. In particular, if cable segments intersect with a small intersection angle (less than $45^\circ$), the tactile exploration could return a more dense tactile point cloud. In general, the rate of information is good, but the real problem is the search for the true endpoints: the sorting algorithm could find more than two endpoints although the point cloud is completely reconstructed. This is a limitation of the proposed approach because the sorting algorithm performs well if the point cloud is not dense; for this reason, downsampling and post-processing of closer points are necessary before sorting the point cloud.

\begin{figure}[th]
    \centering
    \includegraphics[width=0.4\columnwidth]{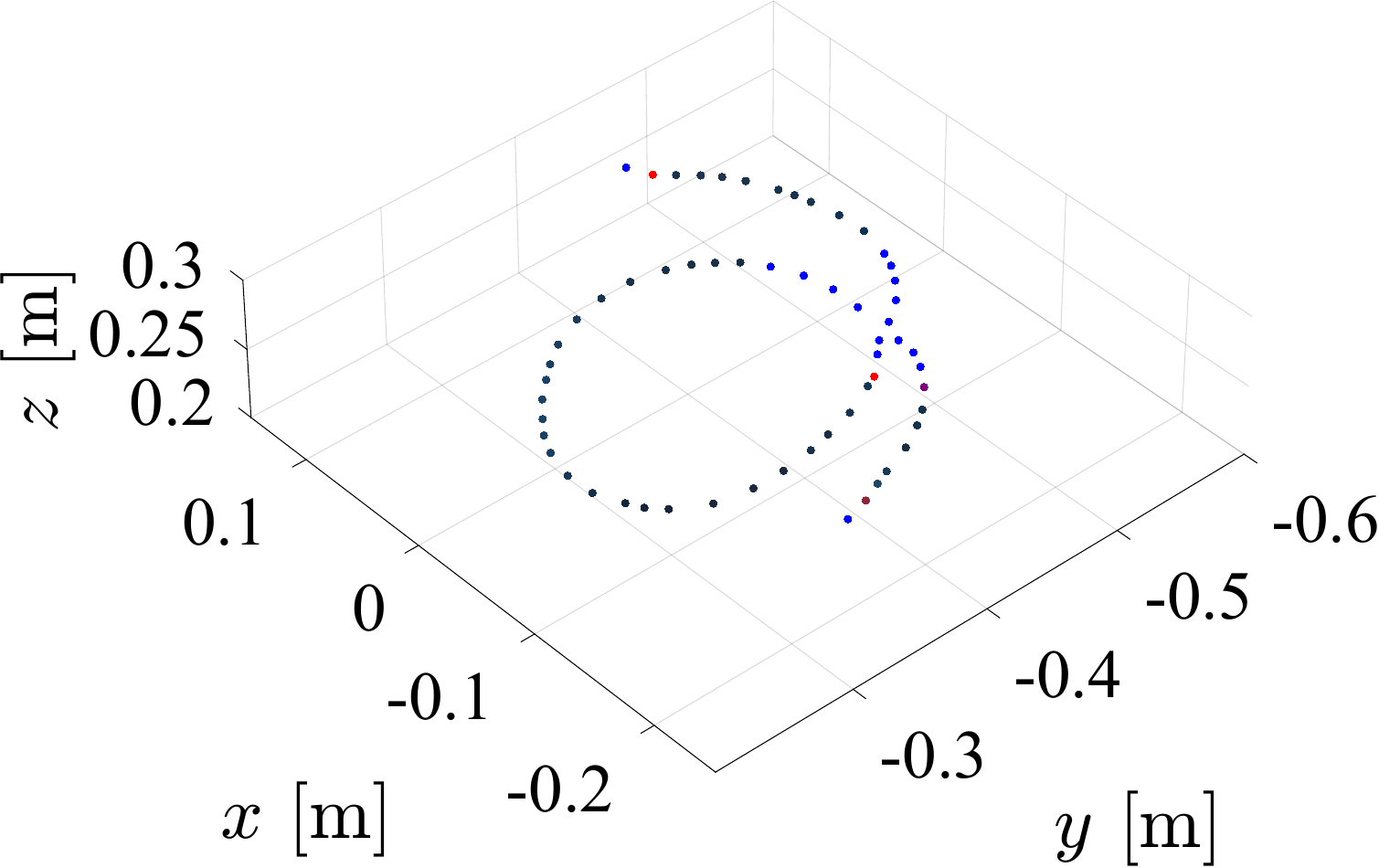}
    \includegraphics[width=0.4\columnwidth]{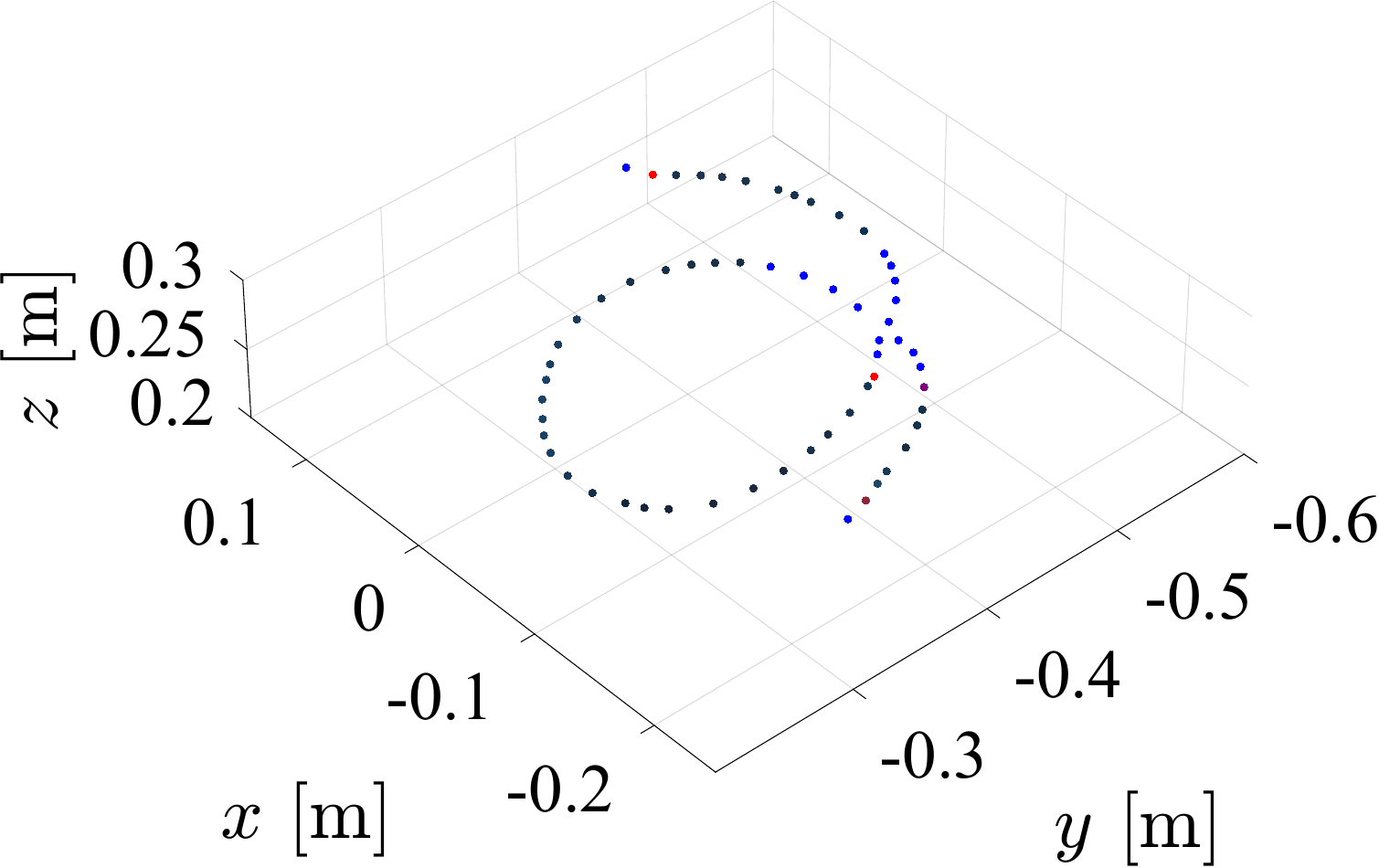}\\[0.5cm]
    \includegraphics[width=0.4\columnwidth]{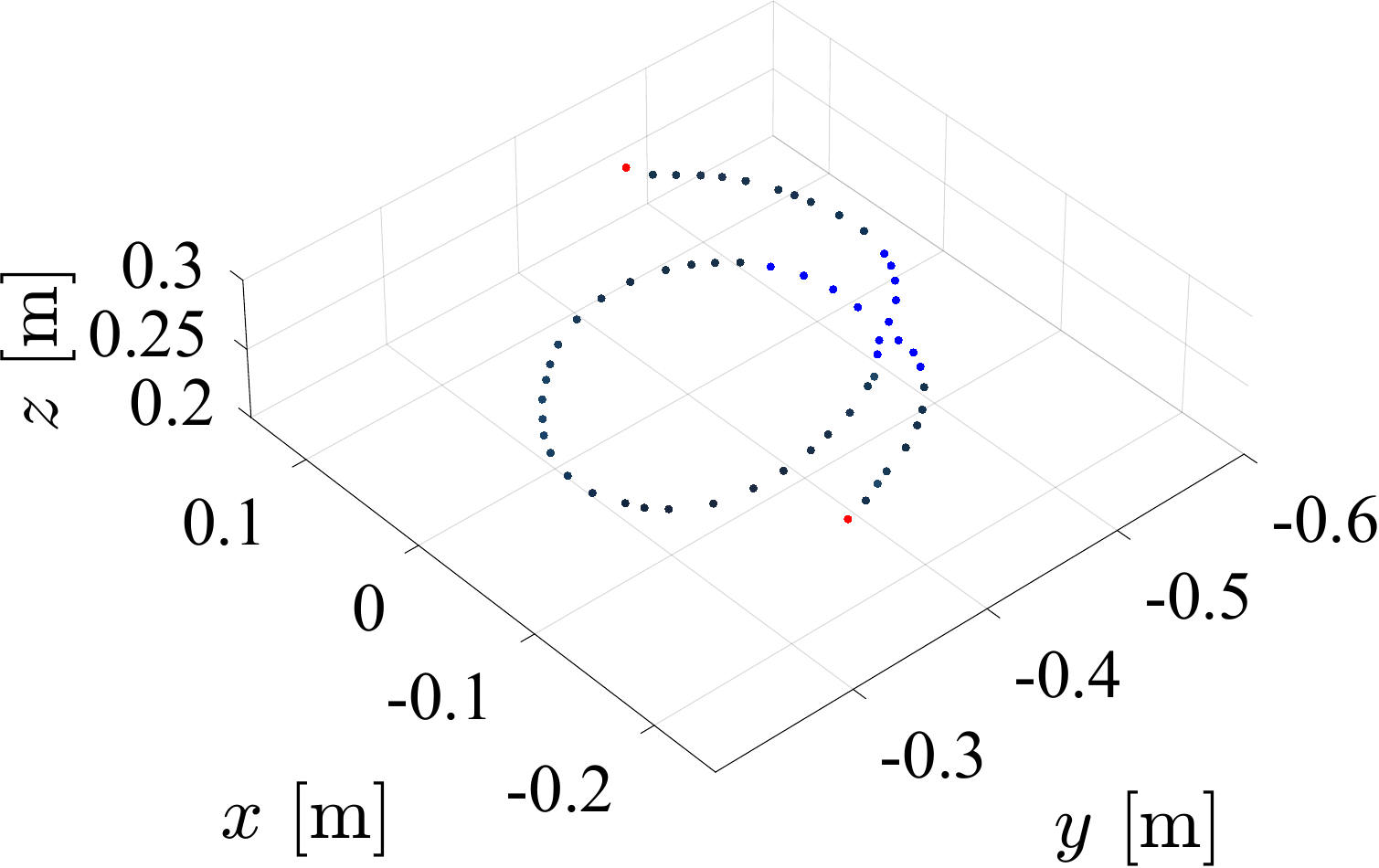}
    \includegraphics[width=0.4\columnwidth]{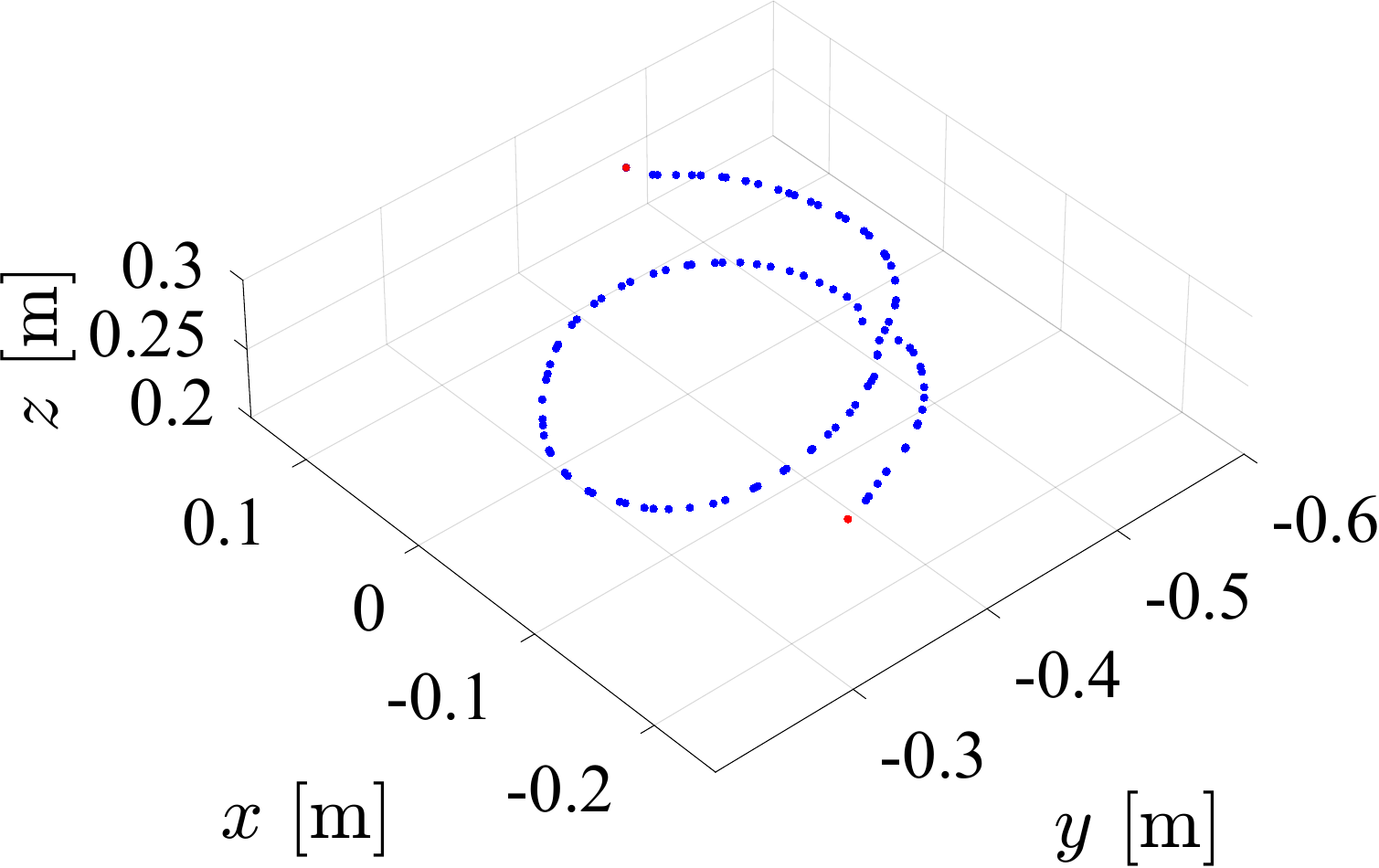}
\caption{CS1 (ii): resulting point cloud after the second downsampling step of the reconstructed point cloud (top-left), point cloud obtained by the second post-processing step of the reconstructed point cloud (top-right), resulting point cloud of the sorting step (bottom-left), interpolated point cloud (bottom-right).}
\label{fig:1cable_Inc_Occ_critic_endpointsPC+reconstruction}
\end{figure}

To mitigate this issue, the reconstructed point cloud needs to be downsampled again and eventually post-processed before sorting the point cloud and obtaining the real endpoints. The results of the second step of downsampling and post-processing, in addition to the sorting and the interpolation, are depicted in Fig. \ref{fig:1cable_Inc_Occ_critic_endpointsPC+reconstruction}. In this case, the post-processing step following the downsampling does not produce any changes in the point cloud because the points are sufficiently spaced. Finally, both the search for the true endpoints and the interpolation are completed.

\subsection{Two cables on a horizontal plane}

In the second case study (CS2) the robot has to reconstruct two cables of different colours on a horizontal plane. To underline the importance of the clustering algorithm, and for brevity, this section does not report the dense point clouds of the skeletonized cable and support surface. 

\begin{figure}[th]
    \centering
    \includegraphics[width=0.4\columnwidth]{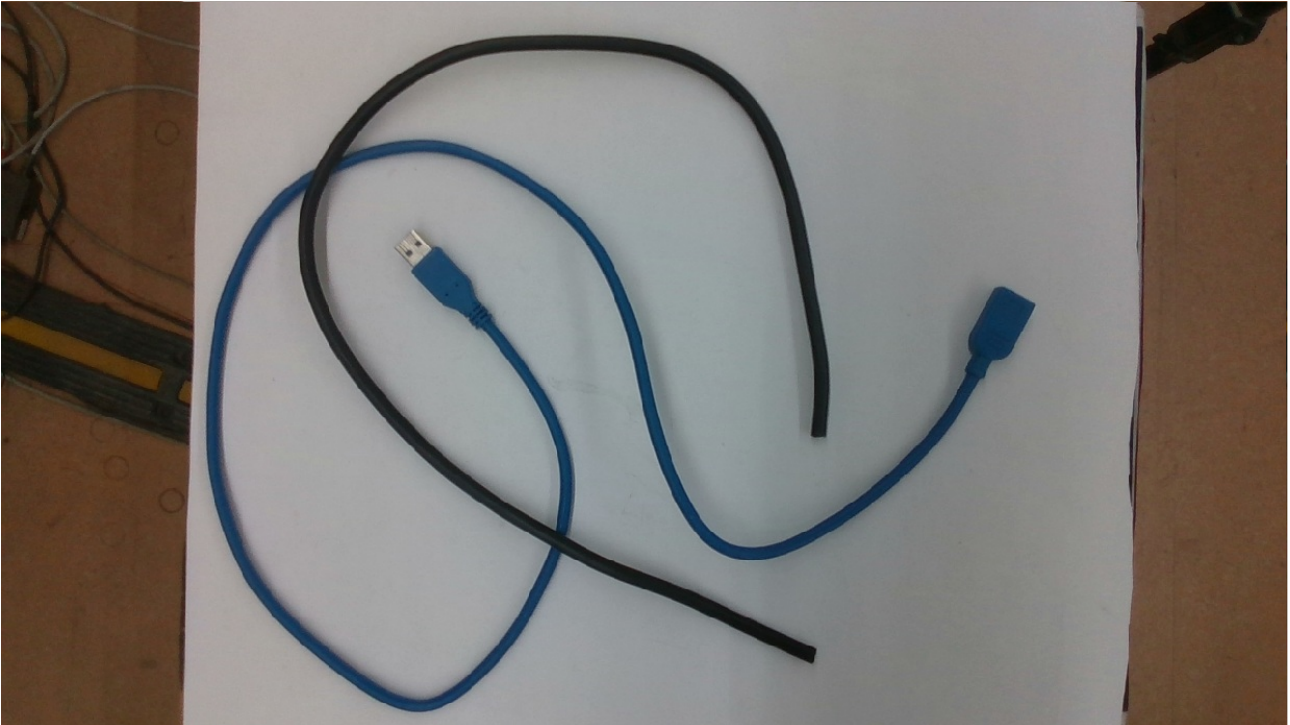}
    \includegraphics[width=0.4\columnwidth]{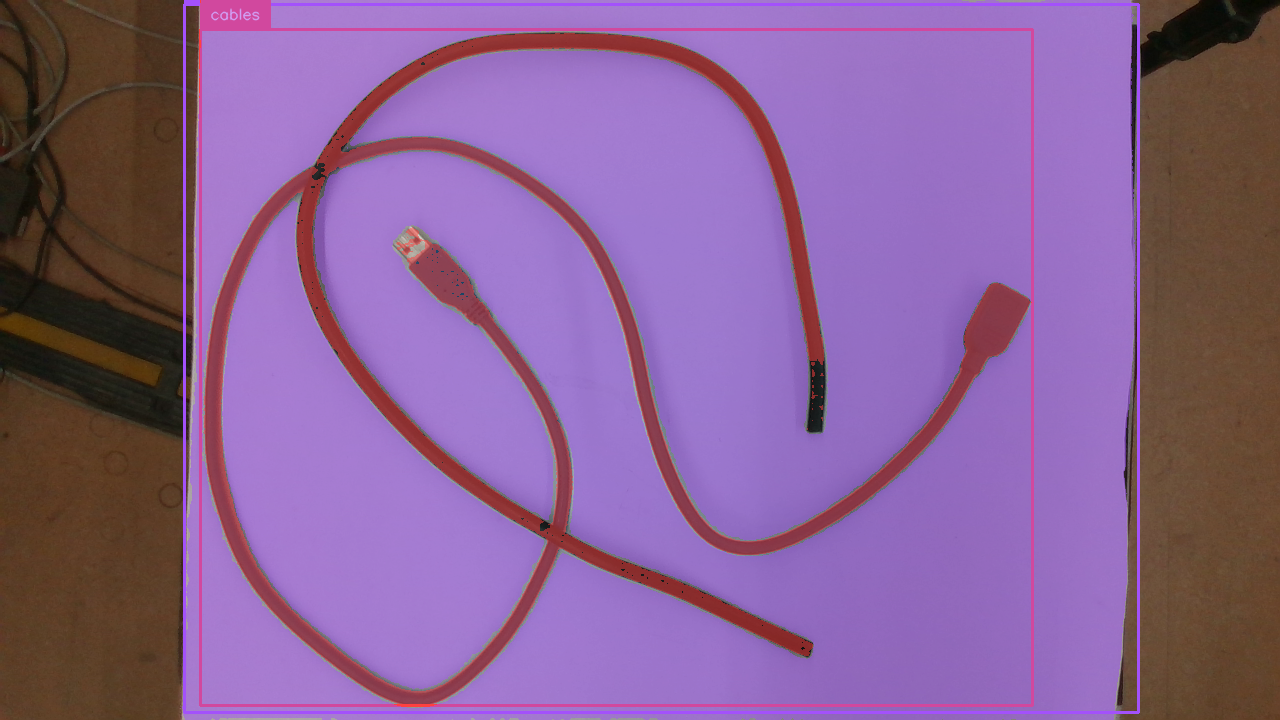}\\[0.5cm]
    \includegraphics[width=0.4\columnwidth]{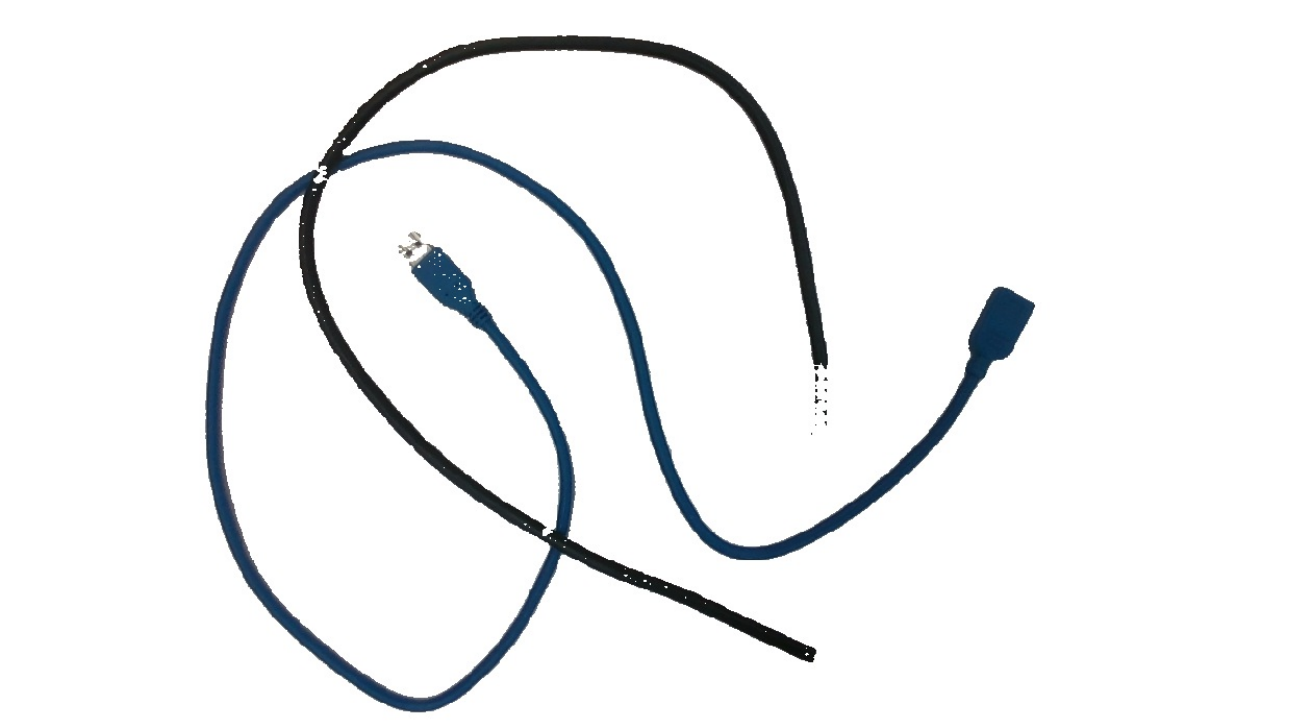}
    \includegraphics[width=0.4\columnwidth]{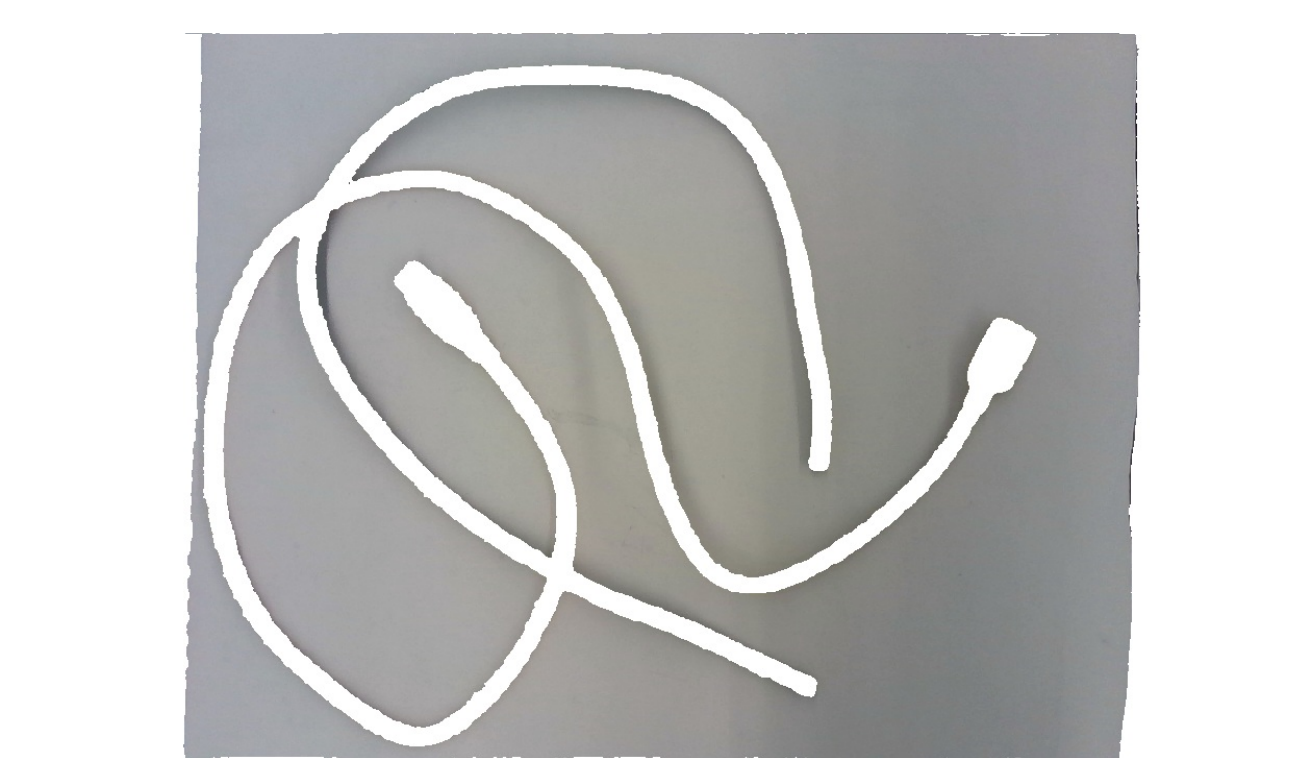}
\caption{CS2 (i): camera RGB image (top-left), result of the semantic segmentation by Florence2/SAM2 (top-right), segmented cables (bottom-left) and support surface (bottom-right).}
\label{fig:2cable_image+sam2}
\end{figure}

\begin{figure}[th]
    \centering
    \includegraphics[width=0.4\columnwidth]{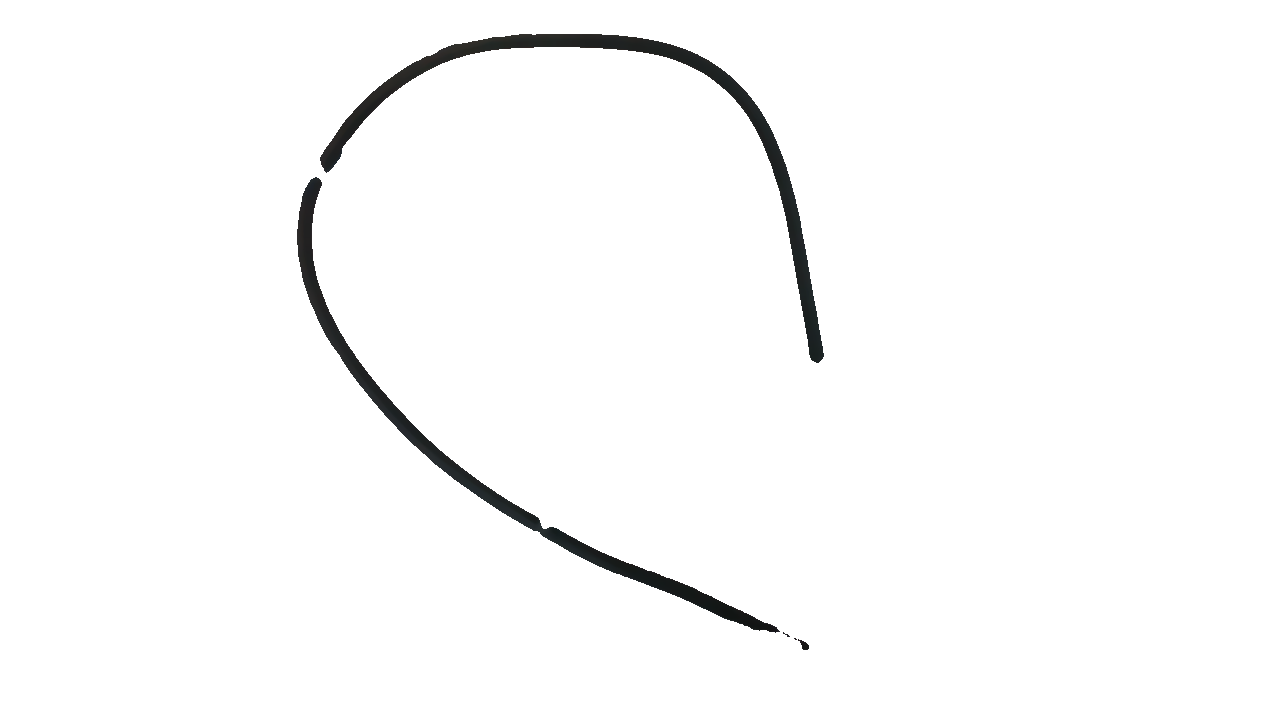}
    \includegraphics[width=0.4\columnwidth]{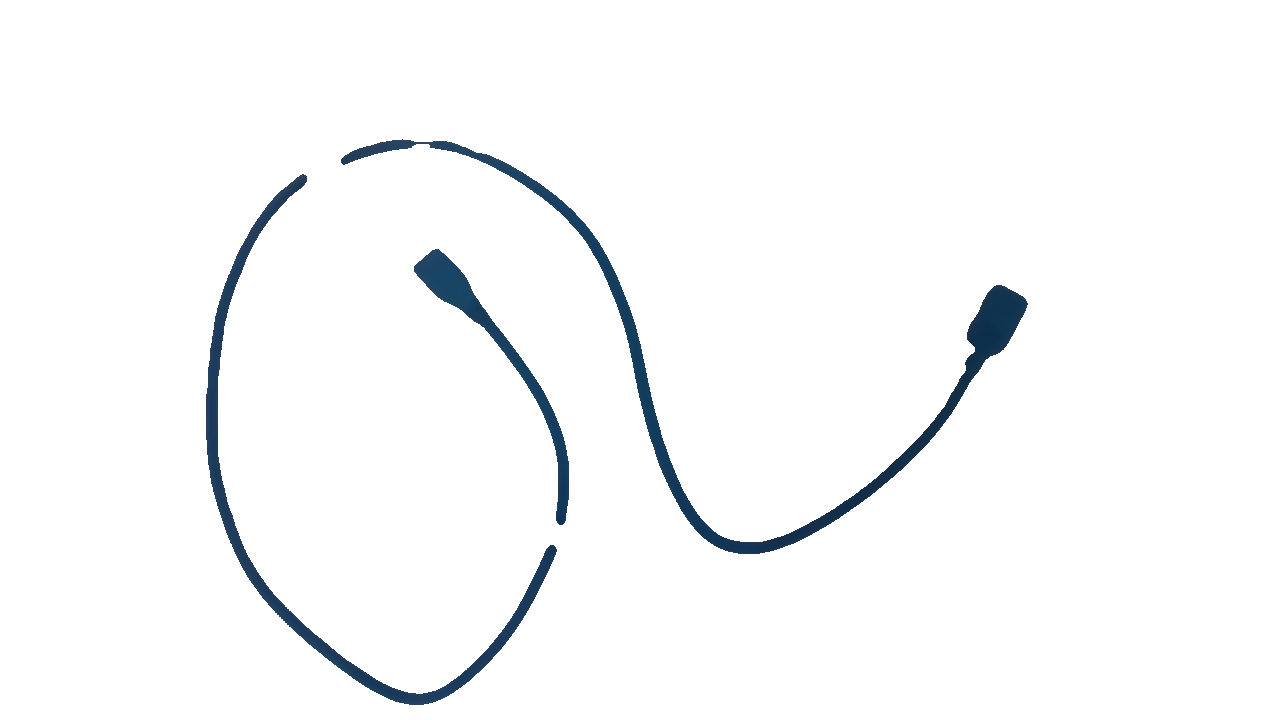}\\[0.5cm]
    \includegraphics[width=0.4\columnwidth]{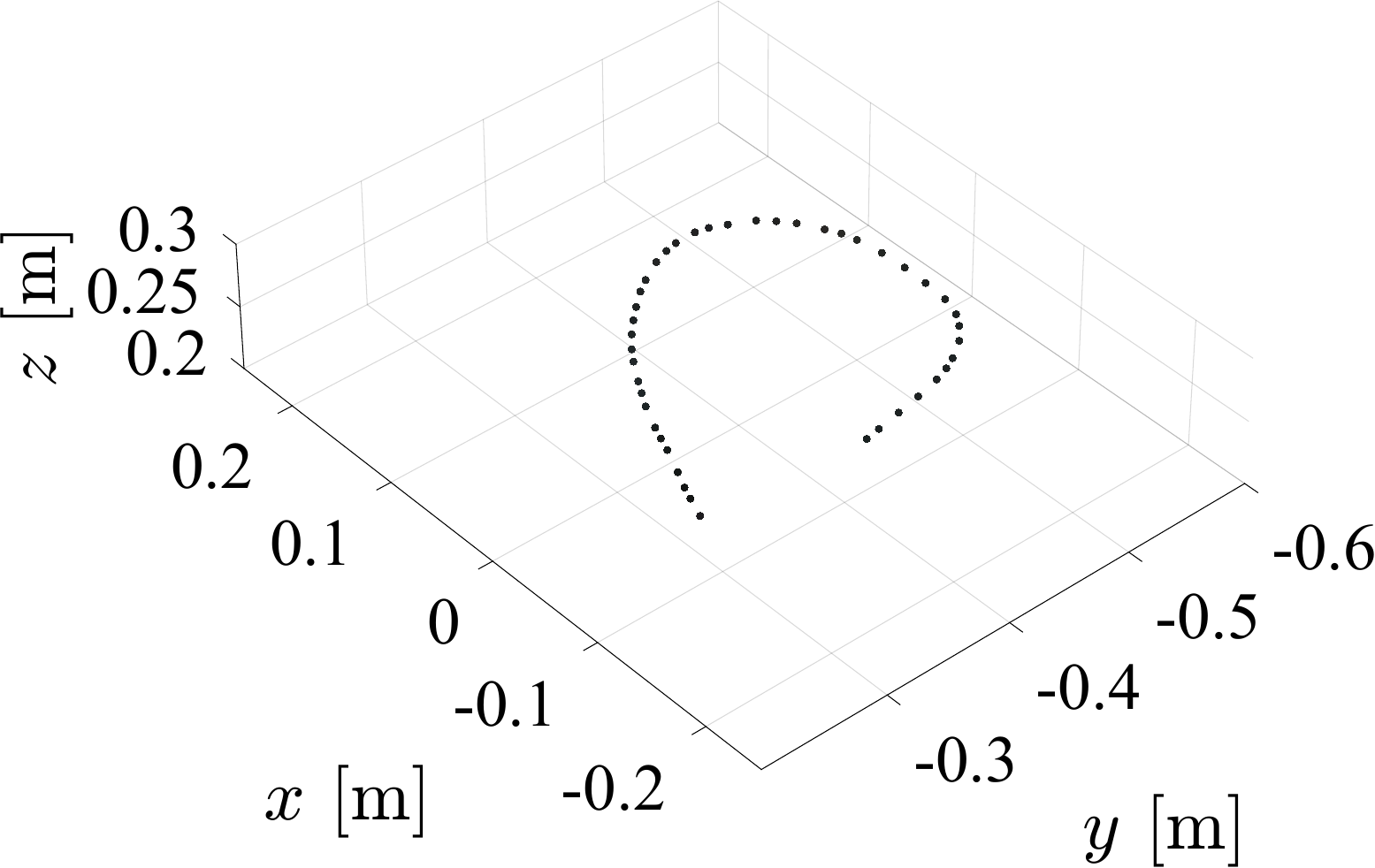}
    \includegraphics[width=0.4\columnwidth]{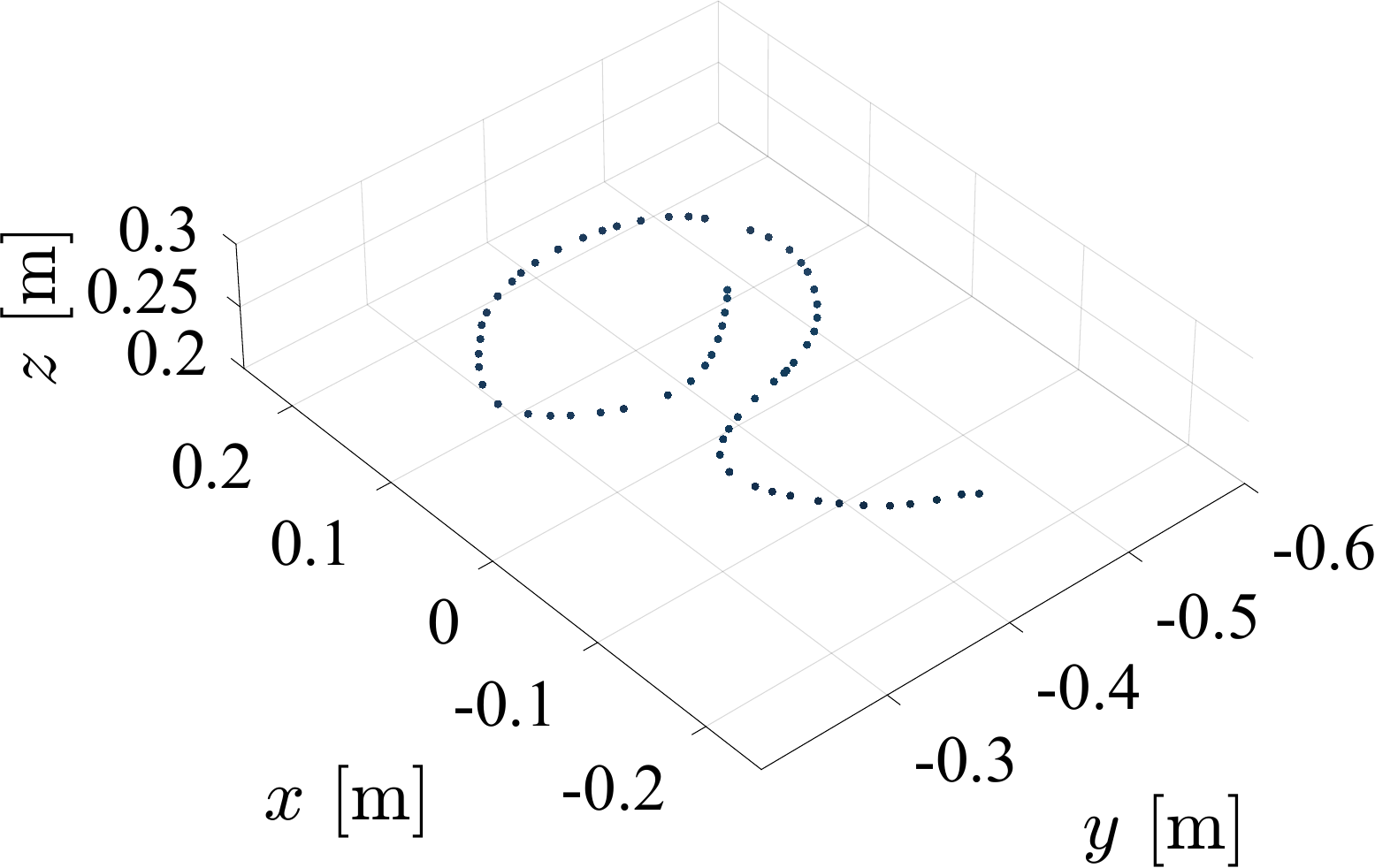}
\caption{CS2 (i): cluster of the black cable (top-left), cluster of the blue cable (top-right), post-processed point cloud of the black cable (bottom-left) and post-processed point cloud of the blue cable (bottom-right).}
\label{fig:2cable_cluster+procPC}
\end{figure}

\subsubsection*{Cables with no occlusions}

\begin{figure}[th]
    \centering
    \includegraphics[width=0.4\columnwidth]{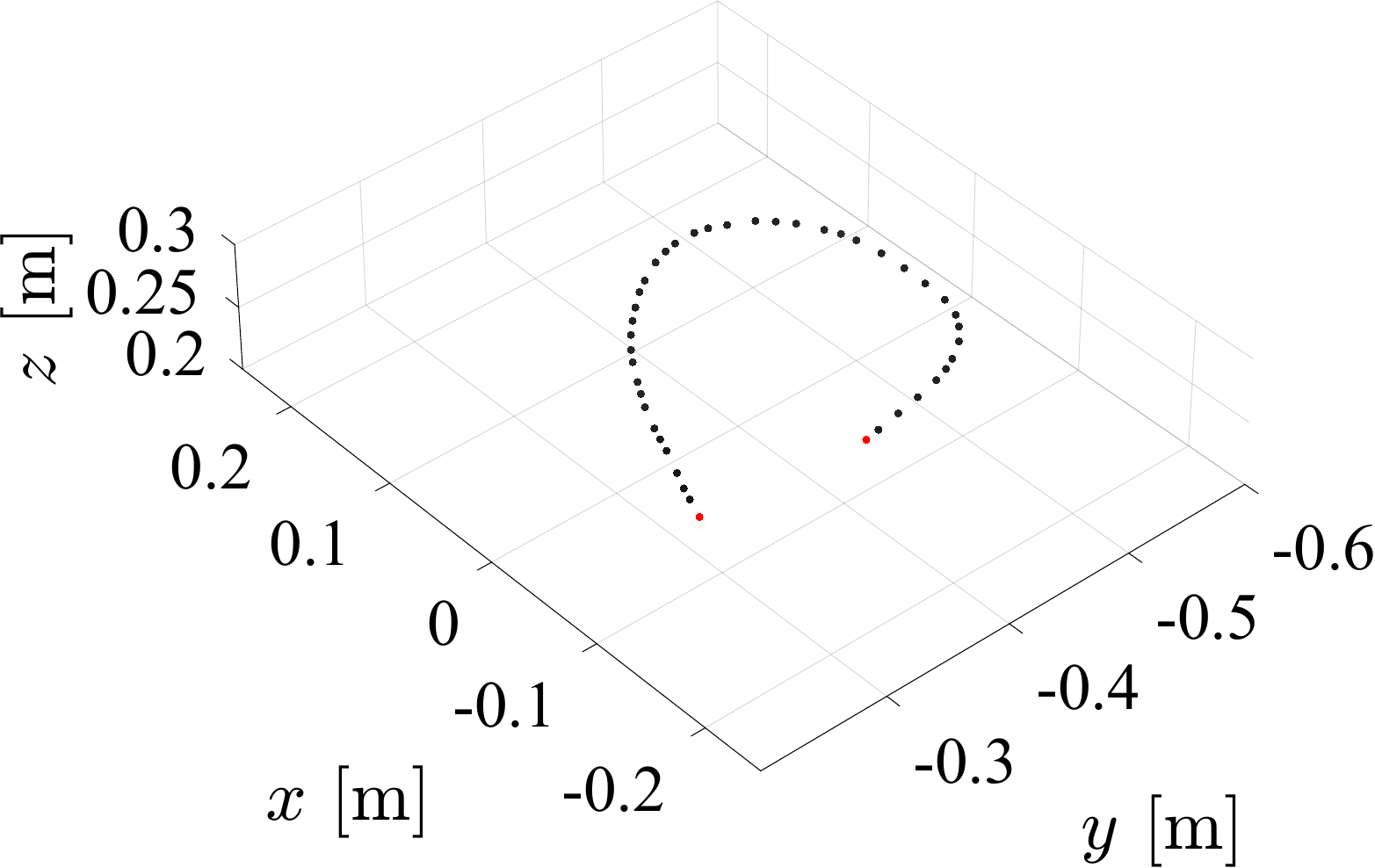}
    \includegraphics[width=0.4\columnwidth]{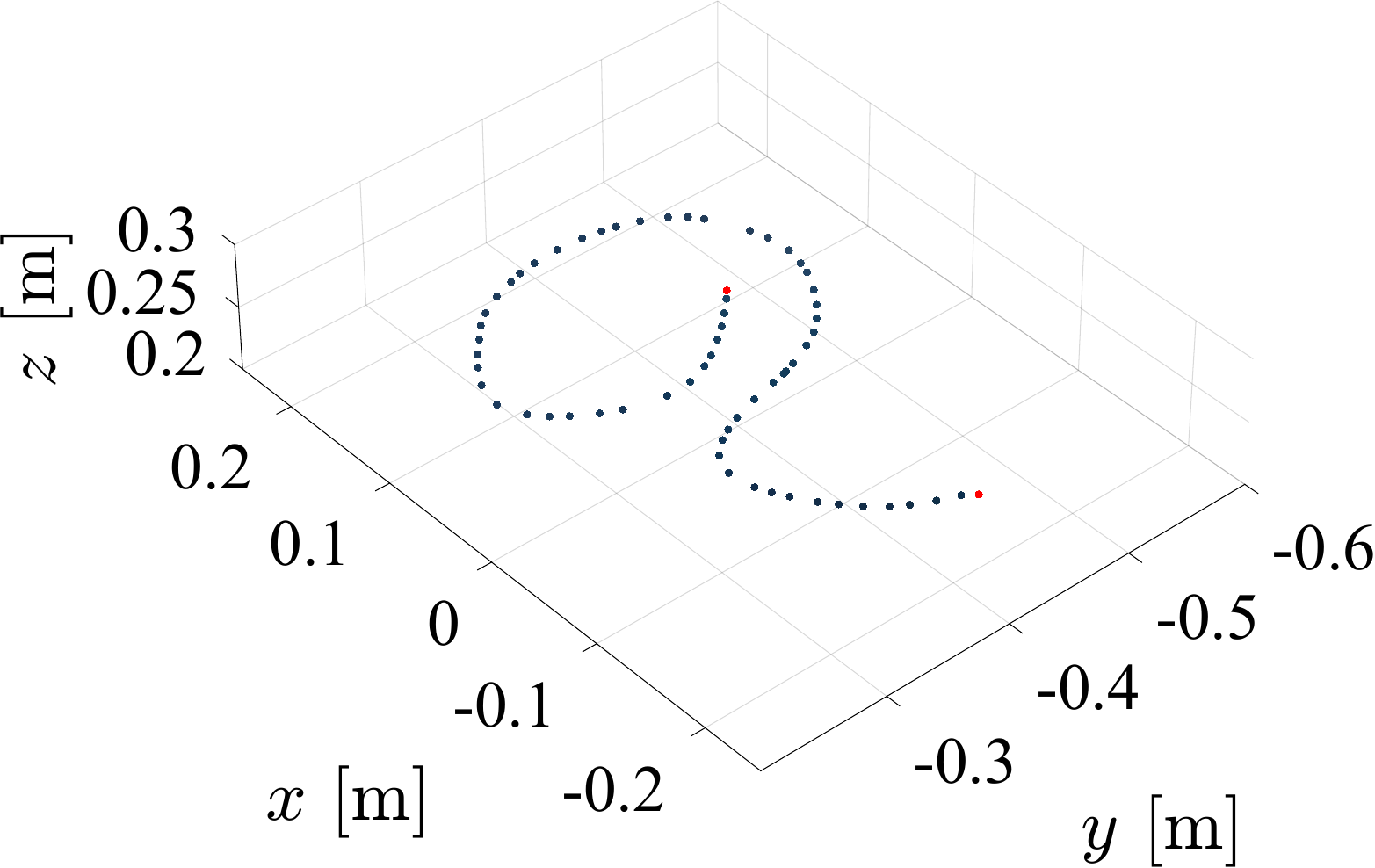}\\[0.5cm]
    \includegraphics[width=0.4\columnwidth]{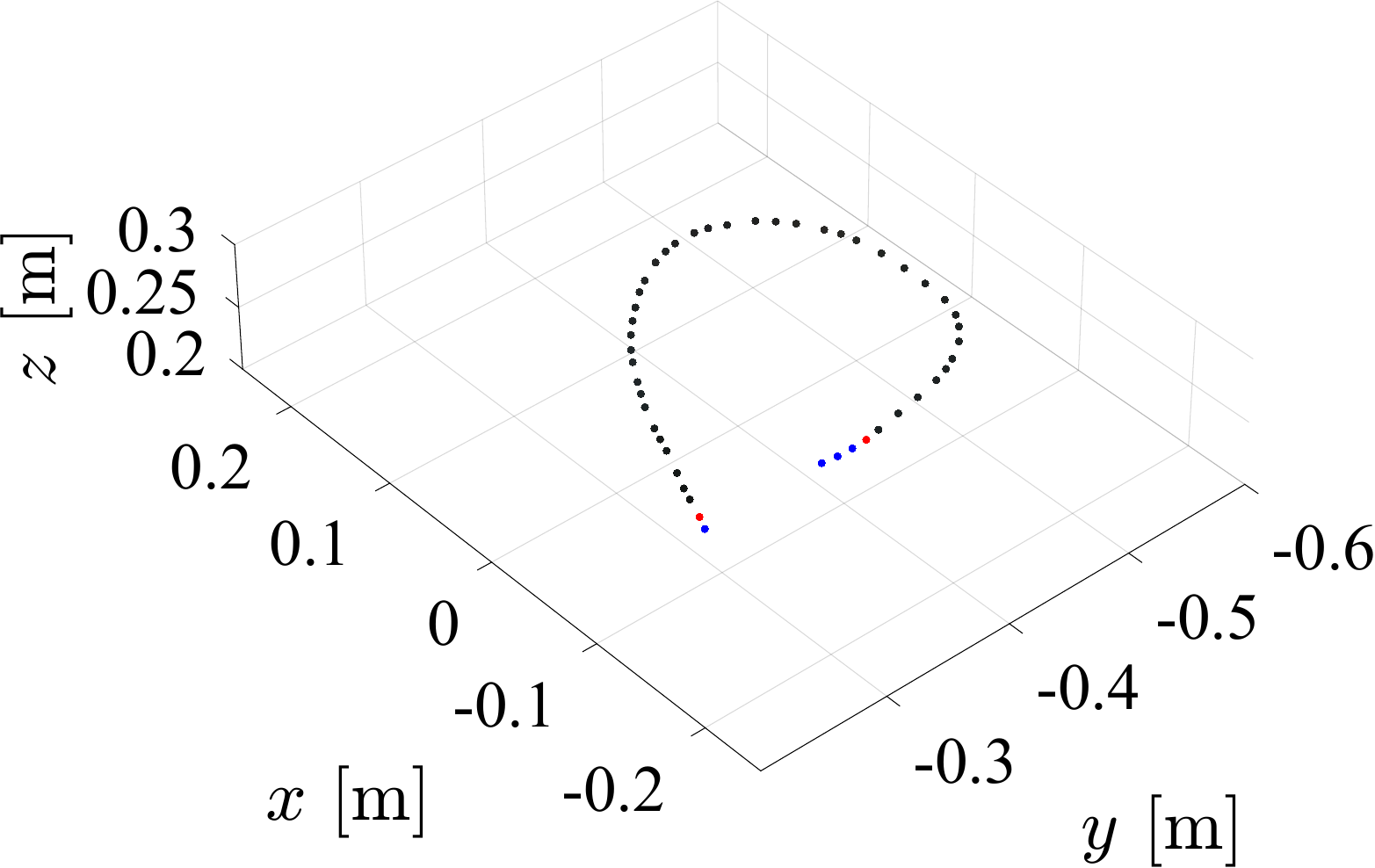}
    \includegraphics[width=0.4\columnwidth]{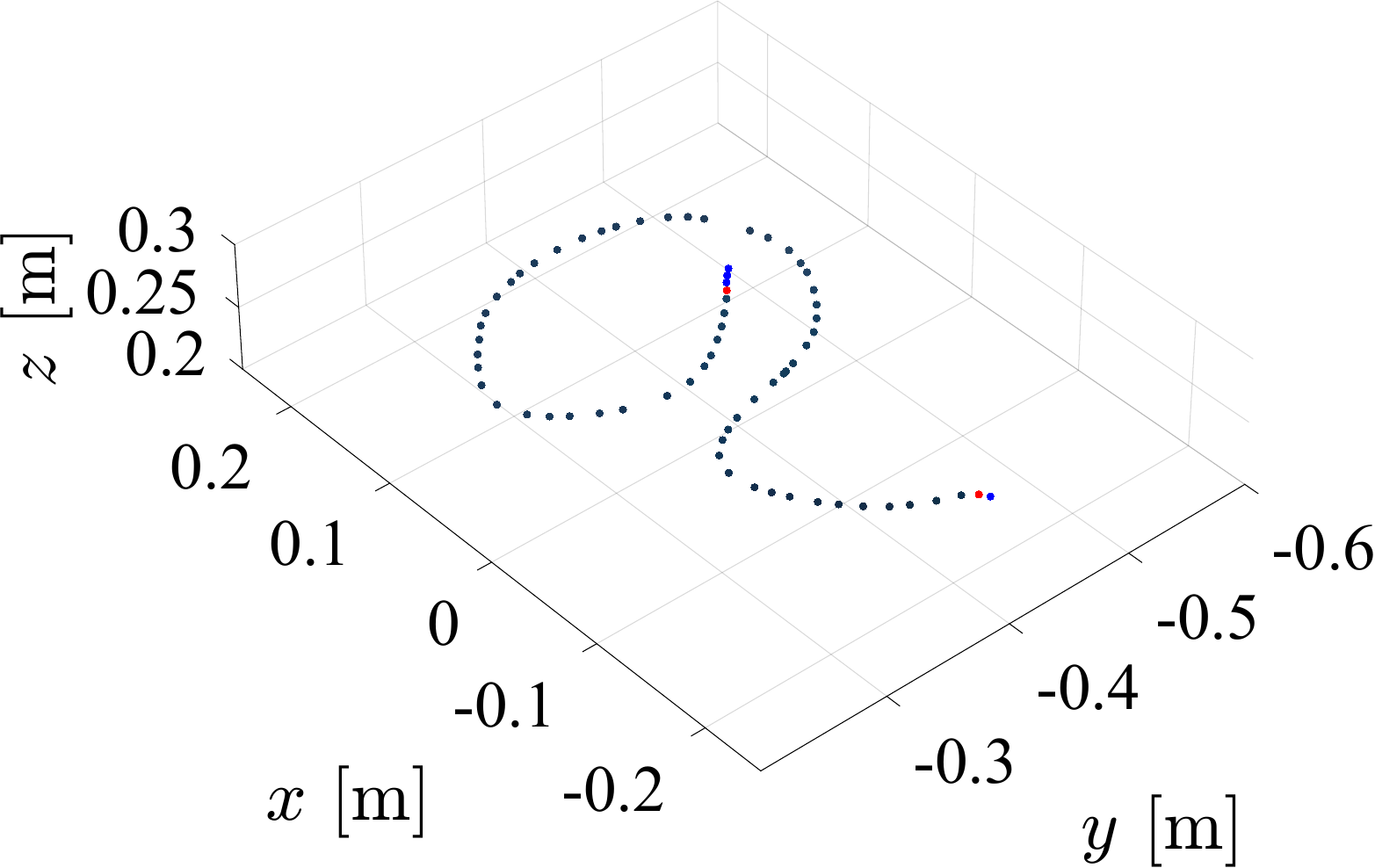}
\caption{CS2 (i): first sorted point cloud of the black cable with endpoints in red (top-left), first sorted point cloud of the blue cable with endpoints in red (top-right), point cloud of the black cable obtained by merging both visual and tactile point clouds (bottom-left) point cloud of the blue cable obtained by merging both visual and tactile point clouds (bottom-right).}
\label{fig:2cable_endpoints+merged}
\end{figure}

\begin{figure}[th]
    \centering
    \includegraphics[width=0.4\columnwidth]{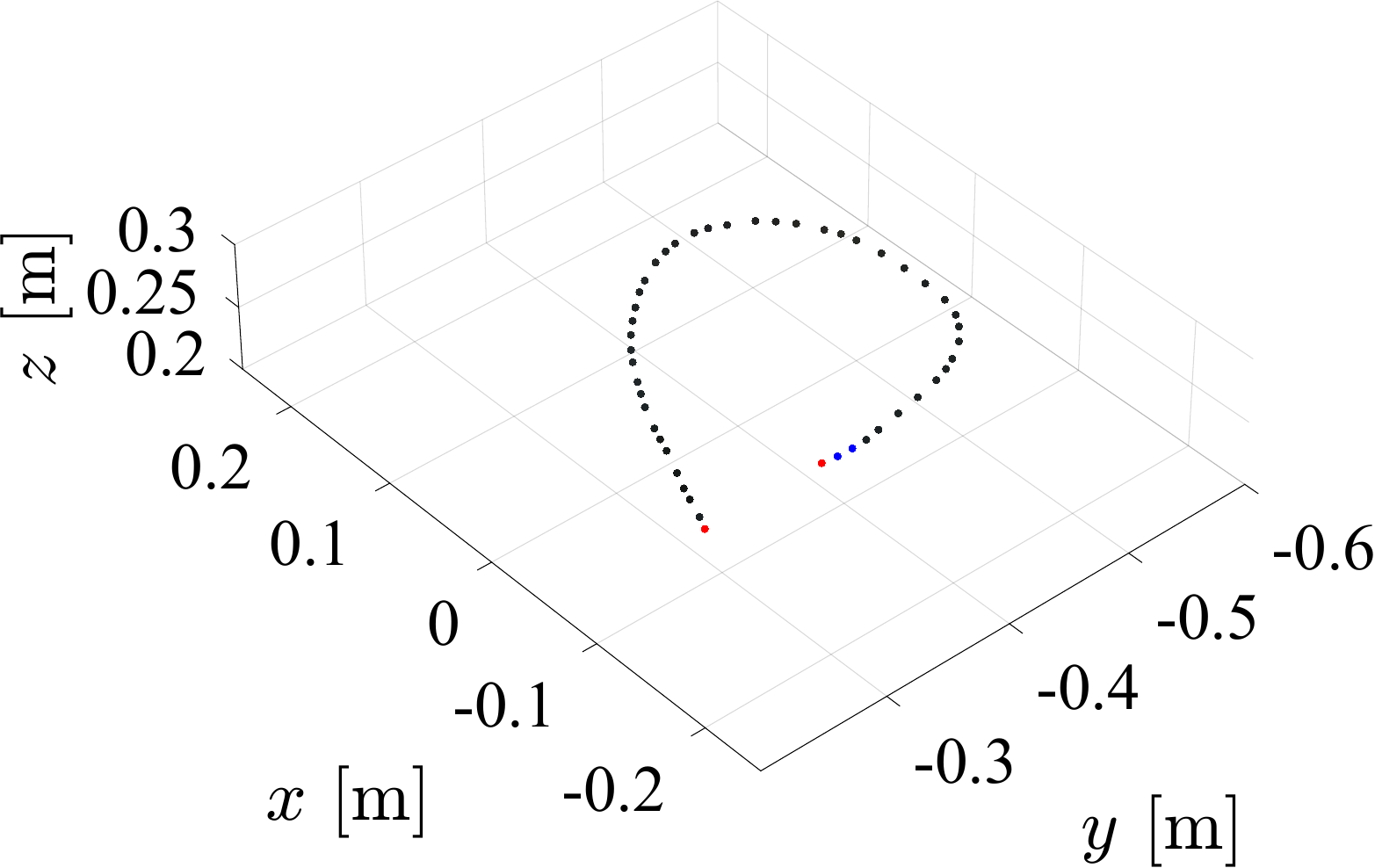}
    \includegraphics[width=0.4\columnwidth]{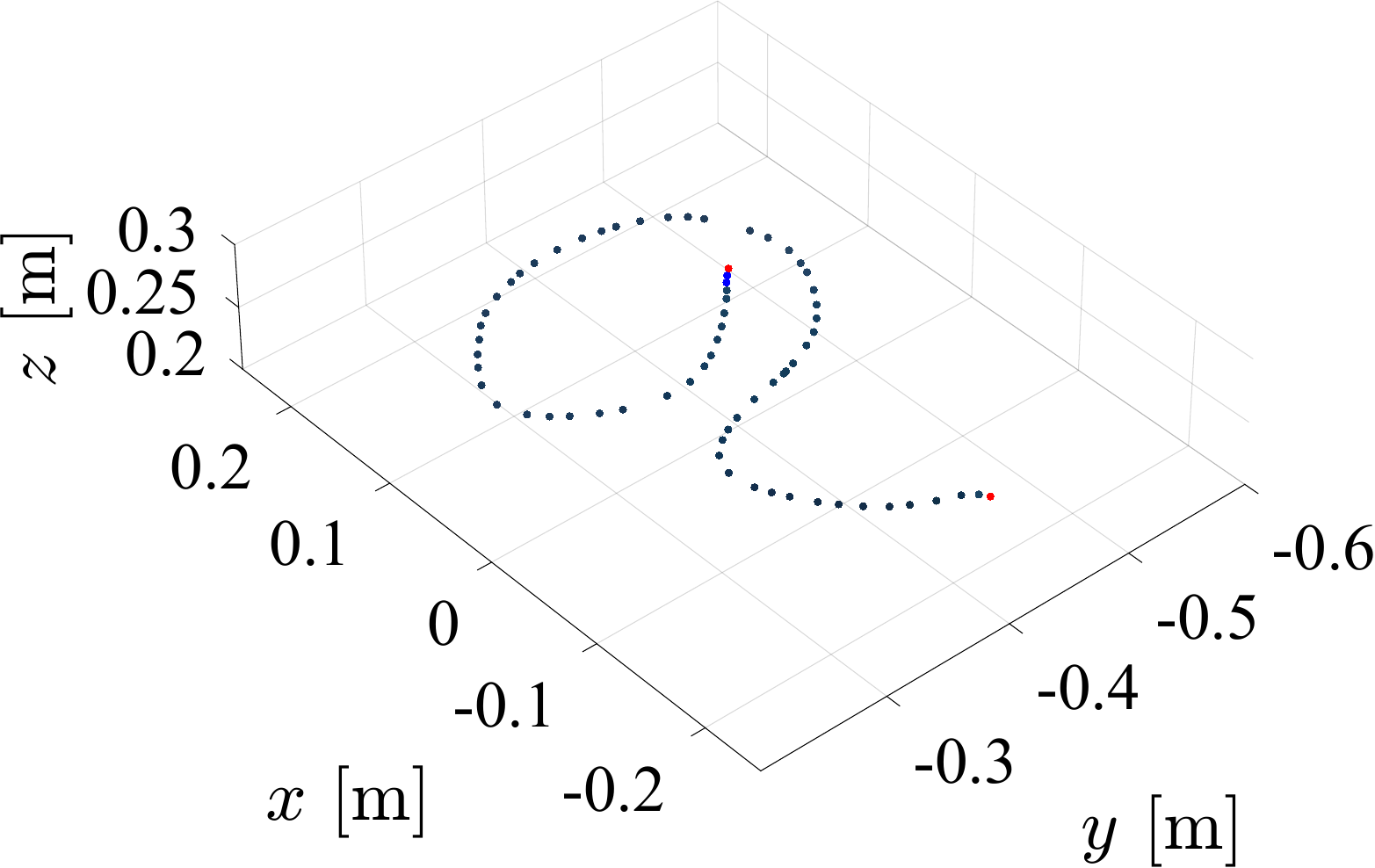}\\[0.5cm]
    \includegraphics[width=0.4\columnwidth]{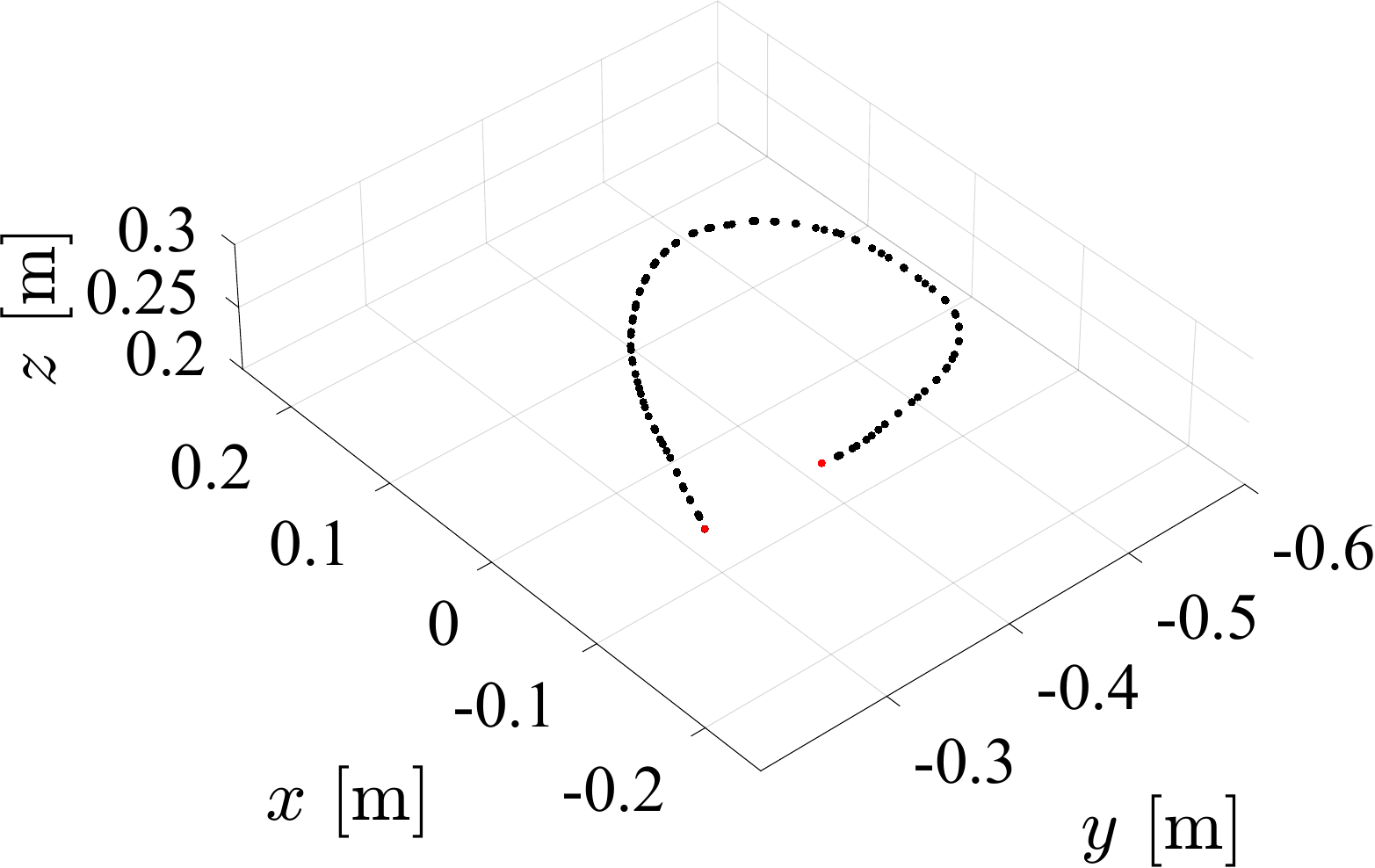}
    \includegraphics[width=0.4\columnwidth]{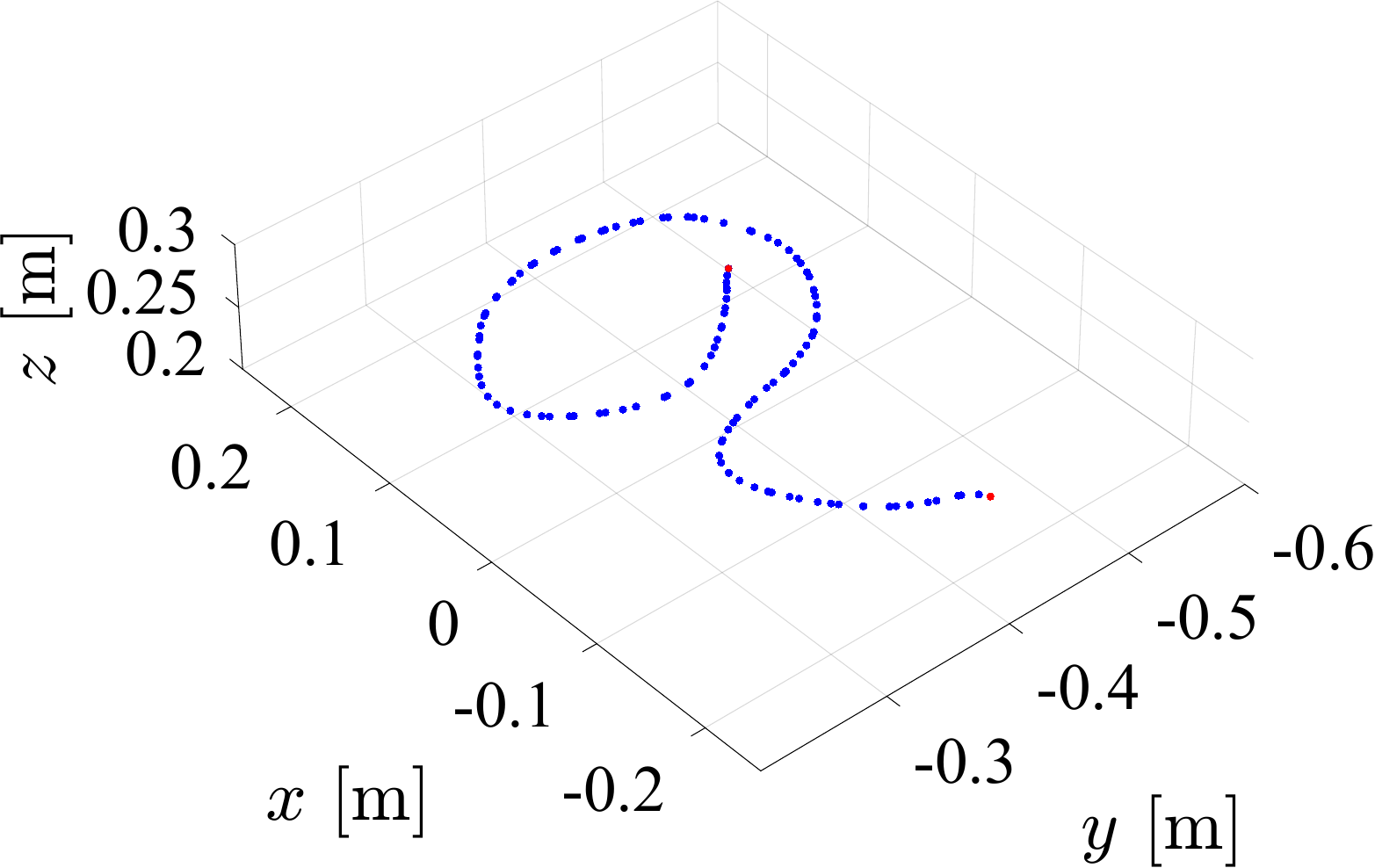}
\caption{CS2 (i): resulting point cloud of the black cable of the new sorting step (top-left), resulting point cloud of the blue cable of the new sorting step(top-right), interpolated point cloud of the black cable (bottom-left), interpolated point cloud of the blue cable (bottom-right).}
\label{fig:2cable_reconstructed+interpolated}
\end{figure}

Fig. \ref{fig:2cable_image+sam2} shows the results of the image processing, while Fig. \ref{fig:2cable_cluster+procPC} illustrates the results of the clustering algorithm and the corresponding post-processed point clouds.
The clusters are independently processed and Fig. \ref{fig:2cable_endpoints+merged} shows the results of the sorting step and of the tactile exploration. For each reconstructed cluster, both the detection of the true endpoints and the interpolation have been performed successfully, as shown in Fig. \ref{fig:2cable_reconstructed+interpolated}.

\begin{figure}[th]
    \centering
    \includegraphics[width=0.4\columnwidth]{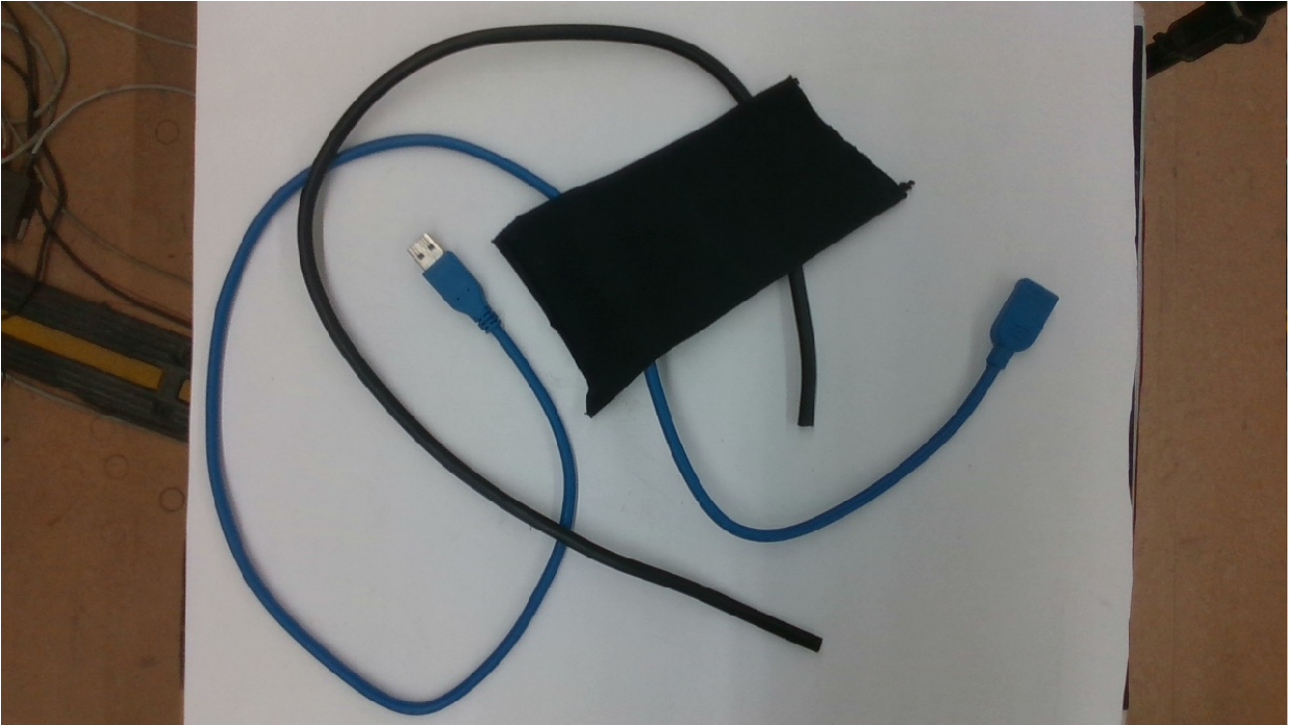}
    \includegraphics[width=0.4\columnwidth]{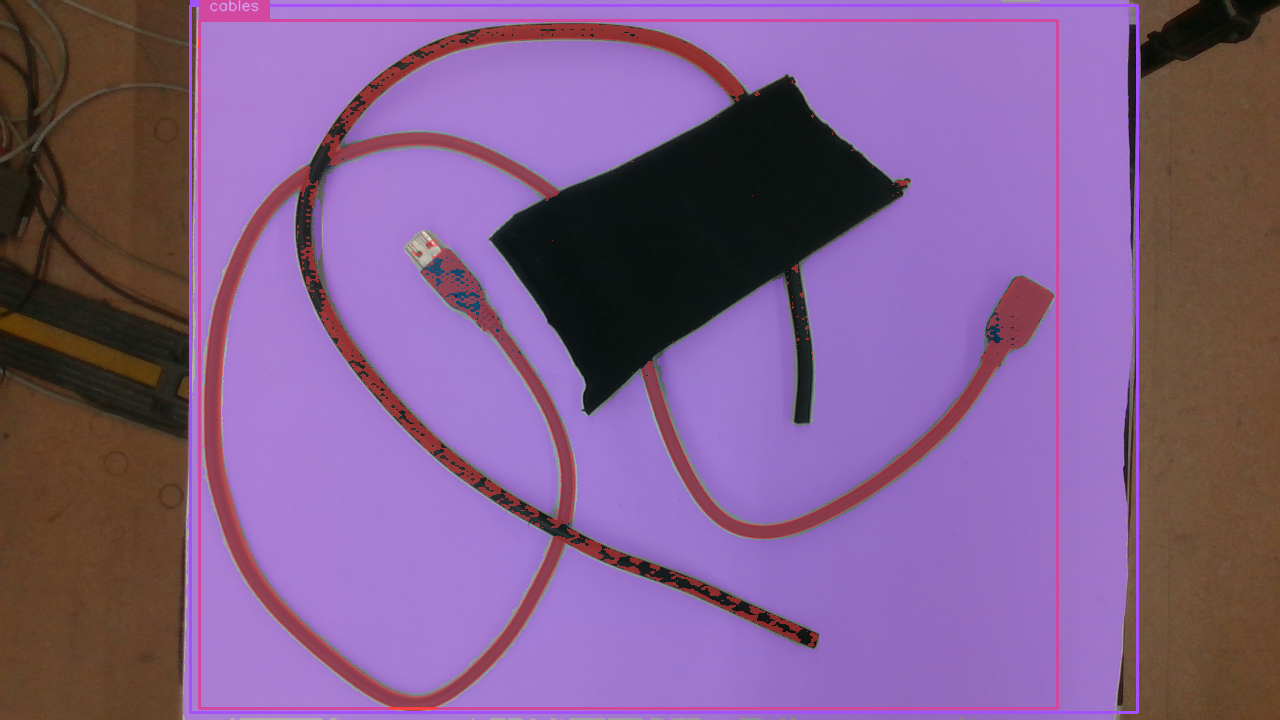}\\[0.5cm]
    \includegraphics[width=0.4\columnwidth]{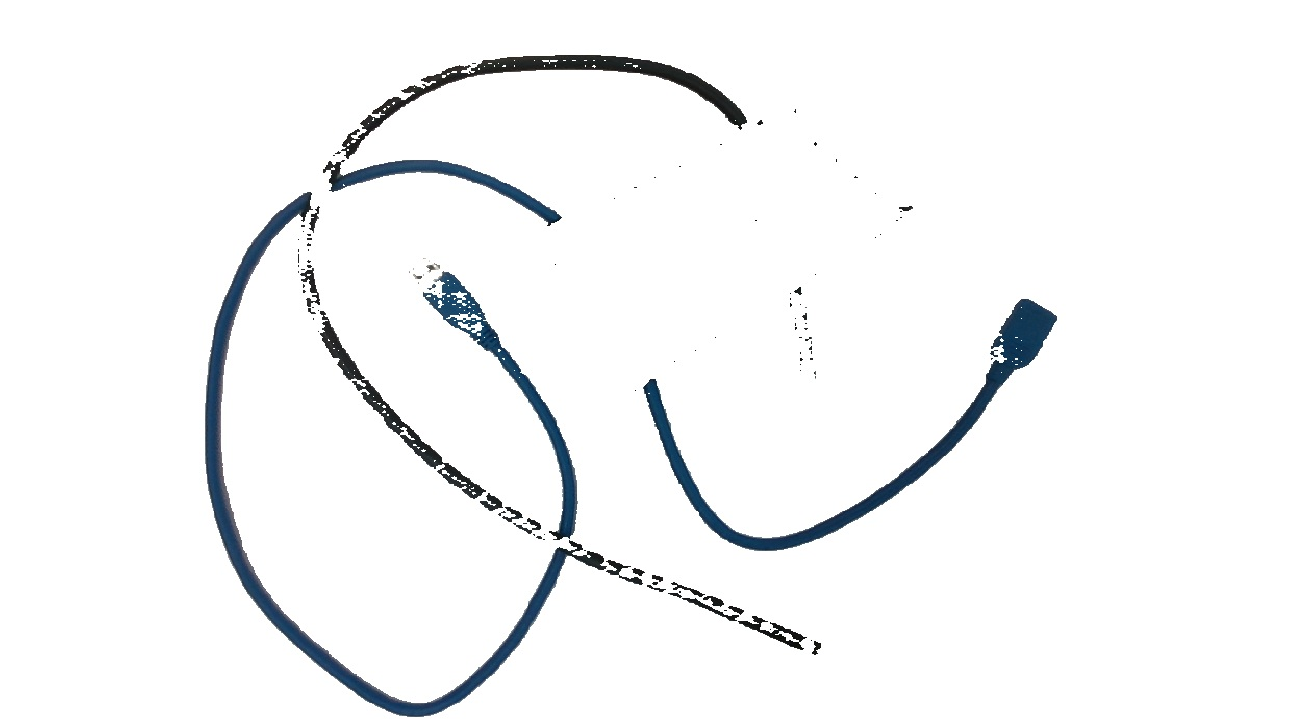}
    \includegraphics[width=0.4\columnwidth]{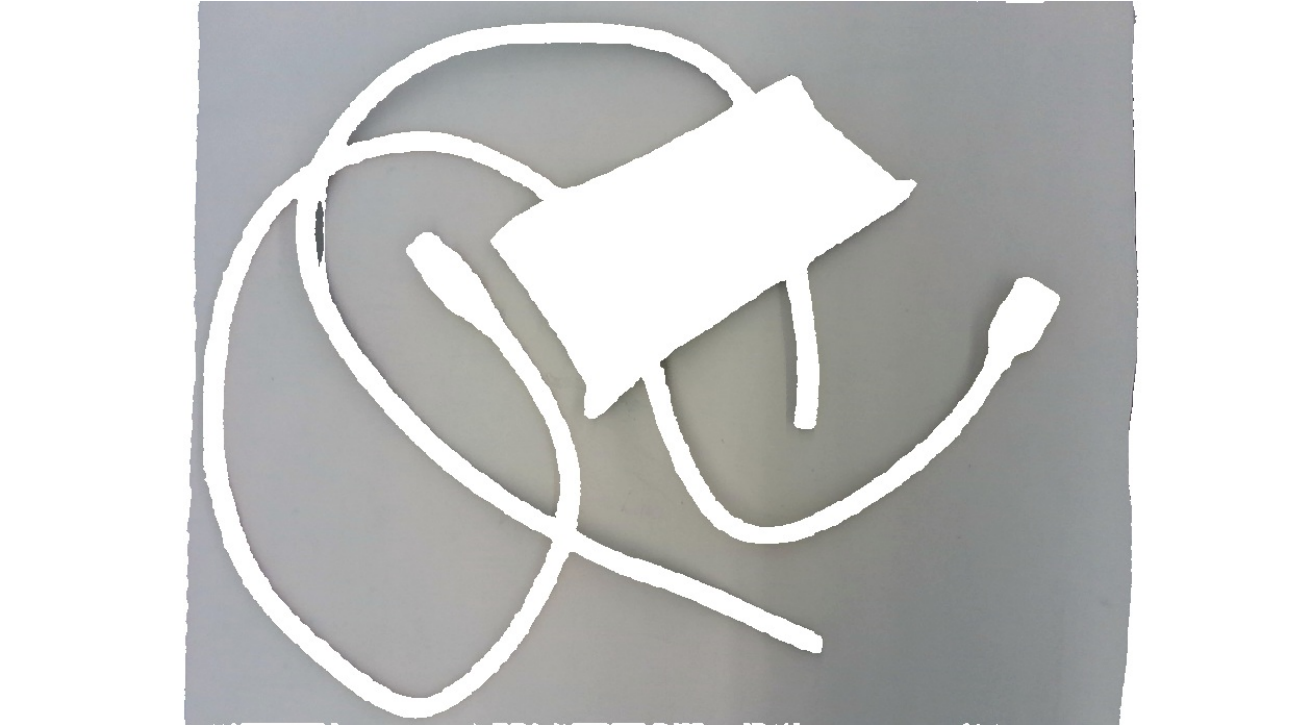}
\caption{CS2 (ii): camera RGB image (top-left), result of the semantic segmentation by Florence2/SAM2 (top-right), segmented cables (bottom-left) and support surface (bottom-right).}
\label{fig:2cable_Occ_image+sam2}
\end{figure}

\subsubsection*{Cables with occlusions}
\begin{figure}[th]
    \centering
    \includegraphics[width=0.4\columnwidth]{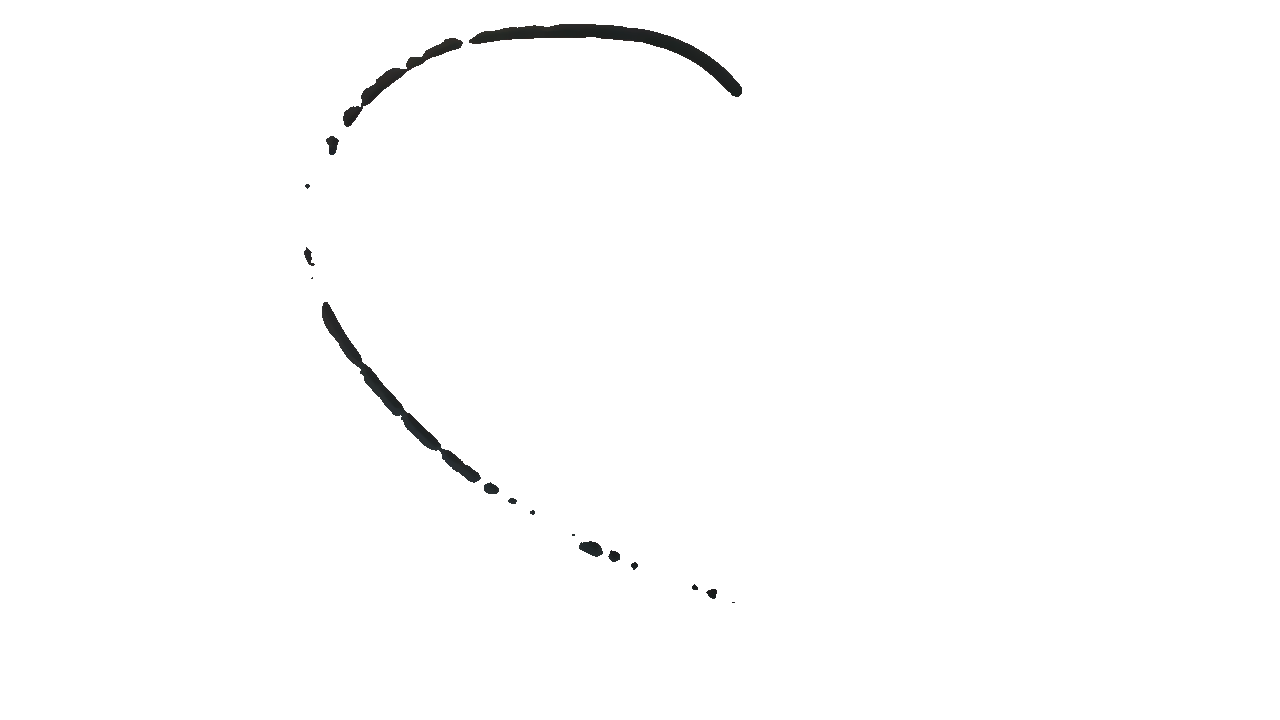}
    \includegraphics[width=0.4\columnwidth]{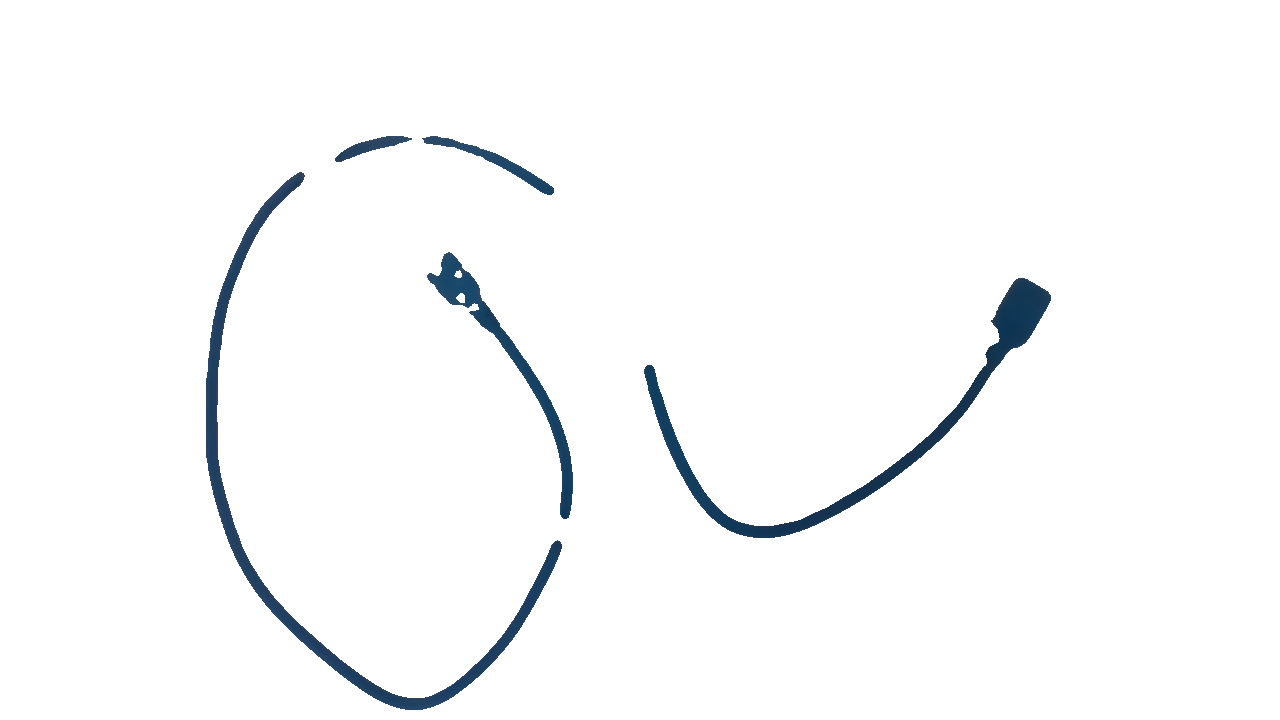}\\[0.5cm]
    \includegraphics[width=0.4\columnwidth]{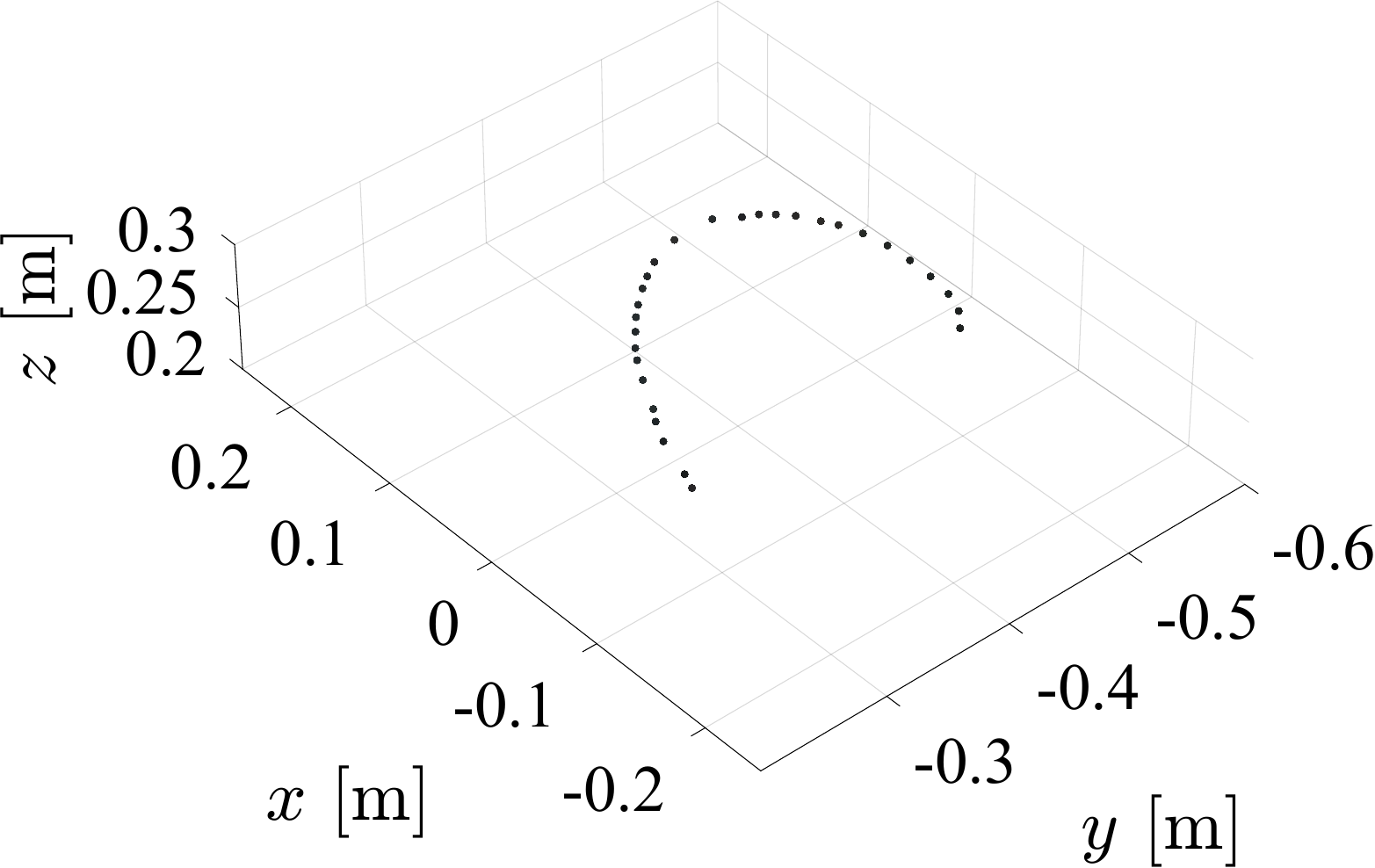}
    \includegraphics[width=0.4\columnwidth]{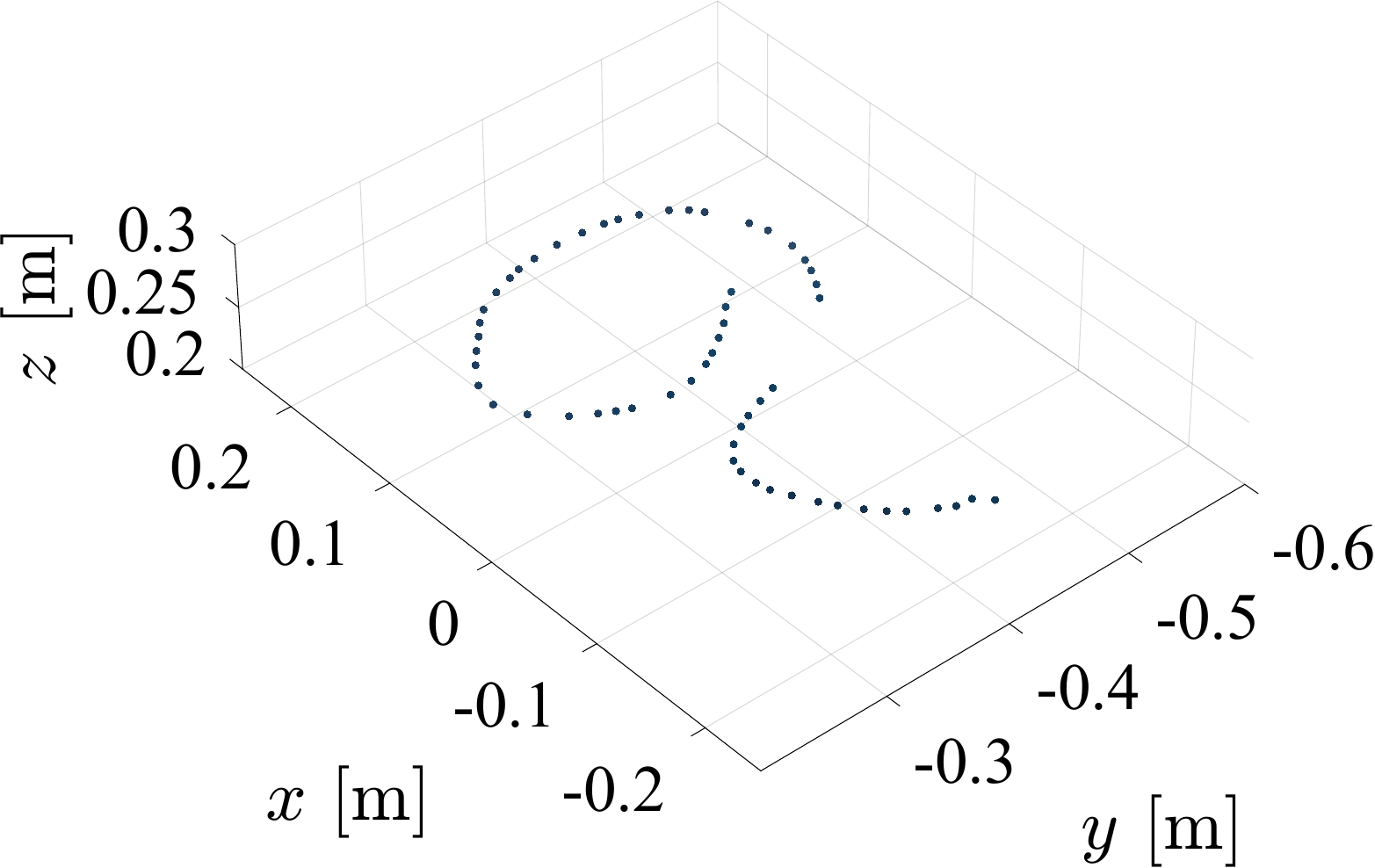}
\caption{CS2 (ii): cluster of the black cable (top-left), cluster of the blue cable (top-right), post-processed point cloud of the black cable (bottom-left) and post-processed point cloud of the blue cable (bottom-right).}
\label{fig:2cable_Occ_cluster+procPC}
\end{figure}

\begin{figure}[th]
    \centering
    \includegraphics[width=0.4\columnwidth]{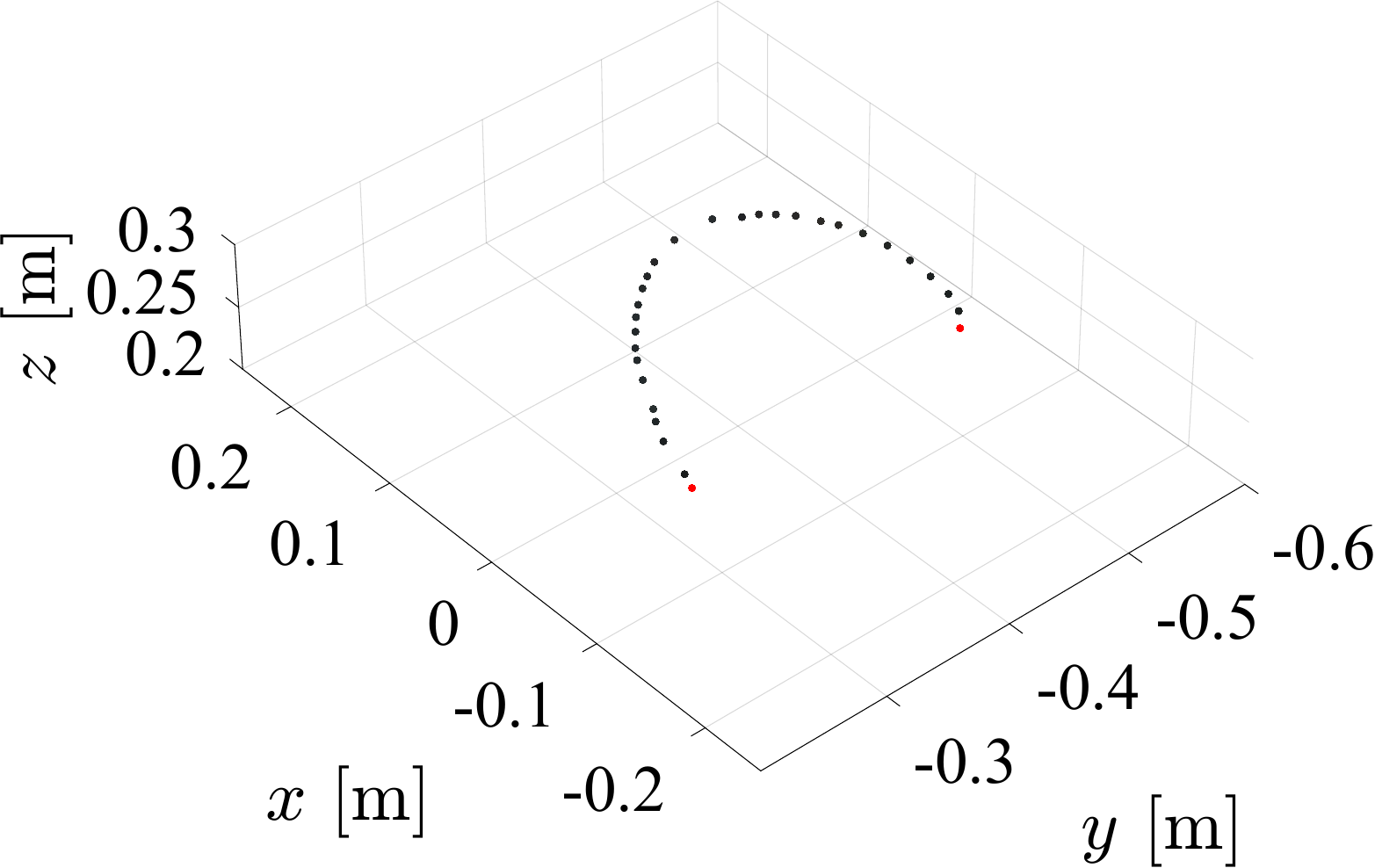}
    \includegraphics[width=0.4\columnwidth]{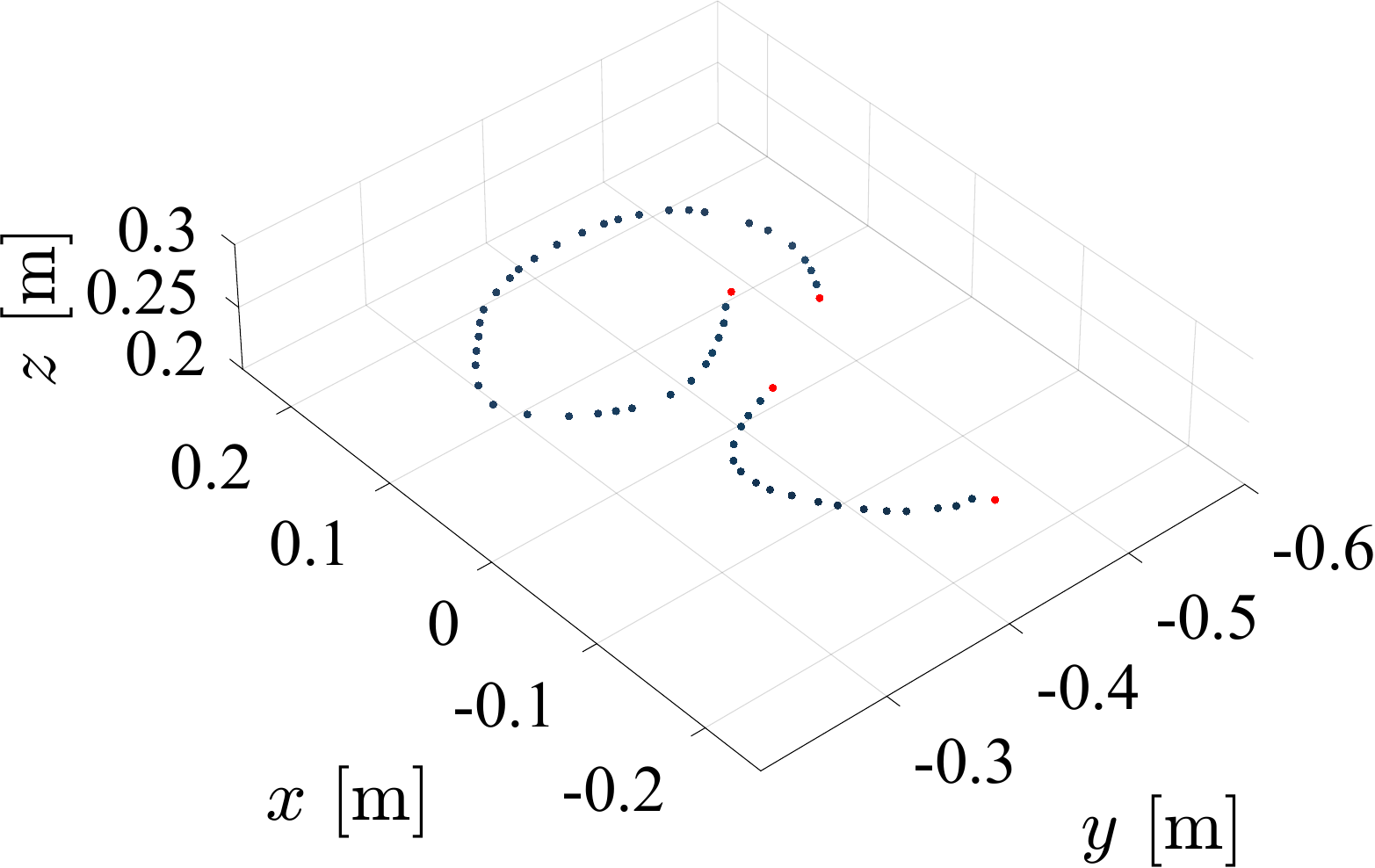}\\[0.5cm]
    \includegraphics[width=0.4\columnwidth]{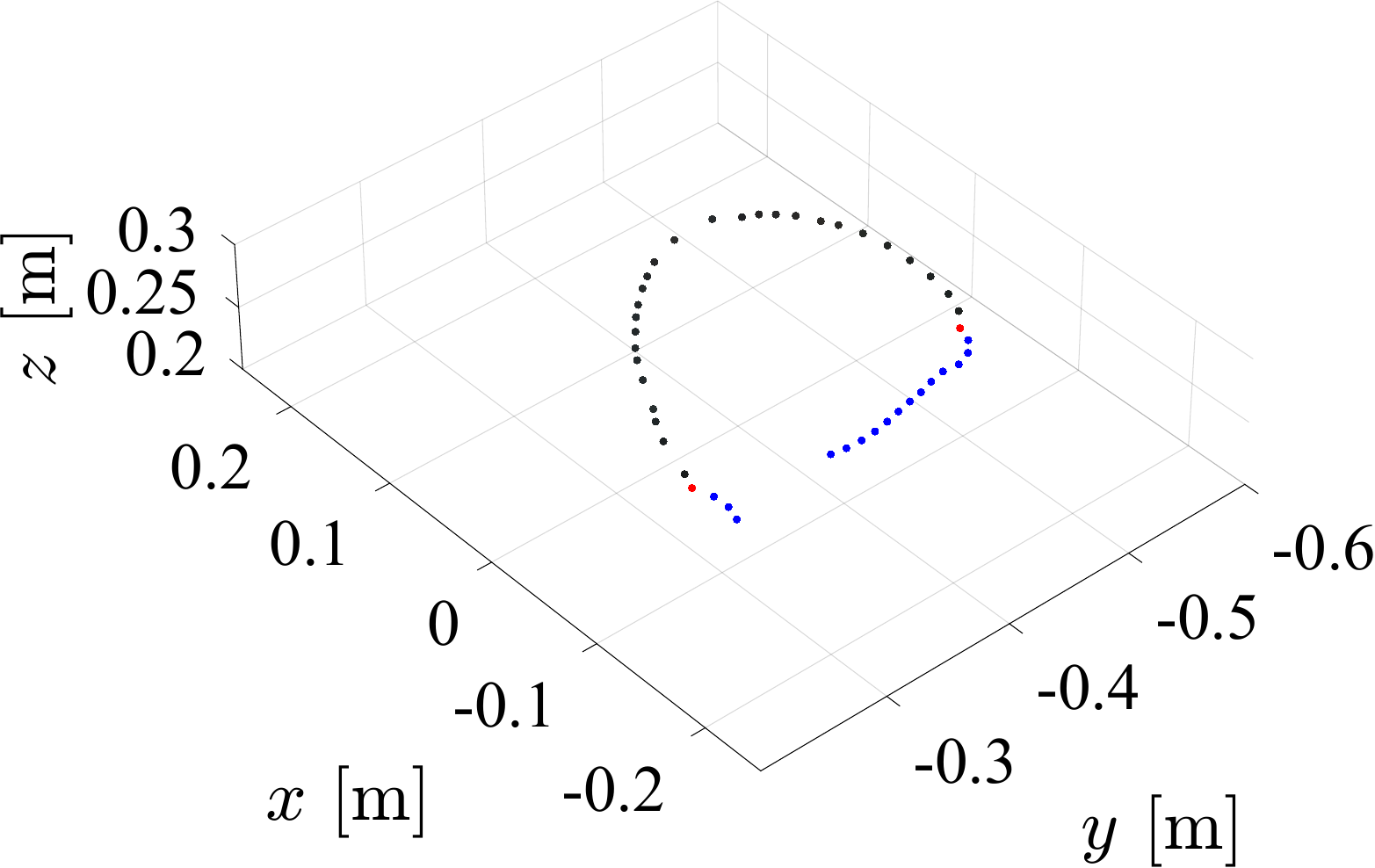}
    \includegraphics[width=0.4\columnwidth]{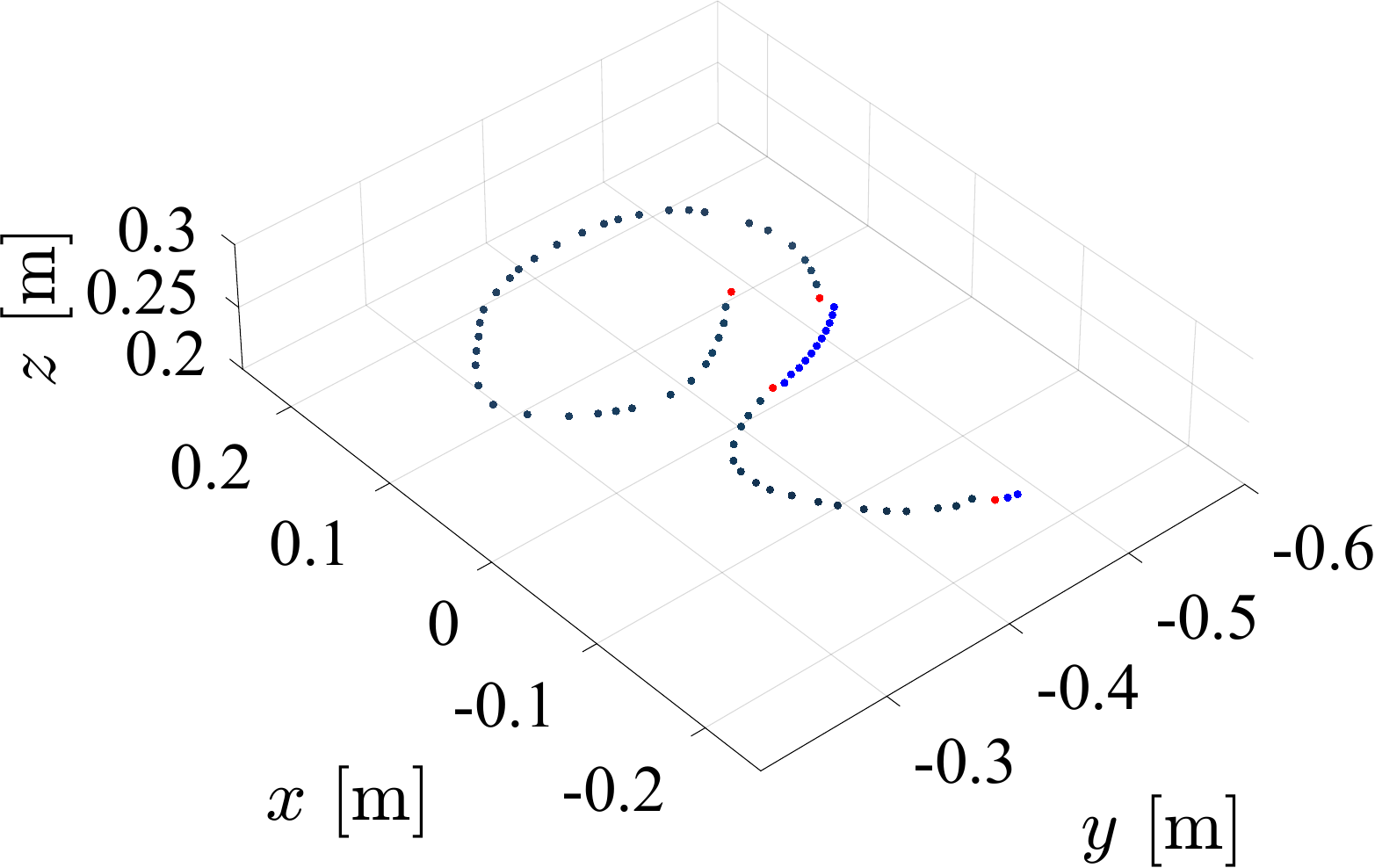}
\caption{CS2 (ii): first sorted point cloud of the black cable with endpoints in red (top-left), first sorted point cloud of the blue cable with endpoints in red (top-right), point cloud of the black cable obtained by merging both visual and tactile point clouds (bottom-left) point cloud of the blue cable obtained by merging both visual and tactile point clouds (bottom-right).}
\label{fig:2cable_Occ_endpoints+merged}
\end{figure}

\begin{figure}[th]
    \centering
    \includegraphics[width=0.4\columnwidth]{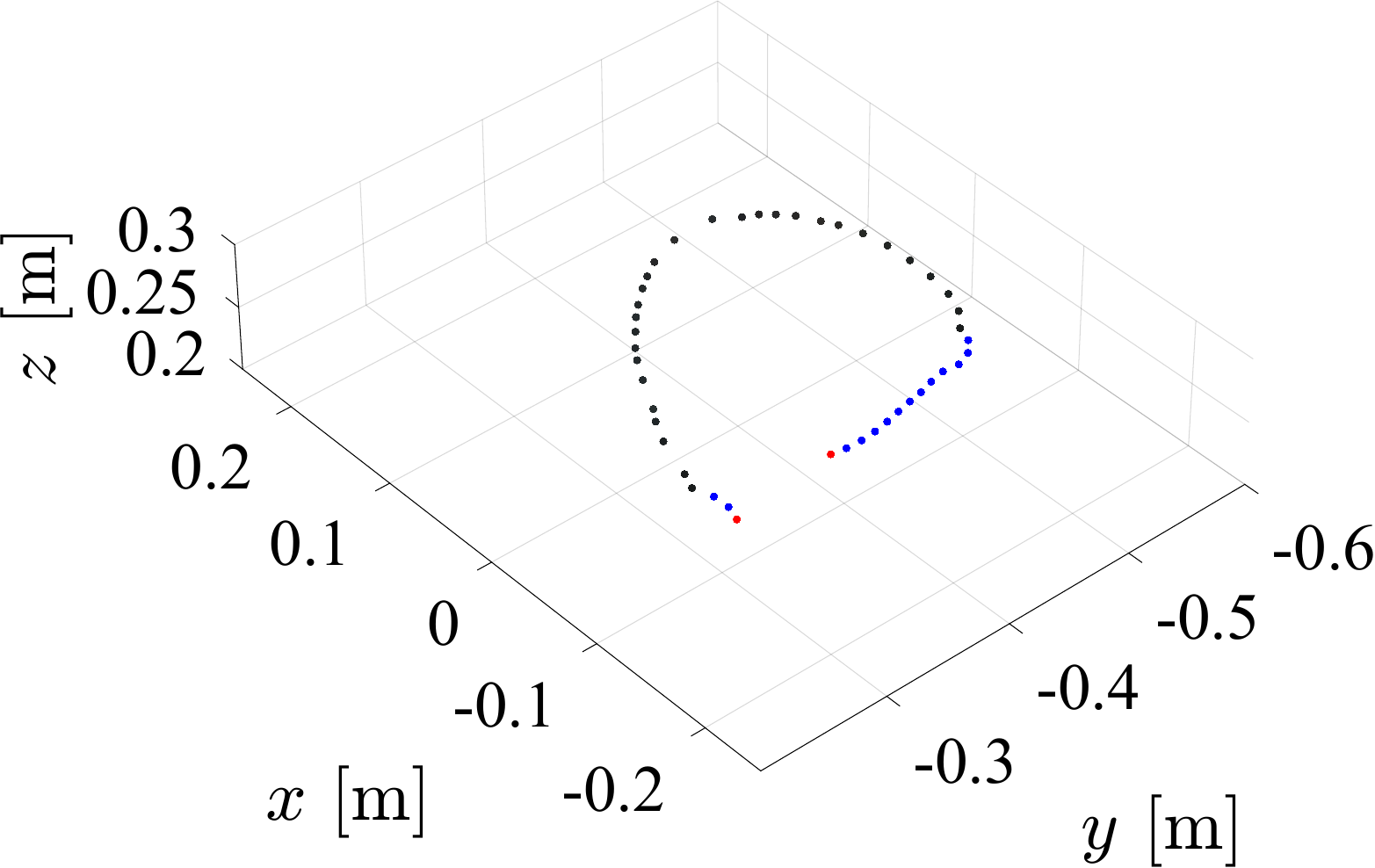}
    \includegraphics[width=0.4\columnwidth]{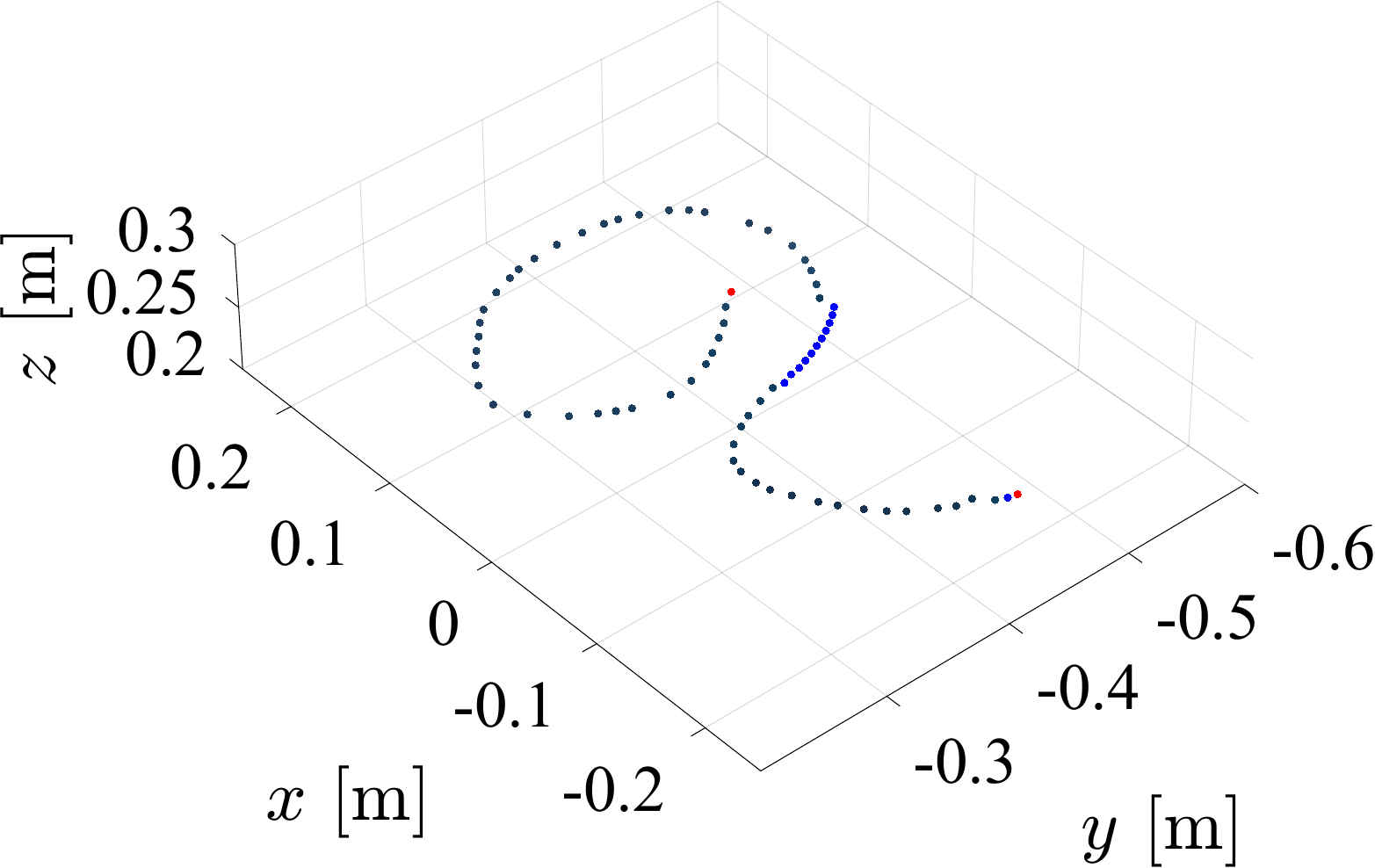}\\[0.5cm]
    \includegraphics[width=0.4\columnwidth]{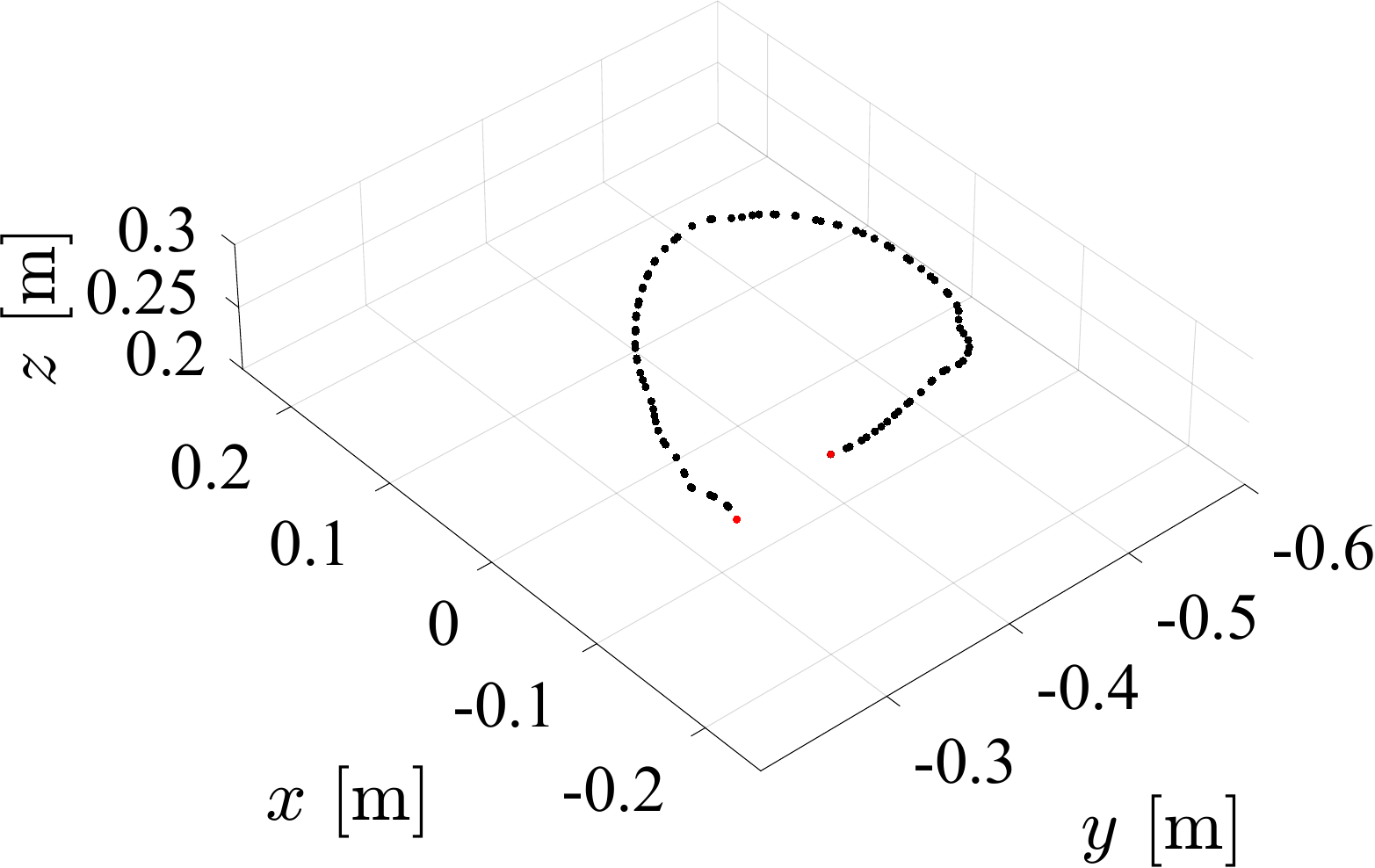}
    \includegraphics[width=0.4\columnwidth]{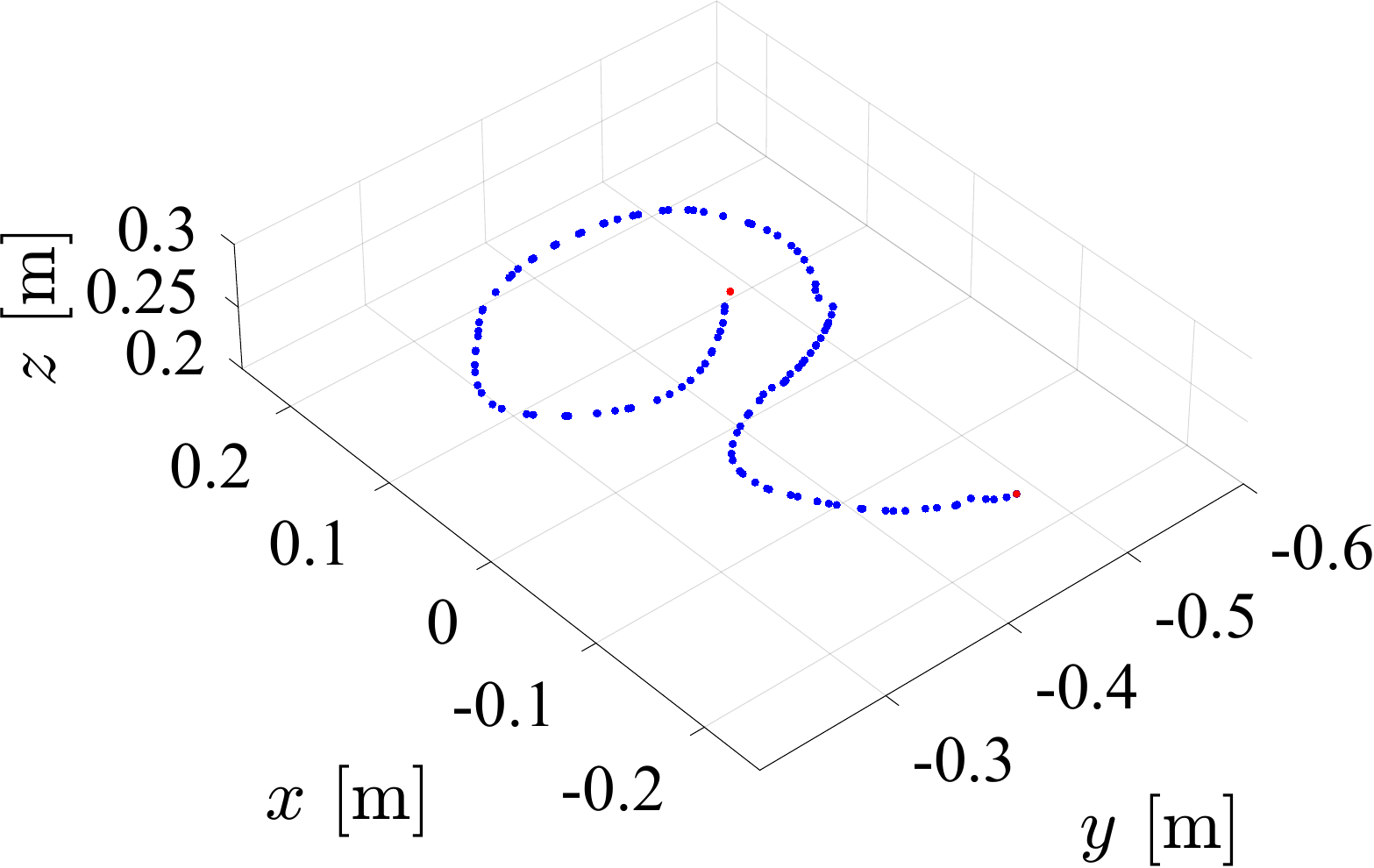}
\caption{CS2 (ii): resulting point cloud of the black cable of the new sorting step (top-left), resulting point cloud of the blue cable of the new sorting step(top-right), interpolated point cloud of the black cable (bottom-left), interpolated point cloud of the blue cable (bottom-right).}
\label{fig:2cable_Occ_reconstructed+interpolated}
\end{figure}

Fig. \ref{fig:2cable_Occ_image+sam2} reports the acquired image and the results of the semantic segmentation. Although the bounding boxes are correctly placed, SAM2 was not able to segment the cables correctly (in particular the black one): some pixels do not belong to the segmentation mask identified by the red colour in Fig. \ref{fig:2cable_Occ_image+sam2} top-right. Therefore, tactile exploration is useful to reconstruct the missing parts arising from semantic segmentation. The results of the clustering and the corresponding post-processed point clouds are shown in Fig. \ref{fig:2cable_Occ_cluster+procPC}. Fig. \ref{fig:2cable_Occ_endpoints+merged} reports the results of the first sorting and of the tactile exploration. Fig. \ref{fig:2cable_Occ_reconstructed+interpolated} illustrates the detection of the true endpoints of the cables and the interpolated point clouds.

\subsection{Evaluation of the reconstruction accuracy}
\begin{figure}[th]
    \centering
    \includegraphics[width=0.4\columnwidth]{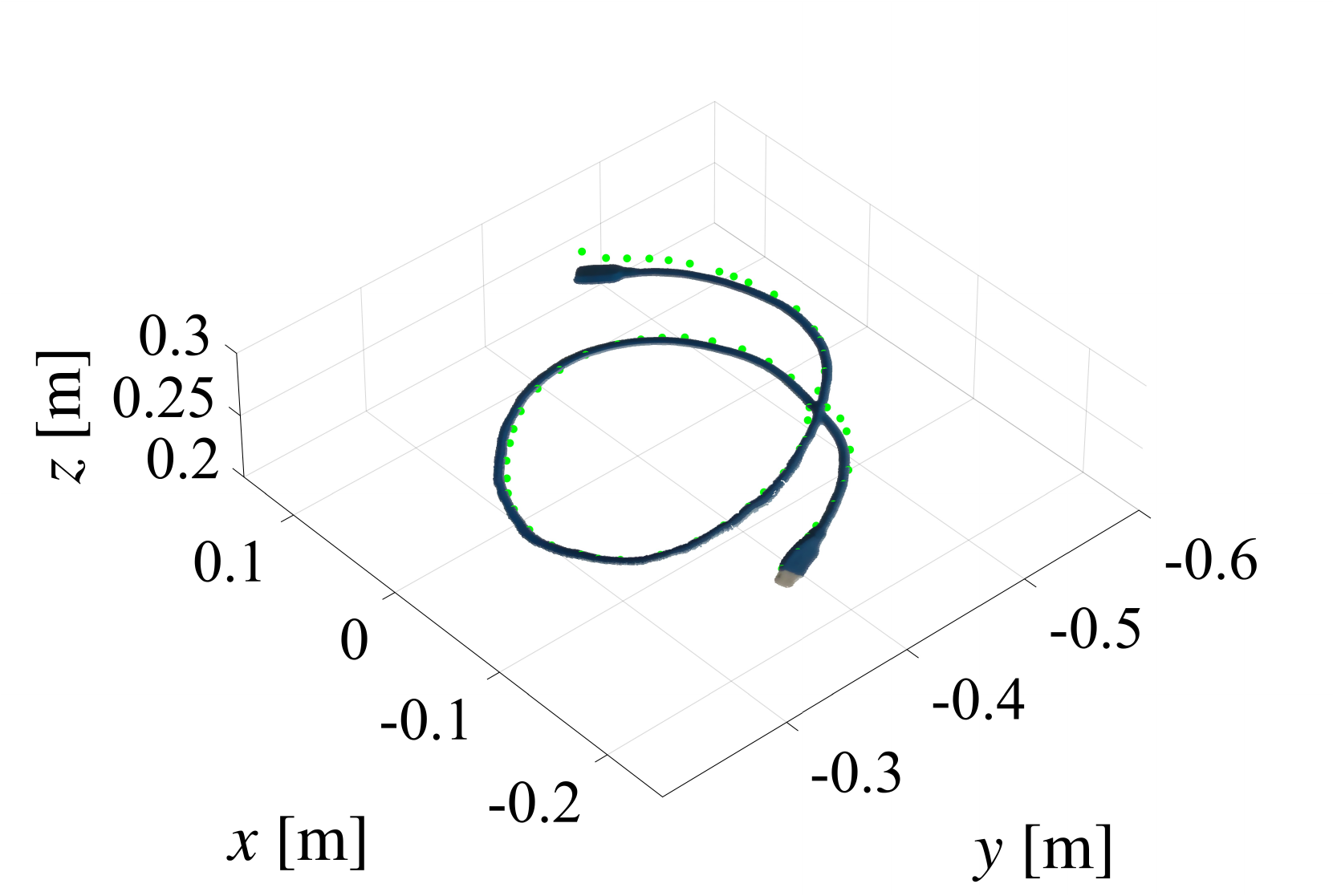}\\
    \includegraphics[width=0.4\columnwidth]{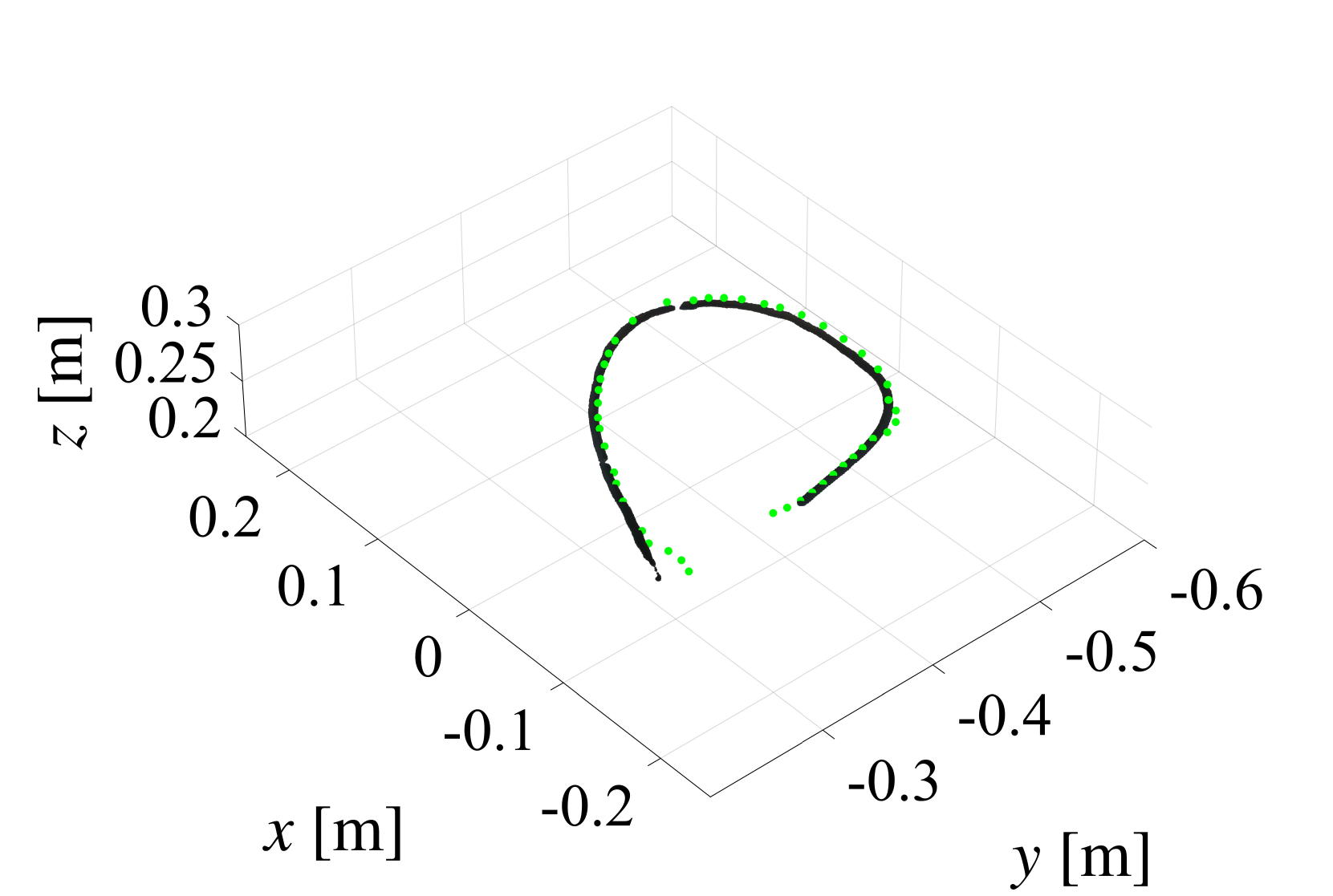}
    \includegraphics[width=0.4\columnwidth]{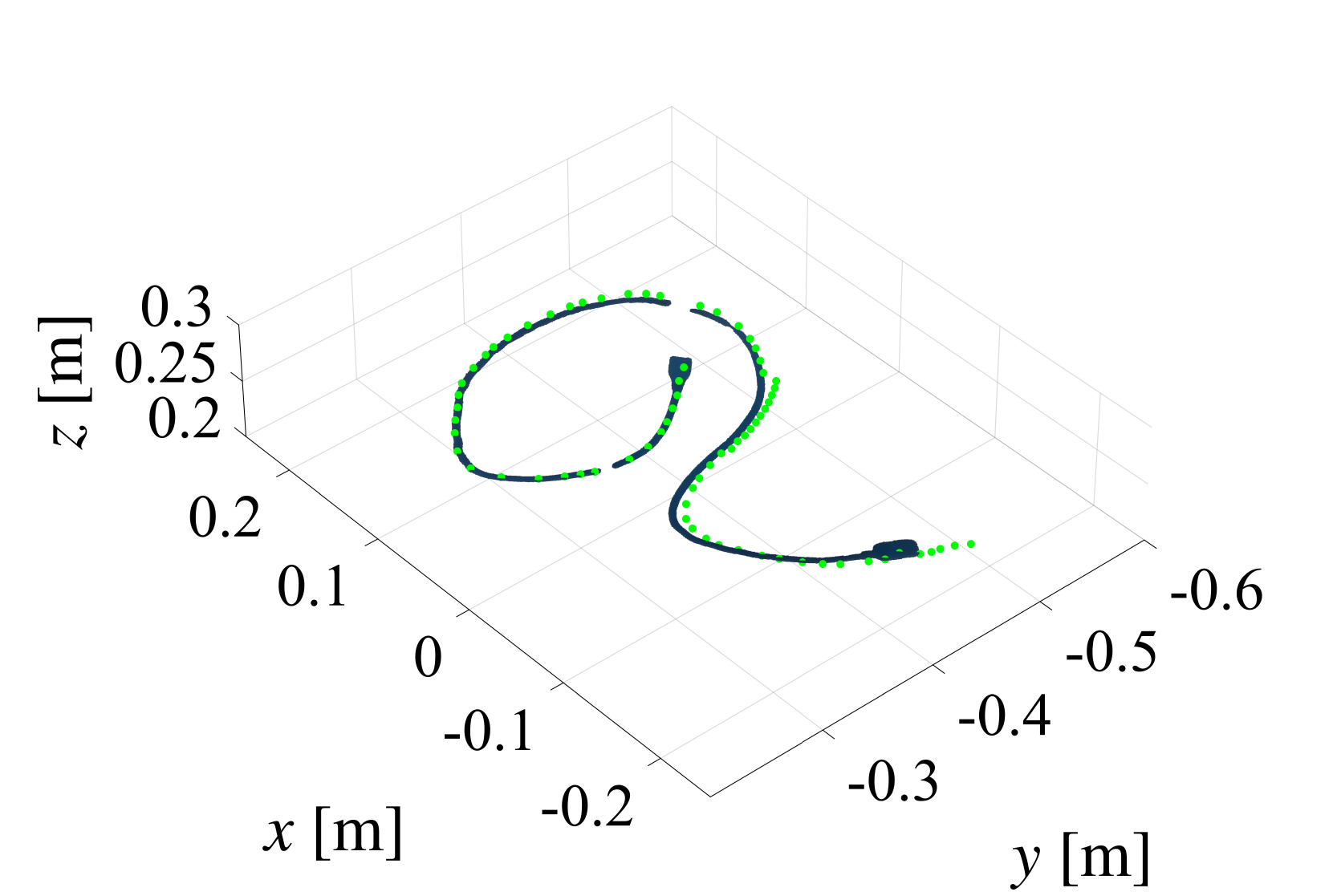}
\caption{Reconstructed point cloud (green dots) of the cable in CS1 (ii) overlapped to the dense target point cloud of the segmented cable in CS1 (i) (top), reconstructed point cloud (green dots) of the black cable in CS2 (ii) overlapped to the dense target point cloud of the segmented cable in CS2 (i) (bottom-left), reconstructed point cloud (green dots) of the blue cable in CS2 (ii) overlapped to the dense target point cloud of the segmented cable in CS2 (i) (bottom-right).}
\label{fig:ICP_evaluation}
\end{figure}

The \textit{Iterative Closest Point} (ICP) algorithm performs registration between two point clouds, specifically aligning a source point cloud with a target one \cite{Besl92}. It can be used to numerically evaluate the accuracy of the reconstruction through the root mean squared error of the performed alignment. The target point clouds are the dense point clouds of the cables obtained by transforming the segmented cables into a point cloud before applying any operation in the scenarios with no occlusions, while the source ones are the reconstructed point clouds obtained by performing the tactile exploration in the presence of occlusions. Fig. \ref{fig:ICP_evaluation} represents the overlapped source and target point clouds for each cable in the presented case studies; Table \ref{tab:icp_table} shows the corresponding root mean squared errors. A direct quantitative comparison with existing approaches is not included, since most prior methods focus on vision-only perception and are not designed for cross-modal visuo–tactile reconstruction with active tactile exploration.

\begin{table}[htbp]
\centering
\caption{Root mean squared errors of the ICP in the two case studies}
\begin{tabular}{|c|c|c|}
\hline
CS1 [m] & CS2 (black cable) [m] & CS2 (blue cable) [m]\\
\hline
0.00174 & 0.00650 & 0.00459 \\
\hline
\end{tabular}
\label{tab:icp_table}
\end{table}

\section{Conclusion}
This paper presented a cross-modal visuo-tactile framework for reconstructing deformable linear objects under partial observability. By combining foundation-model-based visual perception with autonomous tactile exploration, the proposed method addresses the limitations of vision-only pipelines in the presence of occlusions and imperfect segmentation.
The approach exploits a topology-aware representation based on skeletonization and endpoint-guided point sorting, enabling principled fusion of visual and tactile point clouds, and a unified B-spline reconstruction of the cable geometry. Experimental results on a real robotic platform demonstrate reliable reconstruction across a variety of scenarios, including occlusions, inclined planes, self-intersections, and multiple cables.
While the method shows robustness, endpoint detection can become sensitive in cases of dense point clouds arising from small-angle self-intersections. 
Future work will focus on improving topological reasoning and adaptive resampling, as well as tighter integration between perception and manipulation. Overall, the results indicate that foundation-model-enhanced visuo-tactile perception is a promising direction for robust deformable-object reconstruction in real-world robotic applications. 

\bibliographystyle{IEEEtran}

\bibliography{IEEEabrv,bibtex_clean}


 





\end{document}